\documentclass{article}

\PassOptionsToPackage{numbers}{natbib}

\usepackage[preprint]{neurips_2024}

\usepackage[utf8]{inputenc} %
\usepackage[T1]{fontenc}    %
\usepackage{hyperref}       %
\usepackage{url}            %
\usepackage{booktabs}       %
\usepackage{amsfonts}       %
\usepackage{nicefrac}       %
\usepackage{microtype}      %
\usepackage{xcolor}         %
\usepackage{graphicx}
\usepackage{caption}
\usepackage{subcaption}
\usepackage{amsmath,amsfonts,bm,amsthm}
\usepackage[noend]{algpseudocode} 
\usepackage[ruled, linesnumbered, commentsnumbered]{algorithm2e}
\usepackage{adjustbox}
\usepackage{titlesec}
\usepackage{multirow}
\usepackage{hyperref}
\usepackage{changepage}

\titleformat{\subsubsection}[runin]
  {\normalfont\bfseries}{}
  {0em}
  {}

\titlespacing{\subsubsection}
  {0pt}{\parskip}{1em}

\title{Neural Hamilton: \\ Can A.I. Understand Hamiltonian Mechanics?}

\author{Tae-Geun Kim \\
Department of Physics\\
Yonsei University\\
Seoul, 03722, Republic of Korea \\
\texttt{tg.kim@yonsei.ac.kr}
\And
Seong Chan Park \\
Department of Physics\\
Yonsei University\\
Seoul, 03722, Republic of Korea \\
\texttt{sc.park@yonsei.ac.kr}
}

\newtheorem{theorem}{Theorem}

\makeatletter
\newenvironment{numtheorem}[1]
  {\count@\c@theorem
   \global\c@theorem#1 %
    \global\advance\c@theorem\m@ne
   \theorem}
  {\endtheorem
   \global\c@theorem\count@}
\makeatother

\begin{document}

\maketitle

\begin{abstract}

We propose a novel framework based on neural network that reformulates classical mechanics as an operator learning problem. A machine directly maps a potential function to its corresponding trajectory in phase space without solving the Hamilton equations. Most notably, while conventional methods tend to accumulate errors over time through iterative time integration, our approach prevents error propagation. 
Two newly developed neural network architectures, namely VaRONet and MambONet, are introduced to adapt the Variational LSTM sequence-to-sequence model and leverage the Mamba model for efficient temporal dynamics processing. We tested our approach with various 1D physics problems: harmonic oscillation, double-well potentials, Morse potential, and other potential models outside the training data. Compared to traditional numerical methods based on the fourth-order Runge-Kutta (RK4) algorithm, our model demonstrates improved computational efficiency and accuracy. %

\vspace{0.25em}

Code is available at: \url{https://github.com/Axect/Neural_Hamilton}
\end{abstract}

\section{Introduction}

Hamilton's equations form the cornerstone of classical mechanics, providing a powerful framework for describing the evolution of physical systems.
These equations, which govern the dynamics of position and momentum in phase space, find applications across a wide spectrum of physical phenomena, from celestial mechanics to quantum field theory.
Despite their fundamental importance, solving Hamilton's equations for arbitrary potential functions remains a challenging task, often requiring sophisticated numerical methods.

Recent advancements in machine learning, particularly in the domain of operator learning, provide potential new approaches to tackle such challenges in physics.
Operator learning extends the concept of function approximation to mappings between function spaces, potentially allowing for more efficient and generalizable solutions to complex physical problems.
Our study presents a novel perspective by reformulating Hamilton's equations as an operator mapping problem.
We aim to discover an operator that directly maps bounded potential functions to the corresponding time-dependent position $q(t)$ and momentum $p(t)$ functions, without explicitly solving the differential equations.
This approach not only provides a new mathematical formulation of Hamilton's equations but also opens up new possibilities for efficient simulation and analysis of Hamiltonian systems.

A key challenge in applying operator learning to physical systems is the generation of suitable training data.
To address this, we introduce an innovative algorithm for generating diverse and physically plausible potential functions.
Our method combines Gaussian Random Fields with cubic B-Splines, ensuring the generation of smooth, bounded potentials that satisfy the necessary mathematical conditions for Hamilton's equations.
This approach significantly enhances the quality and relevance of the training dataset, a crucial factor in the performance of operator learning models.

Building upon existing operator learning architectures such as DeepONet \citep{lu2021learning} and TraONet \citep{kim2024hyperboliclr}, we introduce two novel network structures: VaRONet and MambONet.
VaRONet adapts the Variational LSTM sequence-to-sequence model \citep{hyun2024unsupervised} for operator learning, while MambONet leverages the recently introduced Mamba model \citep{gu2023mamba} to efficiently process the temporal dynamics of Hamiltonian systems.
These new architectures are designed to capture the complex relationships inherent in Hamiltonian dynamics more effectively than existing models.

To rigorously evaluate the performance of our approach, we conduct extensive tests on a range of potential functions commonly encountered in physics, including harmonic oscillators, double-well potentials, and Morse potentials.
Furthermore, we assess the extrapolation capabilities of our models by testing them on potential functions that lie outside the distribution of the training data.
This evaluation provides insights into the generalization power of our operator learning approach.

A critical aspect of our study is the comparison of our method with traditional numerical techniques for solving Hamilton's equations.
We analyze the computational efficiency, accuracy, and scalability of our approach relative to established numerical methods, highlighting the potential advantages of operator learning in specific scenarios.

Our work contributes to the growing intersection of machine learning and physics in several key ways:

\begin{itemize}
   \item We provide a rigorous mathematical formulation of Hamilton's equations as an operator learning problem, bridging the gap between classical mechanics and modern machine learning techniques.

   \item We introduce a novel algorithm for generating physically meaningful potential functions, addressing a crucial challenge in creating relevant datasets for operator learning in physics.

   \item We develop new neural network architectures specifically designed for learning operators in Hamiltonian systems, potentially offering improved performance over existing models.

   \item We offer comprehensive empirical evidence for the effectiveness of our approach through extensive testing on various potential functions and comparison with traditional numerical methods.
\end{itemize}

Our results demonstrate that the proposed neural network models, particularly MambONet, can effectively solve Hamilton's equations for a wide range of potential functions. Notably, our models outperform the traditional fourth-order Runge-Kutta (RK4) method on several physically relevant potentials in terms of accuracy, with MambONet showing particularly impressive results. In terms of computational efficiency, while not always faster, our models show comparable performance to RK4, with MambONet's computation time being similar to, albeit slightly slower than, RK4. We also observe significant performance improvements when training our models on larger datasets, highlighting the scalability of our approach.

This research not only advances the application of operator learning in physics but also opens new avenues for understanding and simulating complex physical systems.
By demonstrating the potential of machine learning techniques to capture fundamental physical principles, our work contributes to the ongoing dialogue between artificial intelligence and theoretical physics, potentially leading to new insights and more efficient computational methods in both fields.

\section{Related Work \& Background}

\subsection{Operator Learning}

The field of operator learning is built upon the theoretical foundations of neural networks, specifically the Universal Approximation Theorem \citep{cybenko1989approximation, hornik1989multilayer, hanin2019universal}.
This theorem establishes the capacity of neural networks to approximate continuous functions, providing the groundwork for more advanced concepts in machine learning and functional analysis.

Expanding on this foundation, research has shown that neural networks can approximate continuous functionals \citep{chen1993approximations} and nonlinear operators \citep{chen1995universal, back2002universal}.
These theoretical developments established the potential of neural networks in operator learning, although practical implementations remained challenging.

A significant breakthrough in practical operator learning came with the introduction of DeepONet \citep{lu2021learning}.
This architecture presents an efficient approach using two neural networks: a branch network processing discretized input functions at fixed nodes, and a trunk network handling desired output domain values.
DeepONet not only provided a practical framework but also contributed to the theory with a Generalized Universal Approximation Theorem for Operators:

\vspace{0.5em}

\begin{theorem}
   Suppose that $X$ is a Banach space, $K_1 \subset X, K_2 \subset \mathbb{R}^d$ are two compact sets in $X$ and $\mathbb{R}^d$, respectively, $V$ is a compact set in $C(K_1)$.
   Assume that $G: V \rightarrow C(K_2)$ is a nonlinear continuous operator. Then, for any $\epsilon > 0$, there exist positive integers $m, p$, continuous vector functions $\boldsymbol{g}: \mathbb{R}^m \rightarrow \mathbb{R}^p, \boldsymbol{f}: \mathbb{R}^d \rightarrow \mathbb{R}^p$, and $x_1, x_2, \cdots, x_m \in K_1$, such that
   \begin{equation}
   \left|G(u)(y) - \langle \underbrace{\boldsymbol{g}(u(x_1), u(x_2), \cdots, u(x_m))}_{\text{branch}}, \underbrace{\boldsymbol{f}(y)}_{\text{trunk}}\rangle \right| < \epsilon
   \end{equation}
   holds for all $u \in V$ and $y \in K_2$.
\end{theorem}

\vspace{0.5em}

This theorem extends the universal approximation capabilities of neural networks to operators, showing that the inner product of two neural networks (the branch and trunk networks) can approximate any nonlinear continuous operator to arbitrary precision. 
This theoretical foundation underpins the success of DeepONet and similar architectures in various applications of operator learning.

Following DeepONet's success, various neural operator architectures have emerged.
These include the Fourier Neural Operator \citep{li2020fourier}, which utilizes Fourier transforms for efficient computation, and the Deep Green Network \citep{gin2021deepgreen, boulle2022data}, focusing on direct approximation of Green's functions or kernels.
Recent developments have incorporated convolutional neural networks \citep{raonic2023convolutional} and transformer architectures \citep{hao2023gnot} into operator learning, further expanding the field's capabilities.

Alongside architectural innovations, theoretical understanding of operator learning has progressed significantly.
Research has focused on establishing error bounds \citep{kovachki2021universal, lanthaler2022error, de2022generic} and investigating generalization and training dynamics of neural operators \citep{lee2024training}.
These theoretical advances complement practical developments, offering a more comprehensive understanding of operator learning's potential and limitations.

The impact of operator learning extends beyond theoretical realms, finding applications in various scientific domains.
It has been applied to accelerate scientific simulations \citep{azizzadenesheli2024neural}, approximate wavefunction evolution in quantum mechanics \citep{mizera2023scattering}, and even in industrial applications for fault detection in dynamic processes \citep{rani2023fault}.

As operator learning continues to evolve, it promises to bridge the gap between traditional numerical methods and modern machine learning techniques.
This convergence offers new possibilities for solving complex problems in physics, engineering, and other scientific disciplines, potentially revolutionizing approaches to modeling and understanding complex systems.

\subsection{Hamilton's Equations}

In the early 19th century, the mathematical formulation of classical mechanics underwent a profound transformation with the introduction of what would become known as Hamilton's equations \citep{hamilton1833general}. These equations emerged as an elegant reformulation of Newton's laws of motion, offering not just an alternative perspective on classical dynamics, but a framework that would prove fundamental to developments in both classical and quantum physics.

The canonical form of Hamilton's equations is given by:
\begin{equation}
\begin{aligned}
   \dot{q}_i &= \frac{\partial H}{\partial p_i} \\
   \dot{p}_i &= - \frac{\partial H}{\partial q_i}
\end{aligned}
\end{equation}
Here, $H(q, p, t)$ typically represents the total energy of the system. These equations elegantly capture the system's dynamics in terms of partial derivatives of a single scalar function, the Hamiltonian.

Hamilton's equations offer several significant advantages, enhancing their utility across various areas of physics and mathematics.
In contrast to the second-order Euler-Lagrange equations, Hamilton's equations take the form of first-order differential equations.
This characteristic often simplifies numerical integration and analytical treatment, making the Hamiltonian approach particularly valuable for complex systems.

The Hamiltonian formalism introduces a fundamental symmetry between position and momentum variables, providing deeper insights into the structure of phase space \citep{arnol2013mathematical}.
This symmetry not only simplifies many calculations but also offers a more intuitive understanding of a system's dynamics.
Furthermore, the Hamiltonian framework provides a natural setting for canonical transformations, which can significantly simplify the analysis of complex systems by identifying conserved quantities \citep{landau1976mechanics}.

A key feature of Hamilton's formulation is its symplectic structure, which naturally preserves certain geometric properties of phase space.
This structure is intimately connected to Liouville's theorem, ensuring phase space volume conservation - a fundamental principle in statistical mechanics \citep{goldstein2002classical}.
The symplectic nature of Hamilton's equations has profound implications, particularly in the study of long-term system behavior and in the development of numerical integration schemes that preserve the system's underlying structure.

One of the most profound aspects of Hamilton's equations is their connection to quantum mechanics.
The Hamiltonian formalism provides a more direct route to quantum theory through the correspondence between Poisson brackets and quantum commutators \citep{dirac1981principles}.
This connection facilitates the study of semiclassical systems and quantum-classical correspondence, making it an invaluable tool in modern physics.

In recent years, Hamilton's equations have begun to influence artificial intelligence research, particularly in the field of neural networks for solving differential equations.
Hamiltonian Neural Networks (HNNs) \citep{greydanus2019hamiltonian} have emerged as a powerful method for solving differential equations in conservative systems.
Unlike traditional neural differential equations, HNNs approximate the Hamiltonian itself and use Hamilton's equations to update the system state, ensuring energy conservation.

Following the introduction of HNNs, various extensions and variants have been developed.
These include Symplectic ODE-Nets \citep{zhong2019symplectic}, which incorporate control theory, and Symplectic Recurrent Neural Networks \citep{chen2019symplectic}, which utilize recurrent architectures.
Research has also extended to non-conservative systems with the development of Dissipative Hamiltonian Neural Networks \citep{sosanya2022dissipative}.
These approaches continue to be actively studied and refined for solving differential equations \citep{chen2022learning}.

The influence of Hamiltonian mechanics in AI extends to discovering conservation laws.
AI Poincaré employs machine learning for symbolic discovery of conserved quantities \citep{liu2021machine}, while a data-driven ``neural deflation'' approach was introduced to numerically identify conservation laws \citep{zhu2023machine}.
These developments showcase AI's potential in uncovering fundamental principles of Hamiltonian systems.

\subsection{Hamilton's Equations as an Operator}

The formulation of Hamilton's equations as an operator provides a rigorous mathematical framework for understanding classical mechanical systems.
This perspective transforms our view of Hamilton's equations from a traditional initial value problem to a mapping between function spaces.

Consider a classical Hamiltonian of the form:
\begin{equation}
   H: \mathbb{R}^{2m} \rightarrow \mathbb{R}, \quad H(q, p) = \sum_{i=1}^m \frac{p_i^2}{2 m_i} + V(q)
\end{equation}
where $V: \mathbb{R}^m \rightarrow \mathbb{R}$ is the potential energy function. We assume that $H$ is $C^2$ in $(q, p)$. Given initial conditions $q(0)=q_0$ and $p(0)=p_0$, Hamilton's equations yield:
\begin{equation}
\begin{aligned}
   \dot{q}_i &= \frac{\partial H}{\partial p_i} = \frac{p_i}{m_i} \\
   \dot{p}_i &= -\frac{\partial H}{\partial q_i} = -\frac{\partial V}{\partial q_i}
\end{aligned}
\end{equation}
Let $x = (q, p) \in \mathbb{R}^{2m}$.
Then Hamilton's equations can be written in a more compact form:
\begin{equation}
\dot{x} = \mathbf{J} \nabla H(x), \quad \mathbf{J} = \begin{pmatrix} 
    \boldsymbol{0} & \boldsymbol{I_m} \\
    -\boldsymbol{I_m} & \boldsymbol{0}
\end{pmatrix}
\end{equation}
where $\boldsymbol{I_m}$ is the $m \times m$ identity matrix, and $\nabla H$ is the gradient of $H$ with respect to $x$.

Under our assumptions on $H$, the right-hand side of this equation is locally Lipschitz in $x$.
Therefore, by the fundamental theorem of ODEs (also known as the Picard-Lindelöf theorem) \citep{coddington1955theory}, we can state the following theorem:

\vspace{0.5em}

\begin{theorem}
   For any initial condition $x_0 \in \mathbb{R}^{2m}$, there exists a unique maximal solution $x(t) \in C^1(I, \mathbb{R}^{2m})$ to the equation
   \begin{equation}
   \dot{x} = \mathbf{J} \nabla H(x), \quad x(0) = x_0
   \end{equation}
   where $I \subset \mathbb{R}$ is the maximal interval of existence.
\end{theorem}

\vspace{0.5em}

This theorem forms the foundation for defining an operator that encapsulates the process of solving Hamilton's equations.
Let's define the following function spaces:
\begin{itemize}
   \item $\mathcal{H} = C^2(\mathbb{R}^{2m}, \mathbb{R})$: The space of twice continuously differentiable functions from $\mathbb{R}^{2m}$ to $\mathbb{R}$
   \item $\mathcal{T} = C^1(\mathbb{R}_{\geq 0}, \mathbb{R}^{2m})$: The space of continuously differentiable functions from non-negative real numbers to $\mathbb{R}^{2m}$
\end{itemize}
We can now define an operator $G$ that maps the Hamiltonian function $H(q,p)$ to the time-dependent trajectories $x(t) = (q(t), p(t))$:
\begin{equation}
\begin{aligned}
&G: \mathcal{H} \rightarrow \mathcal{T} \\
&G(H)(t) = x(t) = (q(t), p(t))
\end{aligned}
\end{equation}
where $x(t) = (q(t), p(t)) \in \mathcal{T}$ is the unique maximal solution to Hamilton's equations for the given Hamiltonian $H$.

To establish that $G$ is indeed an operator, we can express it as an integral equation:
\begin{equation}
   x(t) = x(0) + \int_0^t \mathbf{J} \nabla H(x(\tau)) \mathrm{d}\tau
\end{equation}
This integral formulation clearly demonstrates that $G$ is an operator, as it maps a function $H \in \mathcal{H}$ to a function $x \in \mathcal{T}$ through a well-defined mathematical operation.

In many physical applications, we are primarily interested in the effect of different potential functions on the system's behavior.
To address this, we introduce a mapping $\Phi$ from potentials to Hamiltonians. Let $\mathcal{V} = C^2(\mathbb{R}^m, \mathbb{R})$ be the space of twice continuously differentiable potential functions, and $\mathcal{H} = C^2(\mathbb{R}^{2m}, \mathbb{R})$ be the space of twice continuously differentiable Hamiltonian functions. We define $\Phi: \mathcal{V} \rightarrow \mathcal{H}$ as:
\begin{equation}
\Phi(V)(q, p) = \sum_{i=1}^m \frac{p_i^2}{2 m_i} + V(q)
\end{equation}
where $V \in \mathcal{V}$, $q = (q_1, \ldots, q_m) \in \mathbb{R}^m$ represents the generalized coordinates, $p = (p_1, \ldots, p_m) \in \mathbb{R}^m$ represents the generalized momenta, and $\{m_i\}_{i=1}^m$ are positive real numbers representing the masses associated with each degree of freedom.

This mapping allows us to construct a Hamiltonian from any given potential function.
To ensure the validity and uniqueness of this construction, we state the following theorem:

\vspace{0.5em}

\begin{theorem}
   The mapping $\Phi: \mathcal{V} \rightarrow \mathcal{H}$ defined above is well-defined and injective.
\end{theorem}

\vspace{0.5em}

This theorem ensures that we can construct a unique Hamiltonian from any given potential function in our space of interest.
The injectivity property guarantees that different potentials always lead to different Hamiltonians, preserving the one-to-one relationship between the potential energy landscape and the resulting dynamics.

Using this mapping, we can construct an operator $\tilde{G}: \mathcal{V} \rightarrow \mathcal{T}$ as follows:
\begin{equation}
\tilde{G} = G \circ \Phi
\end{equation}
The explicit form of $\tilde{G}$ would be:
\begin{equation}
\tilde{G}(V)(t) = G(\Phi(V))(t)
\end{equation}
This construction allows us to study how changes in the potential $V$ affect the system's trajectories, while leveraging the Hamiltonian formalism through $G$.

Proofs of Theorem 2 and Theorem 3 can be found in Appendix \ref{app:proof2} and \ref{app:proof3}, respectively.

By framing Hamilton's equations as an operator, we gain a powerful analytical tool that bridges classical mechanics and functional analysis.
This formulation emphasizes the direct dependence of the system's evolution on the Hamiltonian $H$ and the potential $V$, providing a clear connection between the energy function and the resulting trajectories in phase space.
This approach opens new avenues for both theoretical analysis and computational approaches in Hamiltonian dynamics, potentially leading to new insights into long-term behavior, stability, and chaos in these systems.

\section{Methods}

\subsection{Data Generation}

\subsubsection{Potential Function Generation}

In the context of functional and operator learning, generating a dataset that adequately covers the target input function space is paramount. While traditional approaches in DeepONet-based deep learning often employ Gaussian Random Fields (GRFs) to generate input functions at fixed nodes \citep{lu2021learning}, this method presents significant challenges when applied to our specific task of mapping potentials to trajectories in Hamiltonian systems.

The Universal Approximation Theorem for Operators stipulates that the operator should accept continuous functions defined on a compact subset $K_1$ of a Banach space $X$ as input, and produce continuous functions defined on a compact subset $K_2$ of Euclidean space $\mathbb{R}^d$ as output \citep{chen1995universal}.
In our application, where potential functions serve as inputs, $X$ becomes $\mathbb{R}^n$.
Since continuous functions map compact sets to compact sets, and compactness in Euclidean space implies boundedness and closure \citep{rudin1976principles}, both the domain of definition and range of our potential function must be bounded and closed.
These constraints extend to the output trajectory function $x(t)$ as well.

Consider a Hamiltonian system with initial conditions $q(0) = 0$ and $p(0) = 0$. A critical issue arises when $V'(0) > 0$. In this scenario, $\dot{p}(0) = -V'(0) < 0$, leading to $q(\delta t) < 0$ for a small time increment $\delta t$. This situation results in an ill-posed problem, as the potential function is undefined for negative $q$ values, violating the conditions necessary for the application of the Universal Approximation Theorem for Operators.

Consequently, arbitrary GRFs are unsuitable for potential function operator learning in Hamiltonian systems. Instead, we must consider a class of potential functions that bound trajectories within a specific range. In classical mechanics, such scenarios are known as bounded motions, and the corresponding potentials are termed bounded potential functions \citep{goldstein2002classical}.

To generate bounded motions with initial conditions $x(0) = (q(0), p(0)) = (0, 0)$, we define a differentiable potential $V$ such that $V(0) = V(L)$ and $V(q) < V(0)$ for all $q \in (0, L)$. This condition ensures that the solution $q(t)$ to Hamilton's equations remains bounded within $[0, L]$, satisfying the requirements of the Universal Approximation Theorem for Operators.

These bounded potential functions have significant practical applications in various domains of physics. Examples include the quadratic potential representing the harmonic oscillator, double-well potentials in quantum mechanics and quantum field theory, the Morse potential for diatomic molecule interactions, and effective potentials describing closed orbits of satellites or planets.

To address the challenges of generating suitable potential functions, we propose a novel algorithm that combines Gaussian Random Fields and Cubic B-Splines. This approach satisfies all the necessary conditions for generating suitable potential functions while maintaining the required smoothness and boundary conditions.

\begin{algorithm}[tbh]
\SetAlgoLined
\SetAlgoInsideSkip{5pt}
\DontPrintSemicolon
\SetKwInOut{Input}{Input}
\SetKwInOut{Output}{Output}
\caption{Generate random bounded potential}
\label{alg:potential}
Generate Random Bounded Potential$(L, V_0, n_{\text{GRF}}, \omega, p, l)$\;
\Input{$L$: domain length, $V_0$: potential scale, $n_{\text{GRF}}$: number of GRFs \\
       $\omega$: stability parameter, $p$: B-spline degree, $l$: length scale}
\Output{$V(q)$ for $q \in [0, L]$}
$y_i \gets i / (n_{\text{GRF}}+1)$ for $i = 1, \ldots, n_{\text{GRF}}$\;
$X = [X_1, X_2, \ldots, X_{n_{\text{GRF}}}] \sim \mathcal{G}(0, k(y_i, y_j; l))$ \tcp*{Sample GRF vector}
$X_i \gets V_0 - 2V_0 \cdot \frac{X_i - \min_j X_j}{\max_j X_j - \min_j X_j}$ for $i = 1, \ldots, n_{\text{GRF}}$ \tcp*{Normalize $X$}
$q_i = \begin{cases}
    0 & \text{if } i = 0 \\
    \sim \mathcal{U}(\omega, 1/n_{\text{GRF}} - \omega/2)L & \text{if } i = 1 \\
    \sim \mathcal{U}\left(\frac{i-1}{n_{\text{GRF}}} + \frac{\omega}{2}, \frac{i}{n_{\text{GRF}}} - \frac{\omega}{2}\right)L & \text{if } 2 \leq i \leq n_{\text{GRF}}-1 \\
    \sim \mathcal{U}((n_{\text{GRF}}-1)/n_{\text{GRF}} + \omega/2, 1-\omega)L & \text{if } i = n_{\text{GRF}} \\
    L & \text{if } i = n_{\text{GRF}}+1
\end{cases}$ \tcp*{Define $q_i$}
$V_i = \begin{cases}
    V_0 & \text{if } i = 0 \text{ or } i = n_{\text{GRF}}+1 \\
    X_i & \text{if } 1 \leq i \leq n_{\text{GRF}}
\end{cases}$ \tcp*{Define $V_i$}
$\mathbf{P}_i \gets [q_i, V_i]^T$ for $i = 0, \ldots, n_{\text{GRF}}+1$ \tcp*{Control points}
$C \gets n_{\text{GRF}} + 2$ \tcp*{Number of control points}
$K \gets C - p + 1$ \tcp*{Number of distinct knots}
$u_i = \begin{cases}
    0 & \text{if } 0 \leq i < p \\
    \frac{i-p}{K-1} & \text{if } p \leq i \leq p + K \\
    1 & \text{if } p + K < i \leq K + 2 p
\end{cases}$ \tcp*{Knot vector for clamped spline}
$\displaystyle
    [q(\lambda), V(q(\lambda))]^T \gets \sum_{i=0}^{n_{\text{GRF}}+1} B_{i,p}(\lambda; \mathbf{u}) \mathbf{P}_i \text{ for }\lambda \in [0, 1]
$ \tcp*{Construct B-spline}
\KwRet{$V(q)$ for $q \in [0, L]$}
\end{algorithm}

Our algorithm ensures several key properties:

\begin{enumerate}
    \item It generates diverse and smooth potential functions $V(q)$ defined on a compact domain, satisfying the mathematical conditions required by the Universal Approximation Theorem for Operators while maintaining consistent boundary conditions.
    \item The use of clamped cubic B-splines ensures that the resulting $V(q)$ is a $C^2$ function \citep{bartels1995introduction}, providing the necessary smoothness for our Hamilton's equations model.
    \item The combination of Gaussian Random Fields and B-splines allows for a wide exploration of the potential function space while preserving physical constraints, enhancing the generalization capability of the learned operator.
    \item The method is computationally efficient and easily implementable, facilitating the generation of large, diverse datasets crucial for training deep learning models to solve Hamilton's equations effectively.
\end{enumerate}

By employing this algorithm, we can generate a comprehensive dataset of potential functions and their corresponding trajectories, which is crucial for effectively training our operator network to solve Hamilton's equations. This approach bridges the gap between the theoretical requirements of operator learning and the practical needs of generating physically meaningful potentials in Hamiltonian systems.

\subsubsection{Trajectory Generation}

To implement supervised learning for operator learning, we need to provide output function information as labels for each input potential function during training.
However, since we cannot practically provide functions in infinite-dimensional space, we represent these functions by discretizing a defined time interval into $m$ uniformly spaced nodes and recording function values at these nodes.

The generation of trajectory data is crucial for training our operator learning models.
We employ a two-step process: first, we solve the Hamilton's equations numerically using an ordinary differential equation (ODE) solver, and then we interpolate the solution using a cubic Hermite spline.
This approach offers several advantages over directly sampling the ODE solver output.

Our method allows for flexibility in node spacing, as the ODE solver can use a fine time step for accuracy while we independently choose the number of nodes for our trajectory representation.
This approach ensures numerical stability by avoiding large time steps in the ODE solver that could lead to significant errors.
Simultaneously, it maintains data efficiency by preventing unnecessarily large datasets that would increase computational costs during learning. 

Furthermore, this decoupling of the ODE solution from the final trajectory representation enables the use of adaptive ODE solvers for increased accuracy where needed.
Importantly, the use of cubic Hermite splines provides $C^1$ continuity, aligning with the requirements of our operator formulation of Hamilton's equations, which stipulates that trajectory functions should be $C^1$.

For simplicity and without loss of generality, we set the mass $m=1$ in our Hamilton's equations. This simplification allows us to focus on the potential function's effect on the system dynamics.

\begin{algorithm}[tbh]
\SetAlgoLined
\SetAlgoInsideSkip{5pt}
\DontPrintSemicolon
\SetKwInOut{Input}{Input}
\SetKwInOut{Output}{Output}
\caption{Generate label for each potential function}
\label{alg:trajectory}
\Input{$m$: number of sensors, $V$: potential function, $T$: total time}
\Output{$\mathbf{x}(t)=[q(t), p(t)]$ for $t \in \{0, \frac{1}{m-1}T, \ldots, \frac{m-2}{m-1}T, T\}$}
Define Hamilton's equations: $\dot{q} = p, \quad \dot{p} = -\frac{\partial V}{\partial q}$ \tcp*{ODE system}
$[\mathbf{t}, \mathbf{q}, \mathbf{p}] \gets \text{ODESolver}(\dot{q}, \dot{p}, [0, T], [0, 0])$ \tcp*{Solve ODE}
$q(t) \gets \text{CubicHermiteSpline}(\mathbf{t}, \mathbf{q})$ \tcp*{Interpolate position}
$p(t) \gets \text{CubicHermiteSpline}(\mathbf{t}, \mathbf{p})$ \tcp*{Interpolate momentum}
$\mathbf{x}(t) \gets [q(t), p(t)]$ for $t \in \{0, \frac{1}{m-1}T, \ldots, \frac{m-2}{m-1}T, T\}$ \tcp*{Evaluates at sensors}
\KwRet{$\mathbf{x}(t)$}
\end{algorithm}

In the algorithm \ref{alg:trajectory}, we first compute the gradient of the potential function, which is required for solving Hamilton's equations.
We then use a numerical ODE solver to obtain a solution to these equations.
The choice of ODE solver can be adapted based on the specific requirements of the system; for instance, we might use a 4th order Runge-Kutta method or an adaptive solver for more complex potentials.

The solution from the ODE solver is then interpolated using cubic Hermite splines.
This interpolation step is crucial as it allows us to generate smooth, continuous representations of the trajectories that satisfy the $C^1$ continuity requirement.

\subsection{Model Architectures}

\subsubsection{DeepONet}

The DeepONet architecture \citep{lu2021learning} forms the foundation of our operator learning approach for solving Hamilton's equations.
This architecture consists of two key components: a branch network and a trunk network, designed to efficiently learn mappings between function spaces.

\begin{figure}
    \begin{adjustwidth}{-0.5cm}{-0.5cm}
    \centering
    \includegraphics[width=1.05\textwidth]{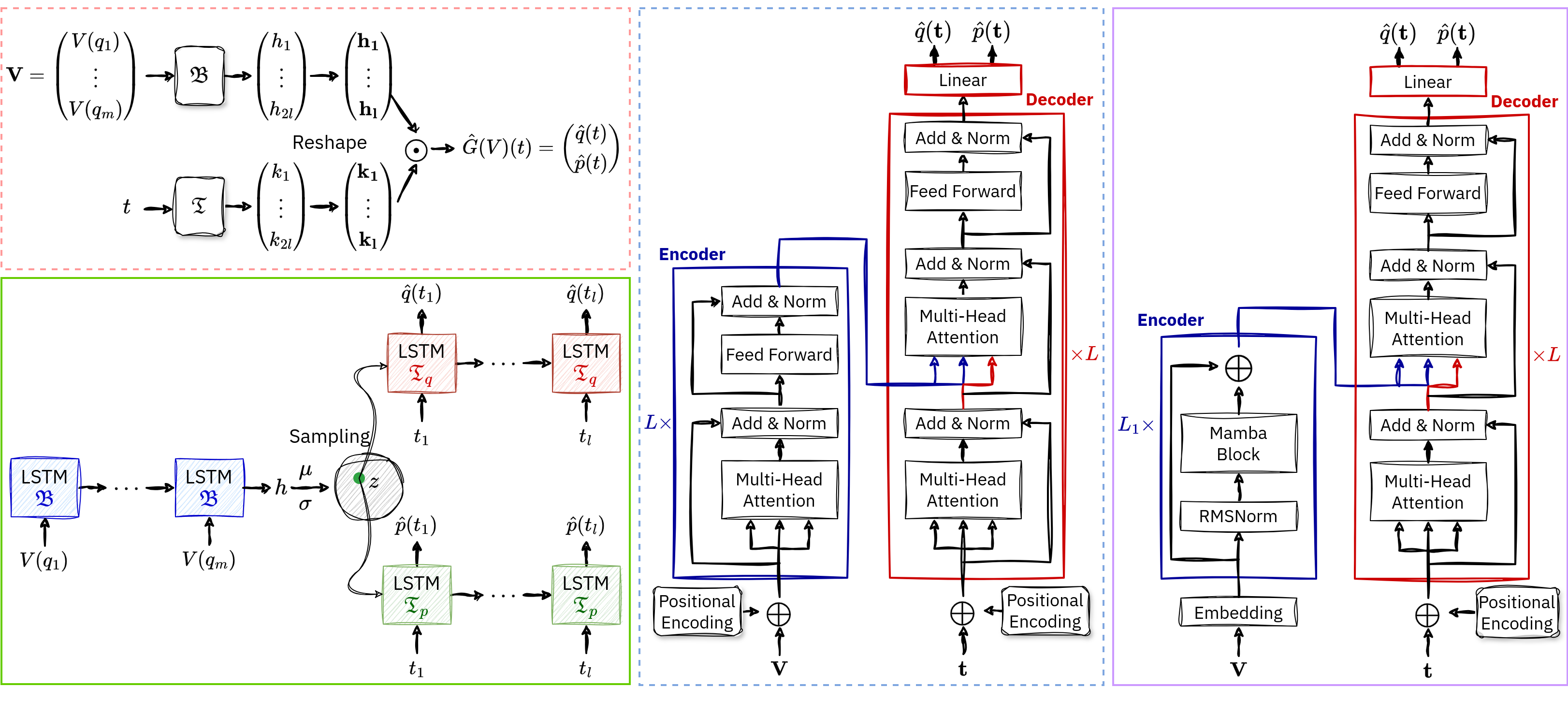}
    \caption{\textcolor{red}{(Red dashed box)} Architectures of DeepONet,
    \textcolor{olive}{(Green solid box)} VaRONet,
    \textcolor{cyan}{(Blue dashed box)} TraONet, and
    \textcolor{violet}{(Purple solid box)} MambONet.
    Each model processes the input potential function \textbf{V} and time \textbf{t} to predict the position $\hat{q}(\textbf{t})$ and momentum $\hat{p}(\textbf{t})$ trajectories.
    DeepONet uses simple feed-forward networks, VaRONet incorporates variational sampling and LSTM layers, TraONet utilizes transformer blocks, and MambONet combines Mamba blocks with transformer decoders.}
    \label{fig:model_architectures}
    \end{adjustwidth}
\end{figure}

In our implementation, the branch network $\mathfrak{B}$ takes the potential function $V(q)$ as input.
To accommodate the infinite-dimensional nature of the input function space, we discretize $V(q)$ using $m$ fixed sensor points uniformly distributed over the domain $[0, 1]$.
This discretization yields an input vector $\mathbf{V} \in \mathbb{R}^m$ for the branch network.

The trunk network $\mathfrak{T}$ processes the query points at which we wish to evaluate the output trajectories.
In our case, these query points represent time values.
While the trunk network can theoretically handle a variable number of query points, we opt for consistency and computational efficiency by using the same number of fixed time points as the potential function discretization, resulting in an input vector $\mathbf{t} \in \mathbb{R}^m$.

Both networks output vectors in $\mathbb{R}^{2l}$, where $l$ is a hyperparameter determining the width of the network outputs.
These outputs are then reshaped into $\mathbb{R}^{l \times 2}$ matrices using a reshape operator $\mathcal{R}$.
The reshaped outputs are combined through a column-wise inner product operation, denoted by $\odot$, to produce the final predictions for position $\hat{q}(\mathbf{t})$ and momentum $\hat{p}(\mathbf{t})$.

The complete DeepONet model for our Hamilton's equations solver can be expressed as:
\begin{equation}
    \hat{G}(V)(\mathbf{t}) = [\hat{q}(\mathbf{t}), \hat{p}(\mathbf{t})] = \mathcal{R}(\mathfrak{B}(\mathbf{V})) \odot \mathcal{R}(\mathfrak{T}(\mathbf{t})) + \mathfrak{b}
    \label{eq:deeponet}
\end{equation}
where $\mathfrak{b} \in \mathbb{R}^{m \times 2}$ is a learnable bias vector that has been shown to enhance performance \citep{lu2021learning}.

The loss function for DeepONet is defined as the mean squared error between the predicted and true trajectories:

\begin{equation}
    \mathcal{L} = \frac{1}{|\mathcal{B}|} \sum_{\mathbf{V} \in \mathcal{B}} \|\hat{G}(V)(\mathbf{t}) - G(V)(\mathbf{t})\|^2
    \label{loss:deeponet}
\end{equation}

where $\mathcal{B}$ represents a batch of potential functions, $\hat{G}(V)(\mathbf{t})$ is the DeepONet's prediction, and $G(V)(\mathbf{t})$ is the true solution of Hamilton's equations.

\subsubsection{VaRONet}

Building upon the DeepONet architecture and inspired by recent advancements in sequence-to-sequence learning, we propose the {\it Variational Recurrent Operator Network} (VaRONet). This novel architecture leverages the sequential nature of our output data—the time-dependent position and momentum functions—while incorporating principles from variational autoencoders and recurrent neural networks.

The VaRONet architecture, illustrated in green solid box of Figure \ref{fig:model_architectures}, draws structural inspiration from both DeepONet and sequence-to-sequence models. While not mathematically equivalent, these architectures share important similarities that make VaRONet particularly suitable for our task of solving Hamilton's equations.

In DeepONet, the branch network processes the input function and outputs a fixed-dimensional vector, which is then combined with the output of the trunk network through an inner product operation. Similarly, in sequence-to-sequence models, an encoder compresses the input sequence into a fixed-dimensional vector, which is then used to initialize the hidden state of the decoder. VaRONet adapts these concepts to the context of Hamiltonian systems, leveraging the time series characteristics of the position and momentum functions.

The encoder $\mathfrak{B}$ (analogous to the branch network) processes the discretized potential function $\mathbf{V}$, compressing it into a latent representation:

\begin{equation}
    h = \mathfrak{B}(\mathbf{V})
\end{equation}

We then employ a variational approach, using two separate networks, $\mathfrak{F}_\mu$ and $\mathfrak{F}_\sigma$, to transform the encoder output into parameters $\mu$ and $\sigma$ of a multivariate Gaussian distribution:

\begin{equation}
    \mu = \mathfrak{F}_\mu(h), \quad \sigma = \exp(\frac{1}{2}\mathfrak{F}_\sigma(h))
\end{equation}

A vector $z$ is sampled from this distribution using the reparameterization trick \citep{kingma2013auto}:

\begin{equation}
    z = \mu + \sigma \odot \epsilon, \quad \epsilon \sim \mathcal{N}(0, I)
\end{equation}

This sampled vector $z$ is then processed to produce initial hidden states for the position and momentum decoders:

\begin{equation}
    h_q = \mathfrak{F}_q(z), \quad h_p = \mathfrak{F}_p(z)
\end{equation}

The decoders $\mathfrak{T}_q$ and $\mathfrak{T}_p$ are recurrent neural networks (specifically, LSTMs) that use these initial hidden states and process the time inputs $\mathbf{t}$ to generate the position and momentum trajectories:

\begin{equation}
    \hat{G}(V)(\mathbf{t}) = [\hat{q}(\mathbf{t}), \hat{p}(\mathbf{t})]; \quad \hat{q}(\mathbf{t}) = \mathfrak{T}_q(\mathbf{t}; h_q), \quad \hat{p}(\mathbf{t}) = \mathfrak{T}_p(\mathbf{t}; h_p)
\end{equation}

This structure is analogous to the inner product operation in DeepONet. While in DeepONet the output of the branch network is combined with the output of the trunk network through an inner product, in VaRONet, the encoded information from the potential function (represented by $h_q$ and $h_p$) is combined with the time input $\mathbf{t}$ through the gating mechanisms of the LSTM cells. These gating operations can be seen as a more complex, nonlinear analogue of the inner product in DeepONet.

The use of variational methods and dual decoders, inspired by the VAE-LSTM Seq2Seq model previously employed for unsupervised sequence-to-sequence learning in biomedical signal processing \citep{hyun2024unsupervised}, enables the model to specialize in predicting position and momentum separately while introducing a stochastic element that helps capture the inherent variability of complex physical systems.

The loss function for VaRONet combines a reconstruction term with a Kullback-Leibler divergence term, typical of variational autoencoders:

\begin{equation}
    \mathcal{L} = \frac{1}{|\mathcal{B}|} \sum_{\mathbf{V} \in \mathcal{B}} \left[\|\hat{G}(V)(\mathbf{t}) - G(V)(\mathbf{t})\|^2 + \beta D_\text{KL}(\mathcal{N}(\boldsymbol{\mu}, \boldsymbol{\sigma}^2\mathbf{I}) \| \mathcal{N}(\mathbf{0}, \mathbf{I}))\right]
\end{equation}

where $\mathcal{B}$ represents a batch of potential functions, $G(V)(\mathbf{t})$ is the true solution of Hamilton's equations, and $\beta$ is a hyperparameter controlling the weight of the KL divergence term, as used in $\beta$-VAE models \citep{higgins2017beta}.

\subsubsection{TraONet}

The Transformer Operator Network (TraONet) leverages the power of self-attention mechanisms to solve Hamilton's equations.
This architecture, initially explored in the context of learning rate scheduling \citep{kim2024hyperboliclr}, has been adapted and optimized for the task of solving Hamilton's equations.

As shown in the blue dashed box of Figure \ref{fig:model_architectures}, TraONet employs a transformer encoder as the branch network and a transformer decoder as the trunk network.
This design aims to effectively process the spatial information of potential functions and the temporal aspects of trajectories simultaneously.

The operation of TraONet begins with the embedding of the discretized potential function.
Each potential value $V(q_i)$ is first embedded into a $d$-dimensional space using a learnable linear transformation $\mathcal{E}_{1 \to d}$:
\begin{equation}
V_\mathcal{E}(q_i) = \mathcal{E}_{1 \to d}(V(q_i)) \in \mathbb{R}^d \quad \text{for } i = 1, 2, \dots, m
\end{equation}
To incorporate spatial information, we apply positional encoding to the embedded potentials:
\begin{equation}
V_\mathcal{P}(q_i) = V_\mathcal{E}(q_i) + \mathcal{P}_i \quad \text{for } i = 1, 2, \dots, m
\end{equation}
where the positional encoding $\mathcal{P}_i$ is defined as:
\begin{equation}
(\mathcal{P}_i)_{2j-1} = \sin(i / 10000^{(2(j-1)) / d}), \quad (\mathcal{P}_i)_{2j} = \cos(i / 10000^{(2(j-1)) / d})
\end{equation}
The branch network $\mathfrak{B}$, implemented as a transformer encoder, processes the sequence of positionally encoded potentials $\mathbf{V}_\mathcal{P} = [V_\mathcal{P}(q_1)^T, \dots, V_\mathcal{P}(q_m)^T] \in \mathbb{R}^{m \times d}$ to produce a memory vector $\mathbf{h} \in \mathbb{R}^{m \times d}$:
\begin{equation}
\mathbf{h} = \mathfrak{B}(\mathbf{V}_\mathcal{P})
\end{equation}
Similarly, we embed and positionally encode the time points:
\begin{equation}
(\mathbf{t}_\mathcal{P})_i = \mathcal{E}_{1 \to d}(t_i) + \mathcal{P}_i \quad \text{for } i = 1, 2, \dots, m
\end{equation}
The trunk network $\mathfrak{T}$, implemented as a transformer decoder, then processes the positionally encoded time points $\mathbf{t}_\mathcal{P}$ and the memory vector $\mathbf{h}$ to predict the trajectories:
\begin{equation}
\hat{G}(V)(\mathbf{t}) = [\hat{q}(\mathbf{t}), \hat{p}(\mathbf{t})] = \mathfrak{T}(\mathbf{t}_\mathcal{P}; \mathbf{h})
\end{equation}
Unlike traditional transformer decoders, TraONet does not employ masking in the self-attention mechanism of the decoder. This design choice reflects our treatment of trajectories as functions of time rather than sequentially evolving entities, allowing the model to capture global temporal dependencies.

The loss function for TraONet is identical to that of DeepONet, as defined in Equation \ref{loss:deeponet}.

\subsubsection{MambONet}

Building on recent advancements in sequence modeling, we introduce the Mamba Operator Network (MambONet), a hybrid architecture that combines the efficiency of the Mamba model \citep{gu2023mamba} with the proven effectiveness of transformers.
MambONet employs Mamba blocks for the branch network and a transformer decoder for the trunk network, as illustrated in the purple solid box of Figure \ref{fig:model_architectures}.

The Mamba model, which has gained attention as a potential alternative to transformers, utilizes a \textit{Selective State Space} approach.
This method allows the model to focus on important information without relying on attention mechanisms, effectively addressing the quadratic complexity issue associated with traditional attention-based models. By incorporating Mamba, MambONet aims to capture long-range dependencies more efficiently, particularly in processing the input potential functions.

The operation of MambONet proceeds as follows. First, each discretized potential value $V(q_i)$ is embedded into a $d$-dimensional space:
\begin{equation}
V_\mathcal{E}(q_i) = \mathcal{E}_{1 \to d}(V(q_i)) \in \mathbb{R}^d \quad \text{for } i = 1, 2, \dots, m
\end{equation}
where $\mathcal{E}_{1 \to d}$ is a learnable linear transformation.

The sequence of embedded potentials $\mathbf{V}_\mathcal{E} = [V_\mathcal{E}(q_1)^T, \dots, V_\mathcal{E}(q_m)^T] \in \mathbb{R}^{m \times d}$ is then processed by multiple Mamba blocks with residual connections, serving as the branch network $\mathfrak{B}$:
\begin{equation}
\mathbf{h} = \mathfrak{B}(\mathbf{V}_\mathcal{E})
\end{equation}
where $\mathbf{h} \in \mathbb{R}^{m \times d}$ is the encoded representation of the potential function.

The trunk network $\mathfrak{T}$, implemented as a transformer decoder, then processes the positionally encoded time points $\mathbf{t}_\mathcal{P}$ (as in TraONet) and the encoded potential $\mathbf{h}$ to predict the trajectories:
\begin{equation}
\hat{G}(V)(\mathbf{t}) = [\hat{q}(\mathbf{t}), \hat{p}(\mathbf{t})] = \mathfrak{T}(\mathbf{t}_\mathcal{P}; \mathbf{h})
\end{equation}
The loss function for MambONet is identical to that of DeepONet, as defined in Equation \ref{loss:deeponet}.

\section{Experiments}

\subsection{Data Preparation}

To effectively train and evaluate our operator learning models for solving Hamilton's equations, we generated comprehensive datasets of potential functions and their corresponding trajectories. We employed Algorithm \ref{alg:potential} to create a diverse set of potential functions, adopting parameters that balance computational efficiency with physical relevance.

For consistency and simplicity, we set the domain length $L$ to 1 and the potential scale $V_0$ to 2. To ensure diversity in the generated potentials, we randomly selected the number of Gaussian Random Fields ($n_\text{GRF}$) from integers between 1 and 7, using a uniform distribution. Additionally, we introduced variability in the kernel window length ($l$) of the GRF, selecting values uniformly at random from the range 0.01 to 0.2. To mitigate the occurrence of overly stiff potentials, we fixed the stability parameter $\omega$ at 0.05.

We generated two distinct datasets: a {\it standard dataset} comprising 10,000 potential functions and an {\it extended dataset} with 100,000 potentials. These datasets, henceforth referred to as the ``baseline'' and ``expanded'' datasets respectively, provide a robust foundation for training and testing our models across various scales of data availability.

For each potential function in both datasets, we applied Algorithm \ref{alg:trajectory} to generate the corresponding trajectory labels. We set the number of sensors ($m$) to 100 and the total simulation time ($T$) to 1, striking a balance between resolution and computational efficiency.

To adapt the potential functions for use in DeepONet-style operator learning, we discretized each potential using a fixed set of sensors. Specifically, we uniformly divided the domain $q=[0, 1]$ into 100 nodes and sampled the potential values at these points. This discretization results in input potential functions with dimensions $N \times 100$, where $N$ is 10,000 for the standard dataset and 100,000 for the extended dataset.

Correspondingly, the input time dimension is also $N \times 100$, with each row representing the same set of 100 evenly spaced time points between 0 and 2. The target trajectories $q(t)$ and $p(t)$ each have dimensions $N \times 100$, aligning with the input time points.

To ensure robust evaluation of our models, we employed a hold-out methodology. For each dataset, we randomly allocated 80\% of the samples for training and reserved the remaining 20\% for validation. This split allows us to assess the generalization capability of our models on unseen data while maintaining a substantial training set.

\begin{figure}
    \centering
    \begin{subfigure}[b]{0.32\textwidth}
        \centering
        \includegraphics[width=\textwidth]{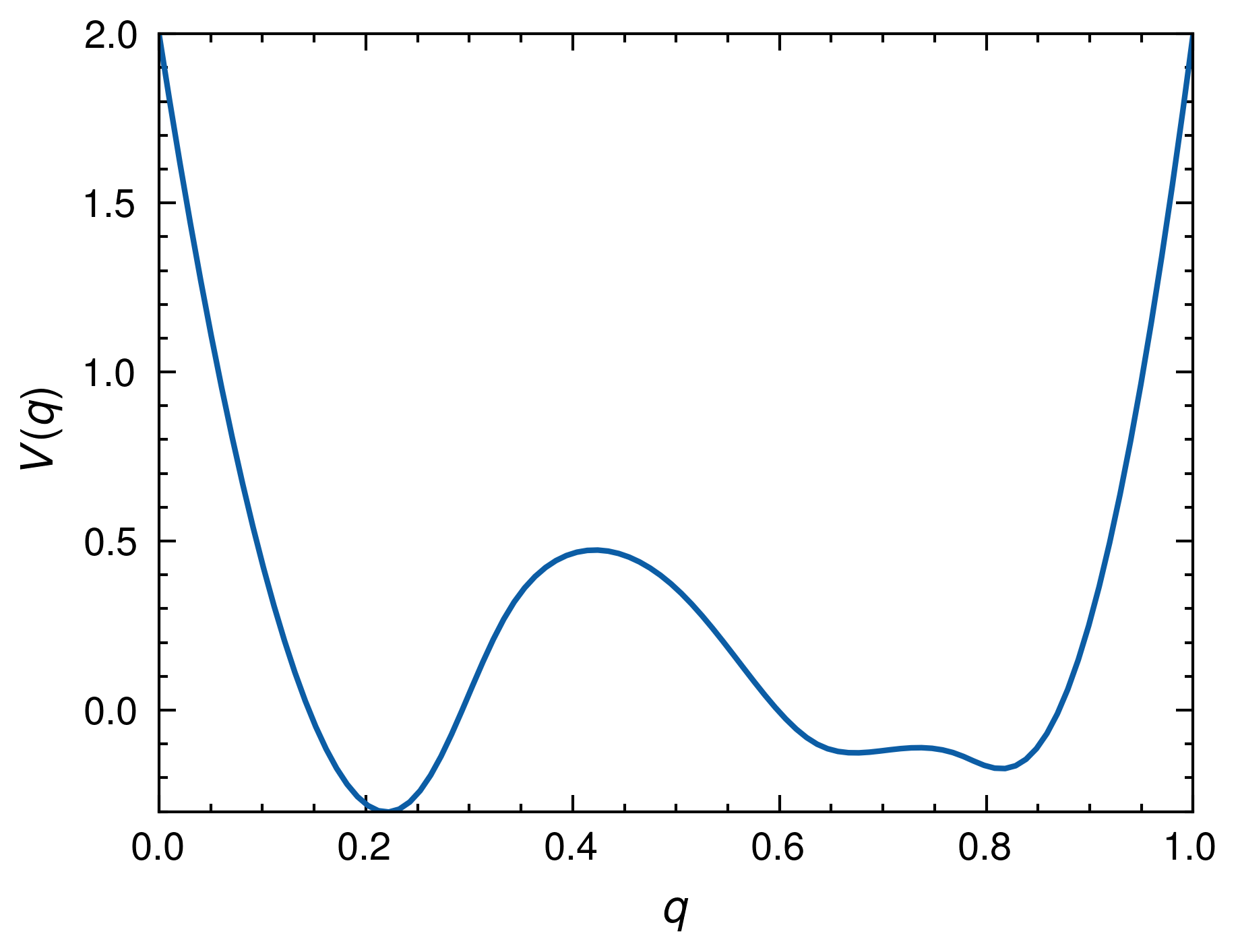}
        \caption{Potential function}
    \end{subfigure}
    \hfill
    \begin{subfigure}[b]{0.32\textwidth}
        \centering
        \includegraphics[width=\textwidth]{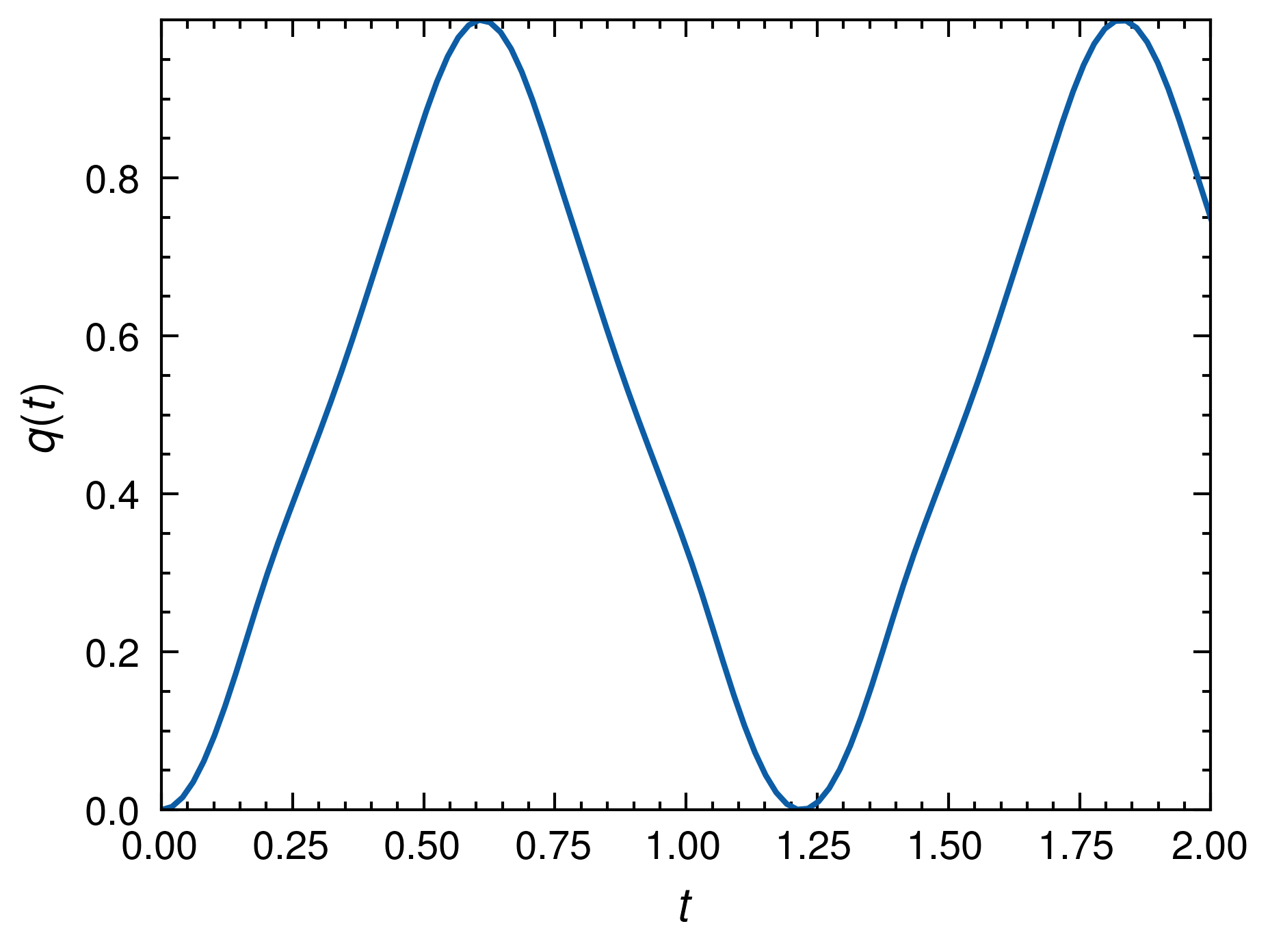}
        \caption{Position $q(t)$}
    \end{subfigure}
    \hfill
    \begin{subfigure}[b]{0.32\textwidth}
        \centering
        \includegraphics[width=\textwidth]{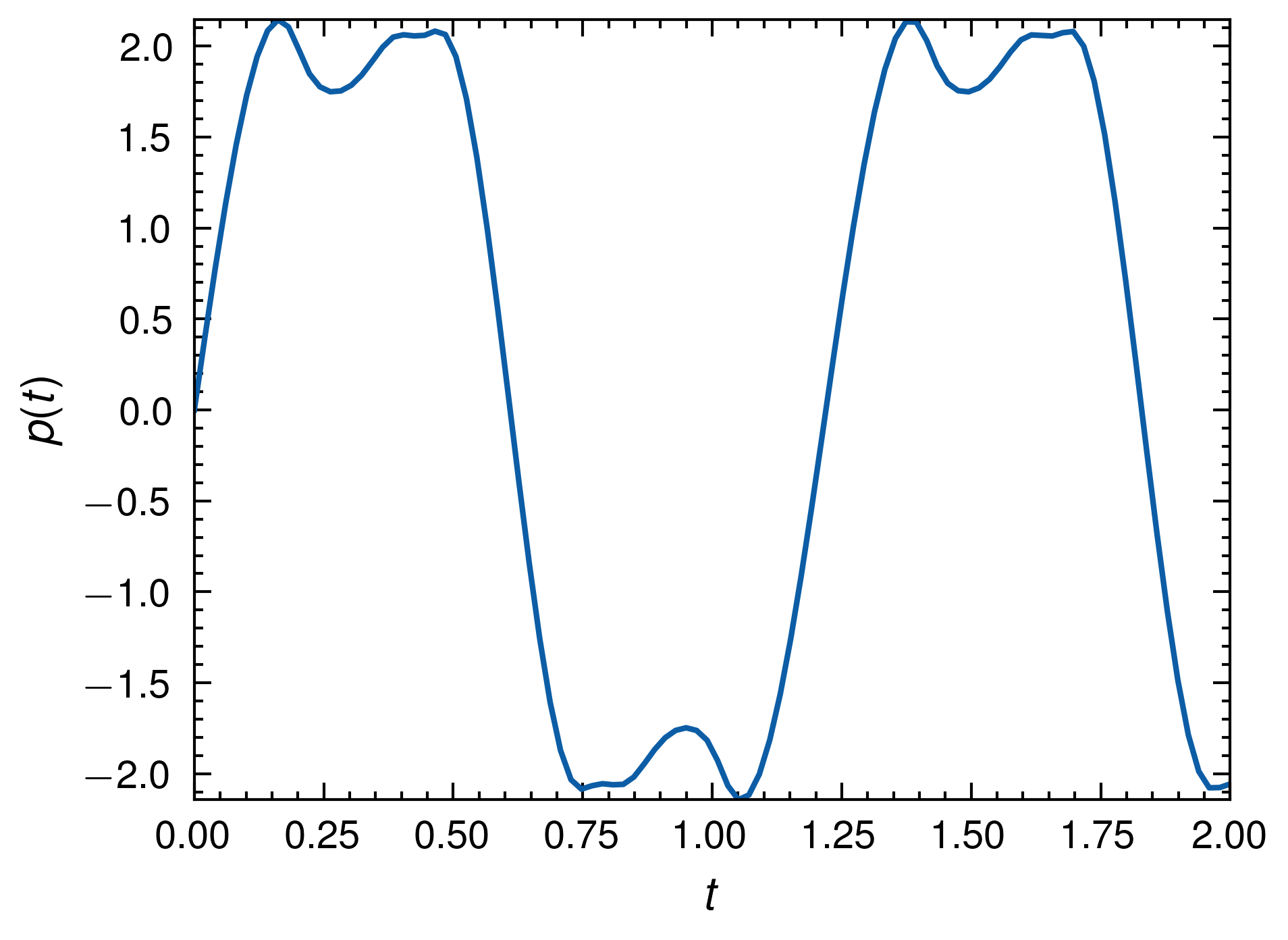}
        \caption{Momentum $p(t)$}
    \end{subfigure}
    \caption{Example of a generated potential function and corresponding trajectories. (a) shows the potential function $V(q)$ generated using Algorithm \ref{alg:potential}. (b) and (c) display the corresponding position $q(t)$ and momentum $p(t)$ trajectories generated from the potential function using Algorithm \ref{alg:trajectory}.}
    \label{fig:potential_and_trajectories}
\end{figure}

Figure \ref{fig:potential_and_trajectories} illustrates an example of a generated potential function and its corresponding position and momentum trajectories. This visualization demonstrates the relationship between the potential shape and the resulting dynamics of the system.

\subsection{Test Set Generation}

To rigorously evaluate our models' generalization capabilities, we generated a separate test set consisting of 4,000 potential functions. This test set was created using the same algorithm described in Algorithm \ref{alg:potential}, but with a different random seed (42) than the one used for generating the training and validation sets (8407). This approach ensures that while the test potentials are drawn from the same distribution as the training data, they represent truly unseen examples for our models.

\subsection{Model Training and Hyperparameter Optimization}

To ensure consistency and optimize the training process across all models, we implemented a standardized training protocol. This approach facilitated fair comparisons between different architectures and allowed for efficient hyperparameter tuning.

For all models, we employed the AdamW optimizer with beta values of (0.85, 0.98), chosen for its effectiveness in handling weight decay and its ability to adapt to different parameter scales \citep{loshchilov2017decoupled}. We used a batch size of 100, balancing computational efficiency with stable gradient estimates. The learning rate was managed using the ExpHyperbolicLR scheduler, a choice motivated by its ability to rapidly optimize parameters in early epochs while maintaining performance stability in later stages of training \citep{kim2024hyperboliclr}.

To account for the stochastic nature of neural network training and ensure robust performance estimates, we conducted all optimization and evaluation processes using five different random seeds: 89, 231, 928, 814, and 269. These seeds were randomly selected to provide a diverse set of initial conditions. Performance metrics reported in our results represent the average across these five runs, offering a more reliable assessment of each model's capabilities.

Our hyperparameter optimization strategy followed a two-stage process. Initially, we fixed the scheduler parameters and performed a grid search over model-specific parameters, optimizing for performance at 50 epochs. This step allowed us to identify promising architectural configurations for each model type. Subsequently, using the optimized model parameters, we fine-tuned the scheduler parameters using the {\it Tree-Structured Parzen Estimator} (TPE) method \citep{bergstra2011algorithms, watanabe2023tree}, again targeting performance at the 50-epoch mark. This approach combines the breadth of grid search for model architecture with the efficiency of Bayesian optimization for learning rate scheduling.
The specific ranges used for hyperparameter optimization and the resulting optimal configurations for each model are detailed in Appendix \ref{app:hyperparams}.

For the standard dataset, we conducted our final training runs using the best-performing configurations identified through our hyperparameter optimization process. Each model was trained for 250 epochs, allowing for thorough convergence and providing insights into long-term learning dynamics.

For the extended dataset, we utilized the same best configurations as determined for the standard dataset. However, to leverage the increased data volume effectively, we applied additional fine-tuning techniques specifically for the Hyperbolic-based learning rate schedulers (HyperbolicLR and ExpHyperbolicLR). These techniques allowed us to adapt the learning rate dynamics to the larger dataset, potentially leading to improved performance. The details of these fine-tuning techniques and their impact on training are provided in Appendix \ref{app:hyperparams}. As with the standard dataset, models trained on the extended dataset were also run for 250 epochs.

\subsection{Evaluation on Physically Relevant Potentials}

While the primary performance comparison of our trained models was based on test loss, we conducted additional tests using potential functions commonly encountered in physics. This approach allows us to assess the models' performance on problems of practical significance and evaluate their generalization capabilities beyond the training distribution.

We selected four distinct potential functions for this evaluation, as summarized in Table \ref{tab:test_potentials}.

\begin{table}
\caption{Test Potentials for Model Evaluation. For the Morse potential, $D_e = 8/(\sqrt{5} - 1)^2$ and $a = 3 \ln ((1+\sqrt{5})/2)$. For the Softened Mirrored Free Fall potential, $\alpha = 20$.}
\label{tab:test_potentials}
\centering
\renewcommand{\arraystretch}{1.25}
\begin{tabular}{lll}
    \toprule
    Potential Type & Function $V(q)$ & Notes \\
    \midrule
    Simple Harmonic Oscillator & $8(q - 0.5)^2$ & Analytic solution available \\
    Double Well & $\frac{625}{8}(q - 0.2)^2(q - 0.8)^2$ & Common in quantum physics \\
    Morse & $D_e (1 - e^{-a (q-1/3)})^2$ & Models molecular bonds \\
    Mirrored Free Fall & $4 |q - 0.5|$ & Test for extrapolation capabilities \\
    Softened Mirrored Free Fall & $\frac{4}{\operatorname{coth}(\alpha / 2)}(q-\frac{1}{2}) \operatorname{coth}(\alpha(q-\frac{1}{2}))$ & Compare to mirrored free fall \\
    \bottomrule
\end{tabular}
\end{table}

The mirrored free fall potential deserves special mention.
It is derived from the free fall potential $V(q) \sim q$, modified to create a bounded motion by mirroring and shifting: $V(q) = 4|q - \frac{1}{2}|$ (4 for consistent scale).
Notably, this potential satisfies $C^0$ continuity but not $C^2$, placing it outside our training data distribution and serving as a test for extrapolation capabilities.
For the Softened Mirrored Free Fall potential, we use the hyperbolic cotangent function $\operatorname{coth}(\alpha(q-\frac{1}{2}))$ to soften the non-differentiable point at $q=\frac{1}{2}$. This results in a potential that is $C^2$ almost everywhere, with a singularity at $q=\frac{1}{2}$.

For the Simple Harmonic Oscillator, which admits an analytic solution, we compared our models' predictions with the exact solutions:

\begin{equation}
    q(t) = \frac{1}{2}(1 - \cos (4t)), \quad p(t) = 2 \sin (4t)
\end{equation}

Also, for the Mirrored Free Fall potential, we compared predictions with the exact solutions:

\begin{equation}
    q(t) = \begin{cases}
        2 t^2 & \text{if } 0 \leq t < 0.5 \\
        -2 t^2 + 4t - 1 & \text{if } 0.5 \leq t < 1.5 \\
        2 (2 - t)^2 & \text{if } 1.5 \leq t \leq 2
    \end{cases}, \quad
    p(t) = \begin{cases}
        4t & \text{if } 0 \leq t < 0.5 \\
        -4t + 4 & \text{if } 0.5 \leq t < 1.5 \\
        -4 (2-t) & \text{if } 1.5 \leq t \leq 2
    \end{cases}
\end{equation}

For the remaining potentials, we compared our models' outputs with numerical solutions obtained using a 4th order Gauss-Legendre implicit ODE solver \citep{iserles2009first}. This solver, implemented in the Peroxide library \citep{Peroxide}, was chosen for its ability to preserve the symplectic property of Hamilton's equations, ensuring high accuracy in our reference solutions.

The shapes of these potential functions and their corresponding $q(t)$ and $p(t)$ trajectories are provided in Appendix \ref{app:test_potentials} for reference. This suite of test potentials allows us to evaluate our models' performance across a range of physically relevant scenarios, from well-behaved harmonic systems to more complex anharmonic potentials and even cases that challenge the models' ability to extrapolate beyond their training domain.

\subsection{Performance Metrics and Evaluation Methodology}

To comprehensively assess the performance of our models, we employed four distinct evaluation metrics. These metrics were designed to capture both the accuracy of the predicted trajectories and the computational efficiency of each method. We applied these metrics to our four proposed models—DeepONet, TraONet, VaRONet, and MambONet—as well as to the widely used 4th order Runge-Kutta (RK4) method, which serves as a benchmark for traditional numerical techniques.

The four evaluation metrics are as follows:

\begin{enumerate}
    \item Mean Squared Error (MSE) of the position function:
    \begin{equation}
        \mathcal{L}_q = \frac{1}{N}\sum_{i=1}^N \lVert q_i(\mathbf{t}) - \hat{q}_i (\mathbf{t}) \rVert^2
    \end{equation}
    
    \item Mean Squared Error (MSE) of the momentum function:
    \begin{equation}
        \mathcal{L}_p = \frac{1}{N} \sum_{i=1}^N \lVert p_i(\mathbf{t}) - \hat{p}_i(\mathbf{t})\rVert^2
    \end{equation}
    
    \item Total loss, defined as the average of position and momentum MSE:
    \begin{equation}
        \mathcal{L}_{\text{tot}} = \frac{1}{2} (\mathcal{L}_q + \mathcal{L}_p)
    \end{equation}
    
    \item Computation time for generating a trajectory from a single potential:
    \begin{equation}
        \mathcal{T} = \sum_{i=1}^N \mathcal{T}_i
    \end{equation}
\end{enumerate}

In these equations, $N$ represents the number of test samples, $q_i(\mathbf{t})$ and $p_i(\mathbf{t})$ are the true position and momentum trajectories, $\hat{q}_i (\mathbf{t})$ and $\hat{p}_i(\mathbf{t})$ are the predicted trajectories, and $\mathcal{T}_i$ is the computation time for the $i$-th point.

The total loss $\mathcal{L}_{\text{tot}}$ serves as our primary metric for comparing model performance, providing a comprehensive measure of prediction accuracy for both position and momentum. The separate position and momentum MSEs ($\mathcal{L}_q$ and $\mathcal{L}_p$) allow us to assess whether certain models excel in predicting one aspect of the system's state over the other. The average computation time $\mathcal{T}$ enables us to evaluate the computational efficiency of each method, which is crucial for practical applications where rapid trajectory computation may be essential.

\subsubsection{Statistical Analysis}

Given the potentially skewed nature of loss distributions, particularly for neural network models, we report both parametric and non-parametric statistics. For each metric, we provide:

\begin{itemize}
    \item Mean and standard deviation: These offer a traditional measure of central tendency and dispersion.
    \item Median and interquartile range (IQR): These non-parametric statistics provide a robust representation of central tendency and spread, minimizing the influence of outliers \cite{dekking2006modern}.
\end{itemize}

This comprehensive set of statistics allows for a more nuanced understanding of model performance, especially in cases where the distribution of losses may not follow a normal distribution.

\section{Results}

\subsection{Performance Comparison on the Test Dataset}

To evaluate the performance of our proposed models, we conducted a comprehensive analysis using both the standard dataset (10,000 potentials) and the extended dataset (100,000 potentials). We compared our models—DeepONet, TraONet, VaRONet, and MambONet—against the traditional 4th order Runge-Kutta (RK4) method, which serves as a benchmark for numerical solutions of Hamilton's equations. 

Our primary focus is on the total loss $\mathcal{L}_{\text{tot}}$, which represents the average of the position and momentum mean squared errors (MSEs). This metric provides a comprehensive measure of the overall prediction accuracy of the models. Additionally, we assessed the computational efficiency through the computation time $\mathcal{T}$.

\begin{figure}
    \centering
    \begin{subfigure}[b]{0.49\textwidth}
        \includegraphics[width=\textwidth]{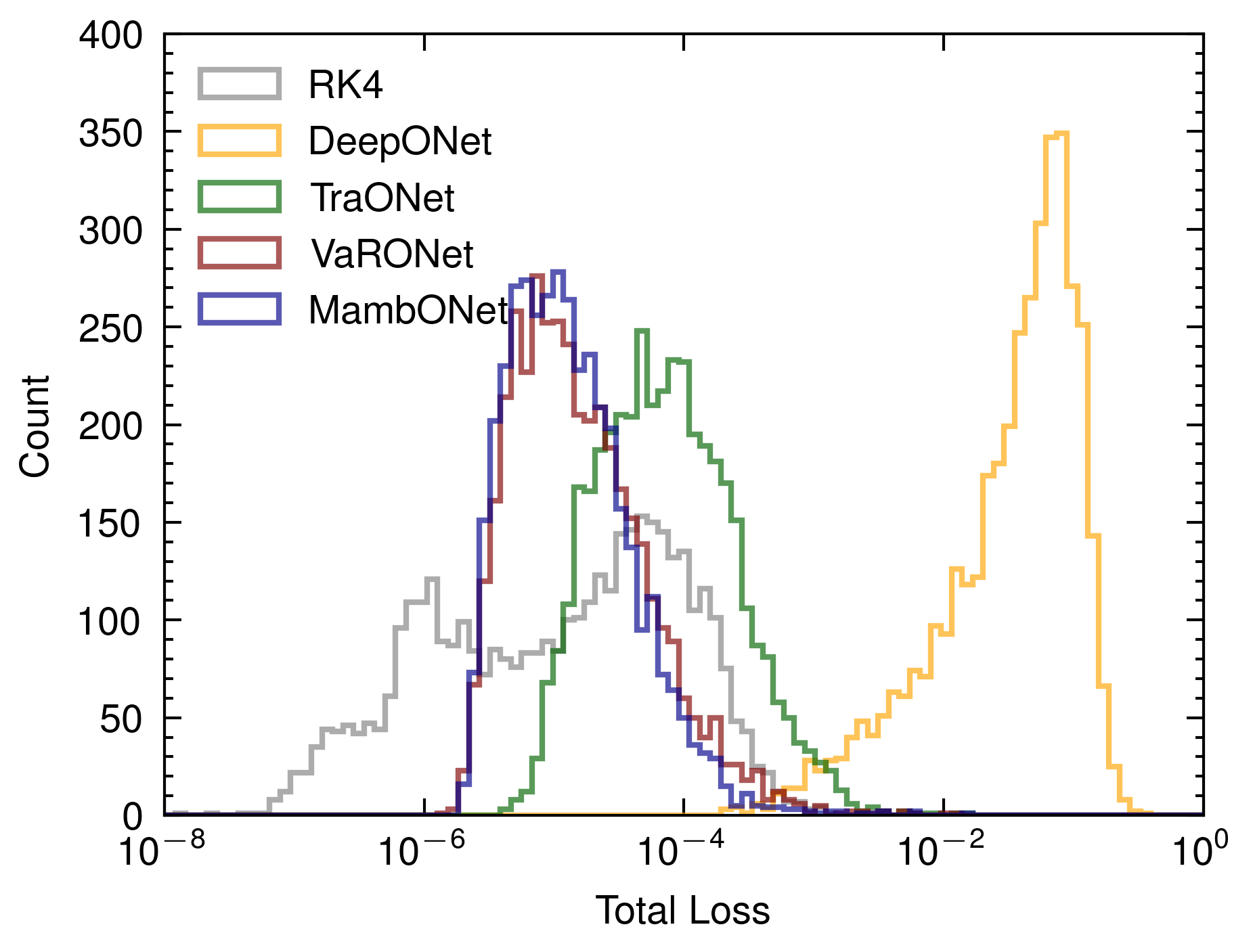}
        \caption{Standard dataset (10,000 potentials)}
    \end{subfigure}
    \hfill
    \begin{subfigure}[b]{0.49\textwidth}
        \includegraphics[width=\textwidth]{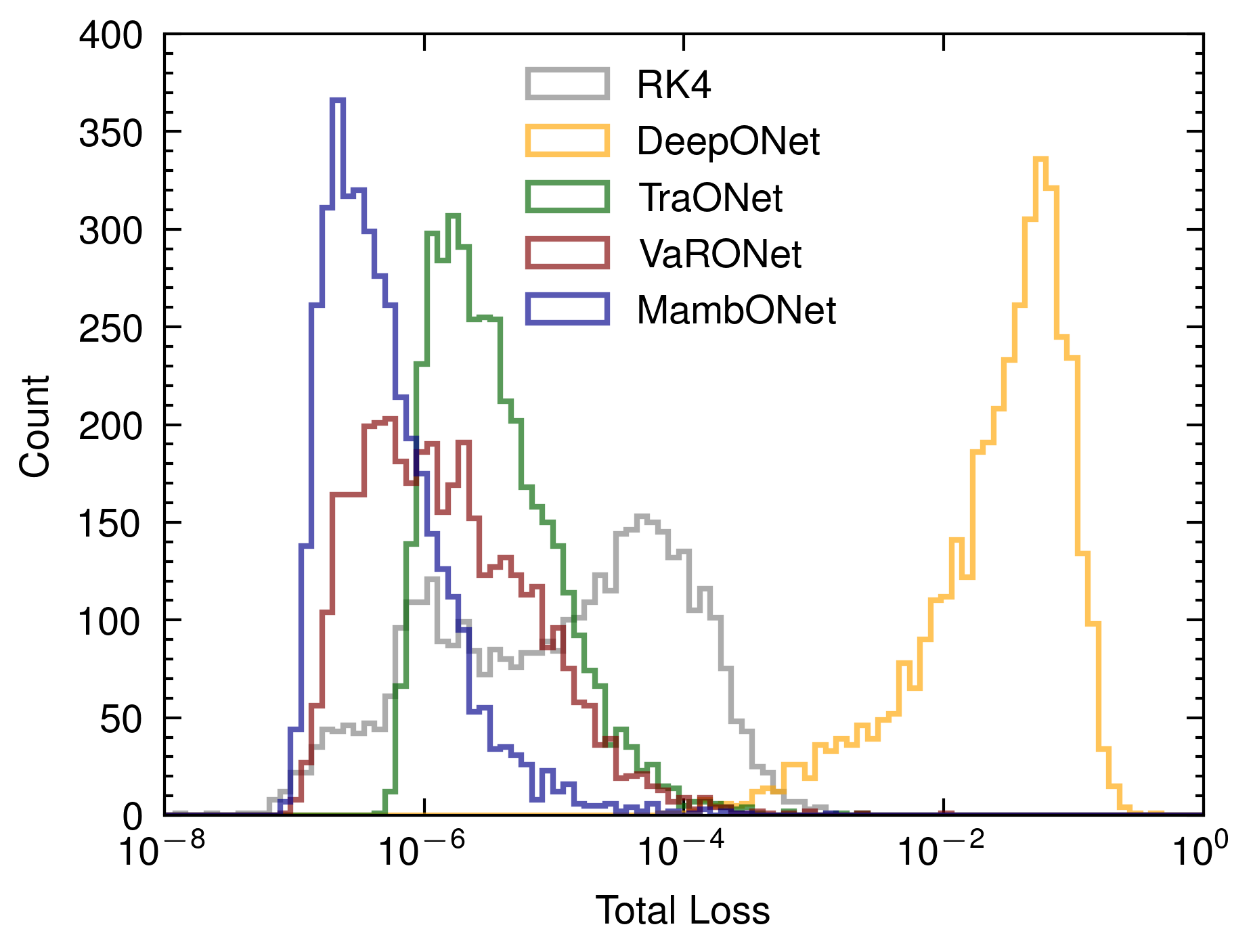}
        \caption{Extended dataset (100,000 potentials)}
    \end{subfigure}
    \caption{Distribution of total test losses ($\mathcal{L}_{\text{tot}}$) for different models. RK4 (Gray), DeepONet (Yellow), TraONet (Green), VaRONet (Red), and MambONet (Blue).}
    \label{fig:loss_distribution}
\end{figure}

Figure \ref{fig:loss_distribution} presents the distribution of the total test losses ($\mathcal{L}_{\text{tot}}$) for each model on both datasets. Tables \ref{tab:standard_dataset_performance} and \ref{tab:extended_dataset_performance} provide detailed statistical metrics for the standard and extended datasets, respectively.

\begin{table}
\caption{Test dataset performance metrics for models trained on the standard dataset. Bold and underlined values indicate the best and second-best performing models, respectively.}
\label{tab:standard_dataset_performance}
\centering
\renewcommand{\arraystretch}{1.25}
\begin{tabular}{lcccc}
\toprule
Model & Mean $\mathcal{L}_{\text{tot}}$ & Std $\mathcal{L}_{\text{tot}}$ & Median $\mathcal{L}_{\text{tot}}$ & IQR $\mathcal{L}_{\text{tot}}$ \\
\midrule
RK4 & $5.2852 \times 10^{-5}$ & $\pmb{1.0586 \times 10^{-4}}$ & $1.4802 \times 10^{-5}$ & $6.0370 \times 10^{-5}$ \\
DeepONet & $5.2533 \times 10^{-2}$ & $4.4131 \times 10^{-2}$ & $4.2962 \times 10^{-2}$ & $6.2856 \times 10^{-2}$ \\
TraONet & $1.6787 \times 10^{-4}$ & $4.6630 \times 10^{-4}$ & $6.9309 \times 10^{-5}$ & $1.3676 \times 10^{-4}$ \\
VaRONet & $\underline{5.2175 \times 10^{-5}}$ & $3.3825 \times 10^{-4}$ & $\underline{1.3215 \times 10^{-5}}$ & $\underline{2.6575 \times 10^{-5}}$ \\
MambONet & $\pmb{3.8228 \times 10^{-5}}$ & $\underline{3.2109 \times 10^{-4}}$ & $\pmb{1.1698 \times 10^{-5}}$ & $\pmb{1.9854 \times 10^{-5}}$ \\
\bottomrule
\end{tabular}
\end{table}

\begin{table}
    \caption{Test dataset performance metrics for models trained on the extended dataset.}
\label{tab:extended_dataset_performance}
\centering
\renewcommand{\arraystretch}{1.25}
\begin{tabular}{lcccc}
\toprule
Model & Mean $\mathcal{L}_{\text{tot}}$ & Std $\mathcal{L}_{\text{tot}}$ & Median $\mathcal{L}_{\text{tot}}$ & IQR $\mathcal{L}_{\text{tot}}$ \\
\midrule
RK4 & $5.2852 \times 10^{-5}$ & $1.0586 \times 10^{-4}$ & $1.4802 \times 10^{-5}$ & $6.0370 \times 10^{-5}$ \\
DeepONet & $4.4341 \times 10^{-2}$ & $4.3616 \times 10^{-2}$ & $3.4747 \times 10^{-2}$ & $5.1906 \times 10^{-2}$ \\
    TraONet & $9.3756 \times 10^{-6}$ & $\underline{5.0914 \times 10^{-5}}$ & $2.9045 \times 10^{-6}$ & $5.4109 \times 10^{-6}$ \\
VaRONet & $\underline{9.3437 \times 10^{-6}}$ & $1.6002 \times 10^{-4}$ & $\underline{1.2522 \times 10^{-6}}$ & $\underline{3.7719 \times 10^{-6}}$ \\
MambONet & $\pmb{1.8749 \times 10^{-6}}$ & $\pmb{2.1925 \times 10^{-5}}$ & $\pmb{3.9598 \times 10^{-7}}$ & $\pmb{6.4351 \times 10^{-7}}$ \\
\bottomrule
\end{tabular}
\end{table}

\subsubsection{Standard Dataset Results}

For the standard dataset, the RK4 method demonstrated a relatively uniform distribution of total losses ($\mathcal{L}_{\text{tot}}$) ranging from $10^{-7}$ to $10^{-3}$. In contrast, our neural network-based models showed more concentrated loss distributions. 

DeepONet, our baseline neural operator model, exhibited the highest total losses among the neural models, with its best performance around $10^{-4}$ and the majority of losses falling between $10^{-2}$ and $10^{-1}$. This suggests that while DeepONet can learn the operator mapping, it struggles to achieve high accuracy consistently across different potentials.

TraONet, which incorporates trajectory information, showed significant improvement over DeepONet, with most losses clustered around $10^{-4}$. This improvement indicates the benefits of leveraging temporal information in the operator learning process.

VaRONet and MambONet demonstrated further improvements, with their best cases approaching $10^{-6}$. While these best cases did not surpass RK4's performance, both models showed more consistent performance, with the majority of their losses concentrated near their best cases. Notably, VaRONet and MambONet achieved mean total losses of $5.2175 \times 10^{-5}$ and $3.8228 \times 10^{-5}$ respectively, outperforming RK4's mean of $5.2852 \times 10^{-5}$. This suggests that these advanced models can compete with, and in some cases surpass, traditional numerical methods in terms of accuracy.

\subsubsection{Extended Dataset Results}

The extended dataset results revealed interesting patterns in model scalability. DeepONet showed only marginal improvement across all metrics, with its mean total loss decreasing from $5.2533 \times 10^{-2}$ to $4.4341 \times 10^{-2}$, suggesting limited benefit from increased data volume for this baseline model.

In contrast, the other neural network models demonstrated substantial improvements. TraONet's performance distribution became sharper and shifted towards lower loss values, with both its mean ($9.3756 \times 10^{-6}$) and median ($2.9045 \times 10^{-6}$) total losses surpassing RK4's performance. This indicates that TraONet's architecture is particularly effective at leveraging larger datasets to improve its operator learning capabilities.

VaRONet maintained its edge over TraONet in most cases, achieving a lower median total loss of $1.2426 \times 10^{-6}$. However, its higher standard deviation ($1.5929 \times 10^{-4}$) compared to TraONet ($5.0914 \times 10^{-5}$) suggests some instability or variability in its performance across different potentials. This variability warrants further investigation to understand the conditions under which VaRONet may underperform.

MambONet exhibited remarkable performance on the extended dataset, excelling across all metrics. Its median total loss ($3.9598 \times 10^{-7}$) was an order of magnitude lower than TraONet's, and its significantly lower interquartile range (IQR) of $6.4351 \times 10^{-7}$ indicated the most robust and consistent performance among all models. This suggests that MambONet's architecture is particularly well-suited for capturing the complexities of Hamiltonian systems when provided with sufficient training data.

\subsubsection{Computational Efficiency}

\begin{table}
\caption{Computational efficiency metrics (time in seconds)}
\label{tab:computation_time}
\centering
\renewcommand{\arraystretch}{1.25}
\begin{tabular}{lcccc}
\toprule
Model & Mean $\mathcal{T}$ & Std $\mathcal{T}$ & Median $\mathcal{T}$ & IQR $\mathcal{T}$ \\
\midrule
RK4 & $\underline{5.1527 \times 10^{-3}}$ & $\pmb{6.4416 \times 10^{-5}}$ & $\underline{5.1391 \times 10^{-3}}$ & $9.9719 \times 10^{-5}$ \\
DeepONet & $8.3430 \times 10^{-3}$ & $\underline{7.3357 \times 10^{-4}}$ & $8.3201 \times 10^{-3}$ & $\underline{8.8453 \times 10^{-5}}$ \\
TraONet & $\pmb{2.4459 \times 10^{-3}}$ & $8.4676 \times 10^{-4}$ & $\pmb{2.4078 \times 10^{-3}}$ & $\pmb{5.4657 \times 10^{-5}}$ \\
VaRONet & $1.7999 \times 10^{-2}$ & $1.5616 \times 10^{-3}$ & $1.7972 \times 10^{-2}$ & $1.1086 \times 10^{-4}$ \\
MambONet & $8.5305 \times 10^{-3}$ & $1.2290 \times 10^{-3}$ & $8.5326 \times 10^{-3}$ & $2.2578 \times 10^{-4}$ \\
\bottomrule
\end{tabular}
\end{table}

We assessed the computational efficiency of our models using those trained on the extended dataset, as the number of parameters remains constant regardless of the training dataset size. Table \ref{tab:computation_time} presents the computational efficiency metrics for each model, calculated with a batch size of 1.

TraONet demonstrates the fastest computation time (mean $2.4459 \times 10^{-3}$ seconds), outperforming RK4 ($5.1527 \times 10^{-3}$ seconds). DeepONet and MambONet show similar computation times (approximately $8.4 \times 10^{-3}$ seconds), about 1.6 times slower than RK4, while VaRONet is the slowest at $1.7999 \times 10^{-2}$ seconds. RK4 exhibits the most consistent performance with the lowest standard deviation.

It's important to note that these results represent sequential processing of individual potentials. Neural network models have the advantage of GPU parallelization for batch processing. In scenarios requiring predictions for multiple potentials simultaneously, increasing the batch size could potentially lead to computation times significantly faster than RK4, a unique advantage over traditional numerical methods.

These findings highlight a trade-off between accuracy and computational efficiency, while also suggesting potential superiority of neural network models in batch processing scenarios. This could be particularly advantageous in applications like large-scale molecular dynamics simulations or parameter sweeps in celestial mechanics calculations.

\subsection{Test Results on Physically Relevant Potentials}

To further evaluate the performance and generalization capabilities of our models, we conducted tests using the physically relevant potentials described in Section 4.3.
These potentials represent common scenarios in physics and provide a robust test for our models' ability to handle diverse and realistic situations.
For these tests, we used our models trained on the extended dataset as they demonstrated superior performance compared to those trained on the standard dataset. 

\begin{table}
\caption{Performance on physically relevant potentials (Total Loss $\mathcal{L}_{\text{tot}}$)}
\label{tab:physical_potentials_performance}
\centering
\renewcommand{\arraystretch}{1.25}
\begin{adjustbox}{center}
\begin{tabular}{lccccc}
\toprule
Model & SHO & Double Well & Morse & MFF & SMFF \\
\midrule
RK4 & $3.3663 \times 10^{-5}$ & $1.4362\times 10^{-2}$ & $2.8753 \times 10^{-4}$ & $1.5224 \times 10^{-4}$ & $4.1551 \times 10^{-5}$ \\
DeepONet & $2.4951 \times 10^{-4}$ & $8.9770 \times 10^{-2}$ & $4.7225 \times 10^{-2}$ & $2.5406 \times 10^{-2}$ & $1.5242 \times 10^{-2}$ \\
TraONet & $1.0145 \times 10^{-6}$ & $2.2207 \times 10^{-6}$ & $7.6594 \times 10^{-6}$ & $\pmb{1.0228 \times 10^{-4}}$ & $\underline{3.3919 \times 10^{-5}}$ \\
VaRONet & $\underline{1.9729 \times 10^{-7}}$ & $\underline{8.8354 \times 10^{-7}}$ & $\underline{2.3114 \times 10^{-6}}$ & $2.0465 \times 10^{-4}$ & $7.2018 \times 10^{-5}$ \\
MambONet & $\pmb{1.4197 \times 10^{-7}}$ & $\pmb{4.3837 \times 10^{-7}}$ & $\pmb{1.3451 \times 10^{-6}}$ & $\underline{1.3426 \times 10^{-4}}$ & $\pmb{1.9754 \times 10^{-6}}$ \\
\bottomrule
\end{tabular}
\end{adjustbox}
\end{table}

The results in Table \ref{tab:physical_potentials_performance} reveal several interesting patterns across the different potentials and models. Firstly, DeepONet consistently shows the poorest performance across all potentials, mirroring its behavior on the test dataset. This suggests that the baseline DeepONet architecture may be insufficient for capturing the complexities of these physically relevant Hamiltonian systems.

For the Simple Harmonic Oscillator (SHO), Double Well, and Morse potentials, our advanced neural network models (TraONet, VaRONet, and MambONet) demonstrate superior performance compared to RK4.
This is particularly noteworthy given that RK4 is a well-established numerical method for solving differential equations.
MambONet exhibits exceptional performance, achieving the lowest total loss for these three potentials.
Its superiority is remarkable across all three, outperforming RK4 by over two orders of magnitude for both the SHO and Morse potentials. The improvement is even more dramatic for the Double Well potential, where MambONet's performance surpasses RK4 by nearly five orders of magnitude.

The Mirrored Free Fall (MFF) potential provides a crucial test for the extrapolation capabilities of our models. Being only $C^0$ continuous, it lies outside the function space of our training dataset. For this potential, RK4 shows an error level comparable to most other potentials (except for the Double Well, where its error significantly increases). In contrast, the neural network models demonstrate a notable decrease in performance for MFF compared to other potentials, highlighting the challenges of extrapolation beyond their training distribution. Despite this difficulty, both TraONet and MambONet still manage to outperform RK4 on the MFF potential, with TraONet achieving a total loss of $1.0228 \times 10^{-4}$ and MambONet $1.3426 \times 10^{-4}$, compared to RK4's $1.5224 \times 10^{-4}$. VaRONet, while not surpassing RK4, shows comparable performance with a loss of $2.0465 \times 10^{-4}$. These results are particularly impressive considering that the MFF potential is outside the training distribution and has lower smoothness than the potentials used in training. They demonstrate a remarkable ability of our neural models, especially TraONet and MambONet, to generalize beyond their training distribution, even under challenging conditions.

The performance improvement becomes even more pronounced for the Softened Mirrored Free Fall (SMFF) potential, which is $C^2$ continuous almost everywhere (except at $q=\frac{1}{2}$) and thus closer to the function space of the training data.
MambONet, in particular, achieves a total loss of $1.9754 \times 10^{-6}$, more than 20 times lower than RK4's $4.1551 \times 10^{-5}$. TraONet also outperforms RK4 with a loss of $3.3919 \times 10^{-5}$. Interestingly, VaRONet shows less consistent performance across MFF and SMFF, with its loss for SMFF ($7.2018 \times 10^{-5}$) being higher than RK4's, despite its strong performance on other potentials.

The significant performance improvement observed in all neural network models when moving from MFF to SMFF underscores the importance of precisely understanding the function space covered by the training data.
This transition from $C^0$ continuity to almost everywhere $C^2$ continuity (with a single point of non-differentiability) leads to a substantial increase in prediction accuracy, demonstrating that these models are highly sensitive to the smoothness properties of the input functions.

\begin{figure}
    \begin{adjustwidth}{-0.5cm}{-0.5cm}
    \centering
    \begin{subfigure}[b]{0.35\textwidth}
        \includegraphics[width=\textwidth]{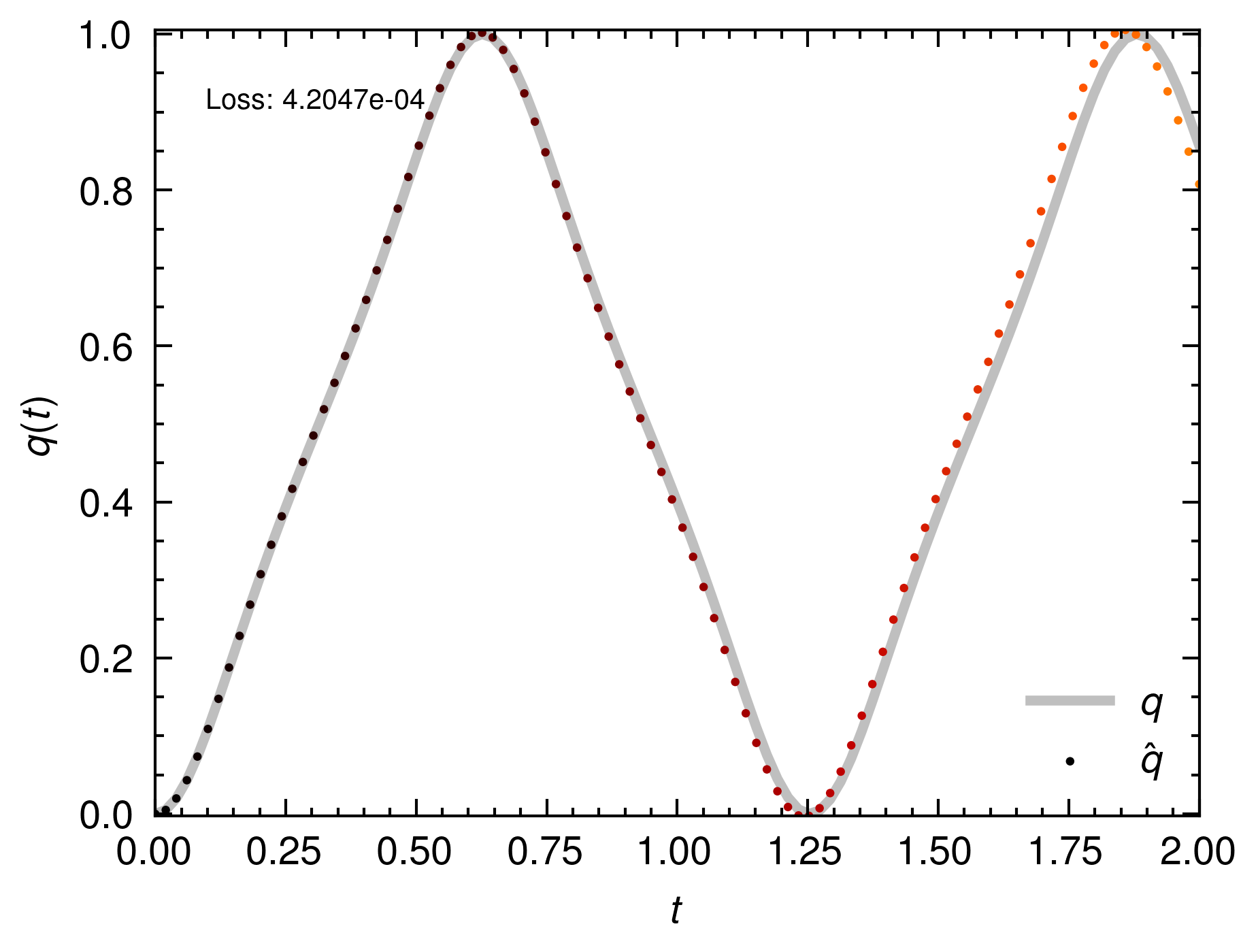}
        \vspace{-0.5cm}
        \caption{Position $q(t)$ (RK4)}
    \end{subfigure}
    \hfill
    \begin{subfigure}[b]{0.35\textwidth}
        \includegraphics[width=\textwidth]{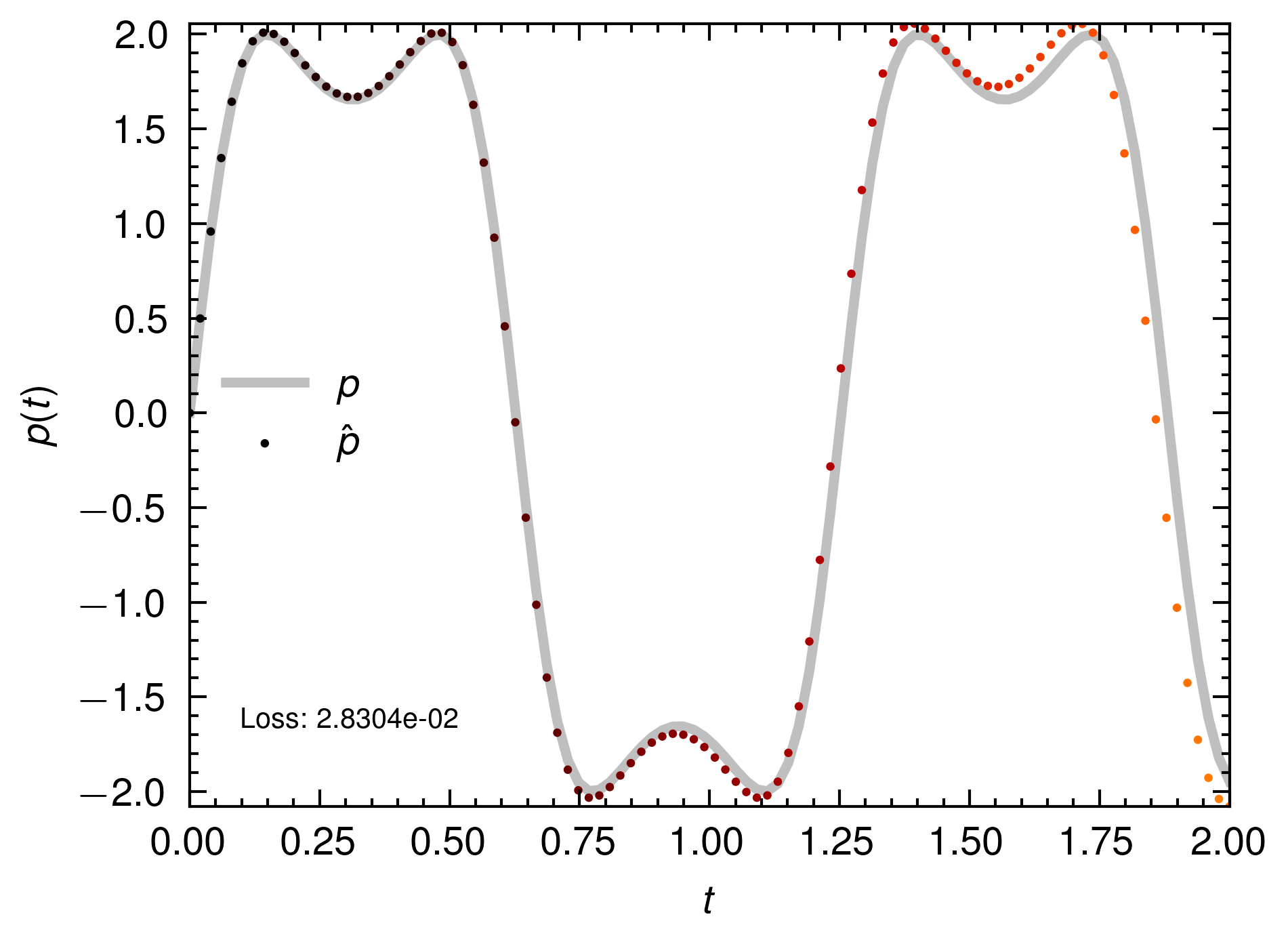}
        \vspace{-0.5cm}
        \caption{Momentum $p(t)$ (RK4)}
    \end{subfigure}
    \hfill
    \begin{subfigure}[b]{0.35\textwidth}
        \includegraphics[width=\textwidth]{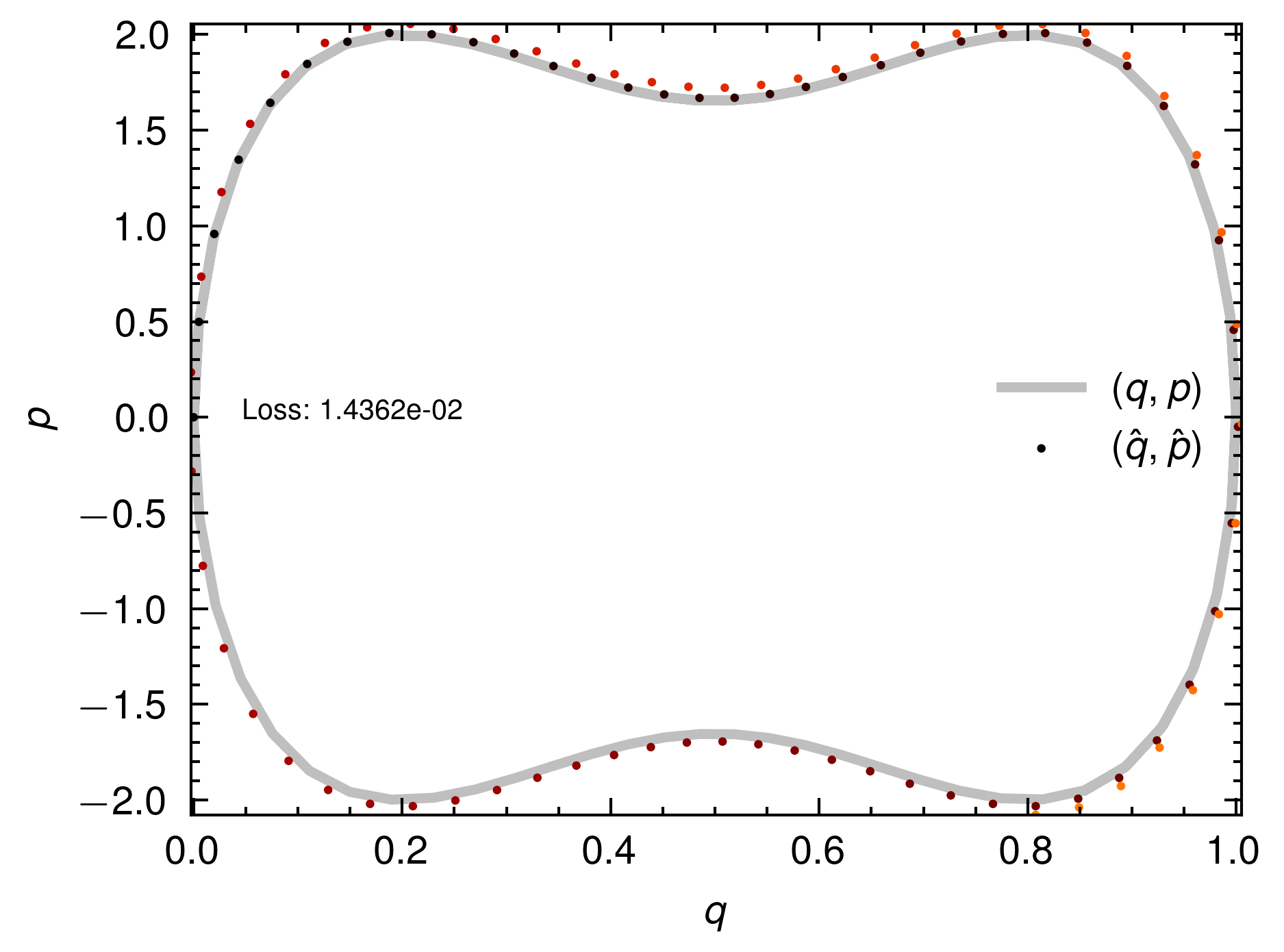}
        \vspace{-0.5cm}
        \caption{Phase plot (RK4)}
    \end{subfigure}
    
    \vspace{0.4cm}

    \begin{subfigure}[b]{0.35\textwidth}
        \includegraphics[width=\textwidth]{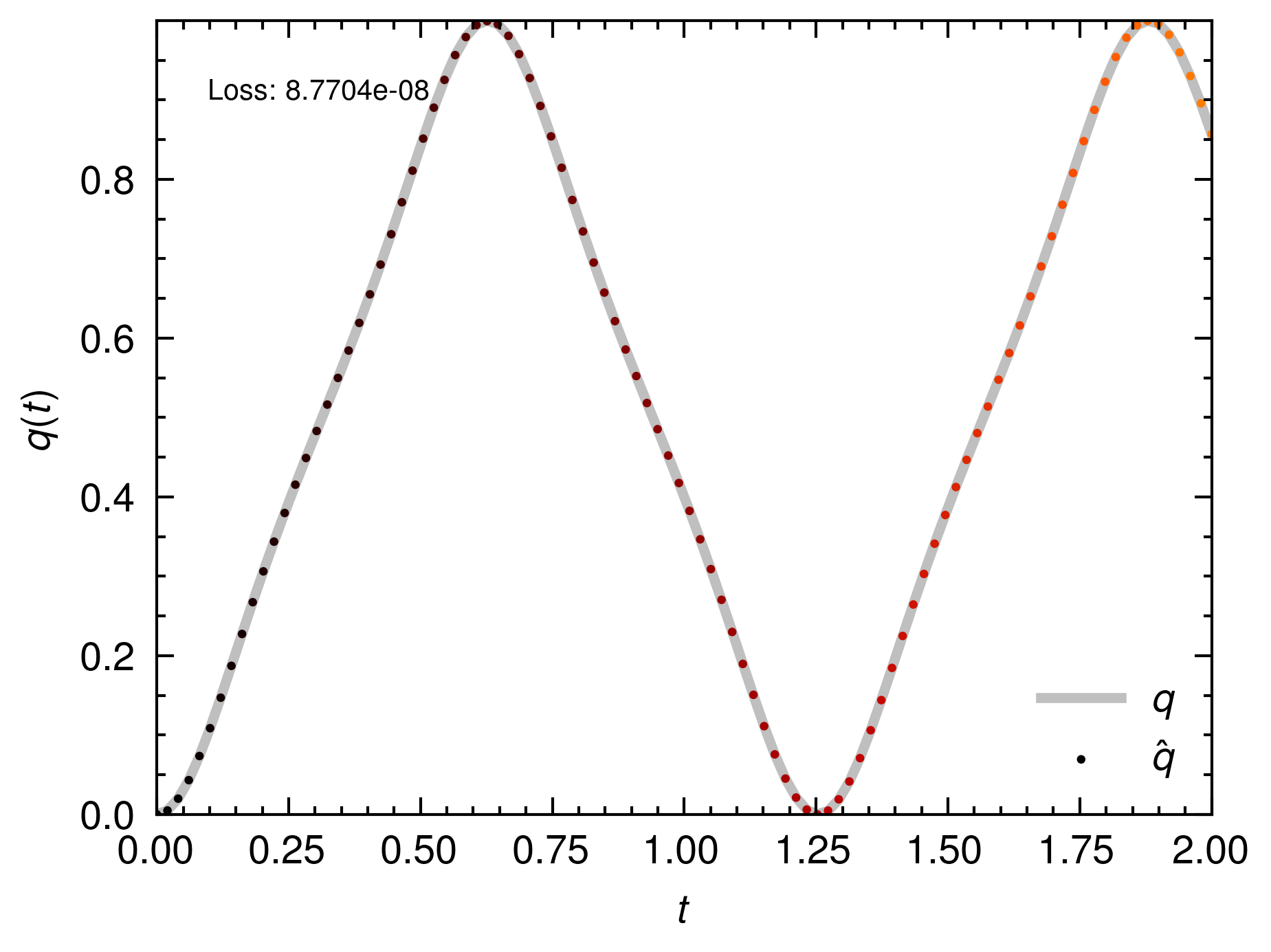}
        \vspace{-0.5cm}
        \caption{Position $q(t)$ (MambONet)}
    \end{subfigure}
    \hfill
    \begin{subfigure}[b]{0.35\textwidth}
        \includegraphics[width=\textwidth]{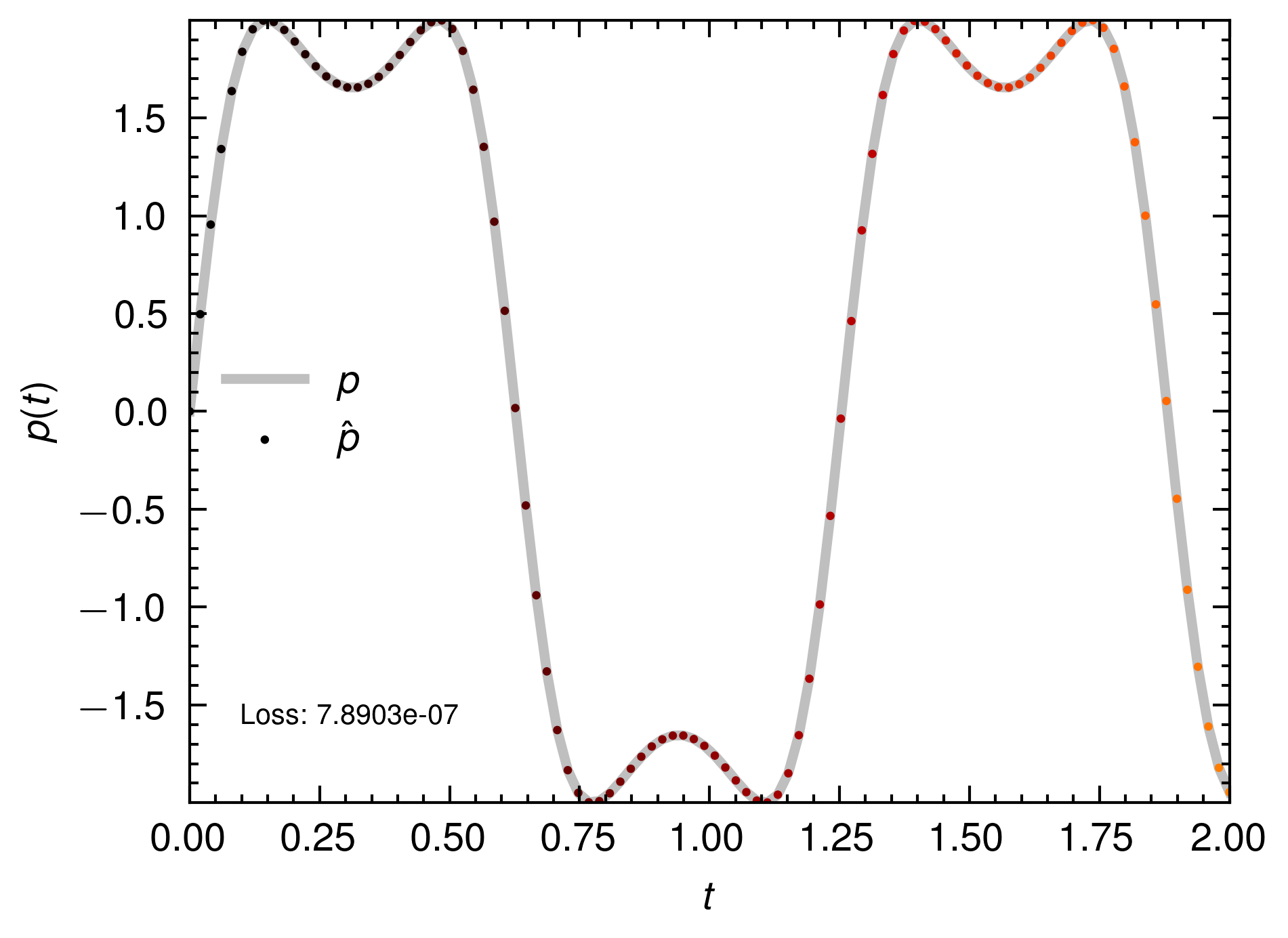}
        \vspace{-0.5cm}
        \caption{Momentum $p(t)$ (MambONet)}
    \end{subfigure}
    \hfill
    \begin{subfigure}[b]{0.35\textwidth}
        \includegraphics[width=\textwidth]{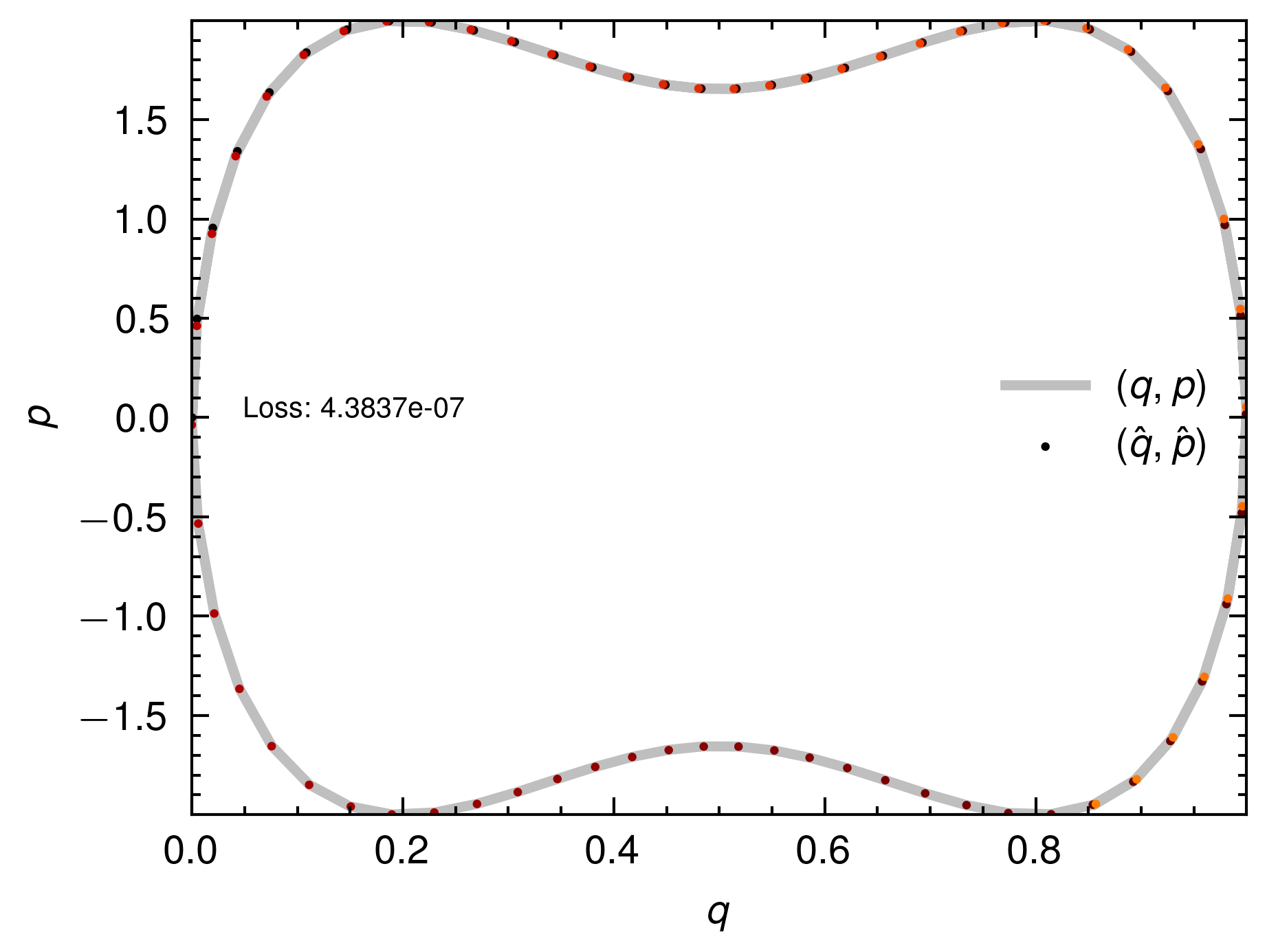}
        \vspace{-0.5cm}
        \caption{Phase plot (MambONet)}
    \end{subfigure}
    \caption{Comparison of RK4 (top row) and MambONet (bottom row) solutions for the Double Well potential.
        Transparent gray lines represent the ground truth, while scatter points transitioning from black to orange over time show the predicted solutions.}
    \label{fig:quartic_comparison}
    \end{adjustwidth}
\end{figure}

To visualize the performance difference between RK4 and our best-performing model, MambONet, we present a comparison of their solutions for the Double Well potential in Figure \ref{fig:quartic_comparison}. The figure clearly illustrates the superior accuracy of MambONet compared to RK4. The MambONet predictions closely align with the ground truth for both position and momentum trajectories, as well as in the phase space representation. In contrast, the RK4 solution exhibits cumulative deviations starting from approximately $t = 0.4$ onwards, becoming increasingly pronounced in both position and momentum predictions. This accumulation of errors leads to significant discrepancies in the phase plot. The comparison underscores MambONet's ability to maintain accuracy over the entire time domain, while RK4's performance degrades as time progresses.

These results underscore the potential of our advanced neural operator models, especially MambONet, in solving Hamilton's equations for a wide range of physically relevant potentials. The models' ability to outperform traditional numerical methods like RK4, even for potentials outside their training distribution, suggests promising applications in various fields of physics and engineering where rapid, accurate solutions to Hamiltonian systems are required. 

Moreover, these findings highlight the importance of function space considerations in the performance of neural operator models. While the models show impressive generalization to potentials outside their training distribution, their performance is notably enhanced when the test potentials more closely align with the smoothness characteristics of the training data. This observation provides valuable insights for future improvements in neural operator learning for physical systems, emphasizing the need for careful consideration of the function spaces involved in both training and application.

\section{Discussion}

\subsubsection{Model Performance Analysis}
The results of our study provide significant insights into the performance of various neural network architectures for solving Hamilton's equations, as well as the impact of dataset size on operator learning. We observed distinct patterns in the performance of DeepONet, TraONet, VaRONet, and MambONet across different scenarios and dataset sizes.

DeepONet's relatively poor performance in our experiments warrants further discussion. It's important to note that this underperformance may not be intrinsic to the DeepONet architecture itself, but rather a result of the optimization process being incompatible with DeepONet's learning dynamics. As mentioned earlier, we observed step-like decreases in DeepONet's loss curve during training, suggesting that its performance at any given epoch might be due to fortunate convergence rather than consistent improvement. Ideally, hyperparameter tuning should be performed at the target number of epochs (250), but this was computationally prohibitive for our study. To ensure fair comparison, we applied the same optimization protocol across all models, which may not have been optimal for DeepONet's specific characteristics.

TraONet, consistent with our previous findings in the HyperbolicLR study \citep{kim2024hyperboliclr}, demonstrated significant improvement over DeepONet. While not the top-performing model in all scenarios, TraONet exhibited robust performance characterized by the lowest standard deviation and the best performance in extrapolation tasks. Moreover, it boasted the fastest single execution speed, making it arguably the most efficient model in our study.

VaRONet generally ranked second in performance among the models. However, it showed limitations in terms of execution speed and extrapolation capabilities. These aspects highlight areas for potential improvement in future iterations of the VaRONet architecture.

MambONet consistently delivered the most impressive performance across various metrics. Its execution speed, while not matching TraONet, was comparable to RK4, indicating its potential for practical applications. The success of MambONet, which combines a Mamba encoder with a Transformer decoder, both of which have shown promise in large language models, raises intriguing questions about the relationship between operator learning and language modeling. This connection could be a fruitful avenue for future research.

\subsubsection{Impact of Dataset Size on Operator Learning}
The impact of dataset size on operator learning performance was particularly noteworthy in our study. We observed substantial improvements in performance when moving from the standard dataset (10,000 potentials) to the extended dataset (100,000 potentials). This improvement was especially pronounced for TraONet and MambONet, which showed approximately tenfold enhancements in performance.

\begin{figure}
    \centering
    \includegraphics[width=0.6\textwidth]{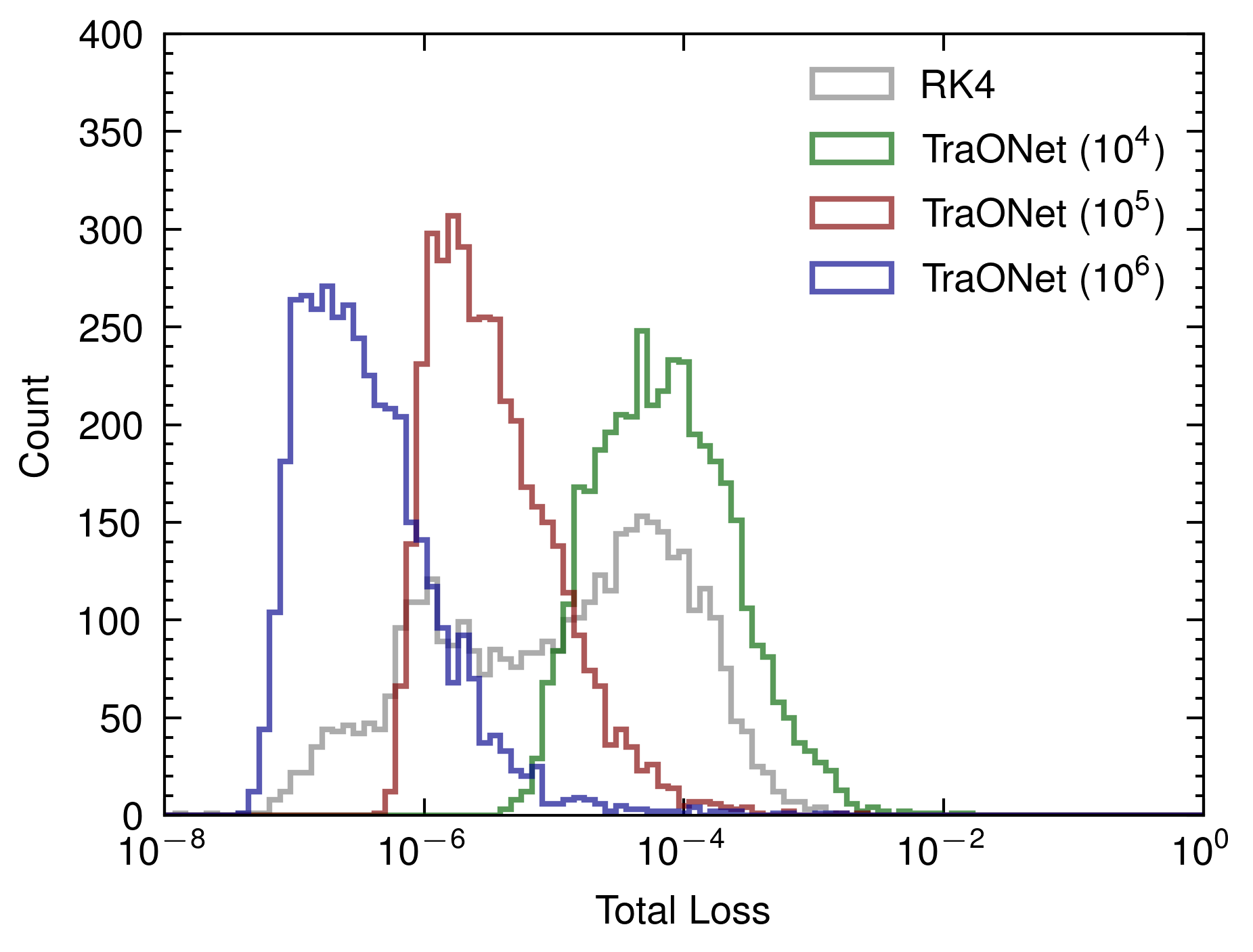}
    \caption{Distribution of total test losses ($\mathcal{L}_{\text{tot}}$) for TraONet trained on different dataset sizes. RK4 (Gray), TraONet($10^4$) (Green), TraONet($10^5$) (Red), TraONet($10^6$) (Blue).}
    \label{fig:traONet_dataset_comparison}
\end{figure}

To further explore the effect of dataset size, we conducted an additional experiment with TraONet using a huge dataset of one million potentials. The results of this experiment were remarkable, as illustrated in Figure \ref{fig:traONet_dataset_comparison}.

The performance improvement observed with the huge dataset was not only significant but also consistent across various types of potentials, including physically relevant ones. Table \ref{tab:traONet_huge_performance} presents the performance of TraONet trained on one million potentials for physically relevant test cases.

\begin{table}
\caption{Performance of TraONet trained on $10^6$ potentials for physically relevant potentials ($\mathcal{L}_{\text{tot}}$)}
\label{tab:traONet_huge_performance}
\centering
\renewcommand{\arraystretch}{1.25}
\begin{adjustbox}{center}
\begin{tabular}{lccccc}
\toprule
Model & SHO & Double Well & Morse & MFF & SMFF \\
\midrule
TraONet ($10^6$) & $7.8275 \times 10^{-8}$ & $3.0411 \times 10^{-7}$ & $6.9798 \times 10^{-7}$ & $1.6294 \times 10^{-5}$ & $2.4104 \times 10^{-6}$ \\
\bottomrule
\end{tabular}
\end{adjustbox}
\end{table}

These results are comparable to or better than those achieved by MambONet trained on 100,000 potentials for most potential types. The performance on the Mirrored Free Fall (MFF) potential is particularly noteworthy, showing a tenfold improvement compared to other models. This significant enhancement in extrapolation capability suggests that the larger dataset expands the function space covered by the model, effectively reducing the distance between the training function space and the MFF potential.

While we would have liked to test other models, such as MambONet, on the huge dataset, the computational resources required were beyond the scope of this study. TraONet's efficiency allowed for this extended experiment, but other models would have required substantially more time and computational power. This limitation highlights a potential direction for future research: with sufficient GPU resources, training large-scale "Neural Hamilton" models on diverse and extensive potential datasets could yield even more impressive results.

The findings from our TraONet experiment with the huge dataset underscore the potential benefits of large-scale training in operator learning for Hamiltonian systems. They suggest that with sufficient data and computational resources, we may be able to develop highly accurate and generalizable models for solving Hamilton's equations across a wide range of potential functions.

These results open up several avenues for future research. First, investigating the performance of other models, particularly MambONet, on similarly large datasets could provide valuable insights into the scalability of different architectures. Second, exploring the relationship between dataset size, model complexity, and performance could help in developing guidelines for efficient training of operator learning models in physics. Finally, the observed improvements in extrapolation capabilities with larger datasets warrant further study into the nature of function spaces learned by these models and how they relate to the physical systems they represent.

\subsubsection{Error Propagation in Operator Learning vs. Traditional Methods}
Another striking aspect of operator learning revealed in our study is its resilience to cumulative errors over time, contrasting sharply with iterative ODE solvers. This difference is particularly evident in the performance of our models on the Mirrored Free Fall (MFF) potential, which serves as an extrapolation test case due to its non-differentiability at $q=0.5$.

\begin{figure}
    \begin{adjustwidth}{-0.5cm}{-0.5cm}
        \centering
        \begin{subfigure}[b]{0.52\textwidth}
            \centering
            \includegraphics[width=\textwidth]{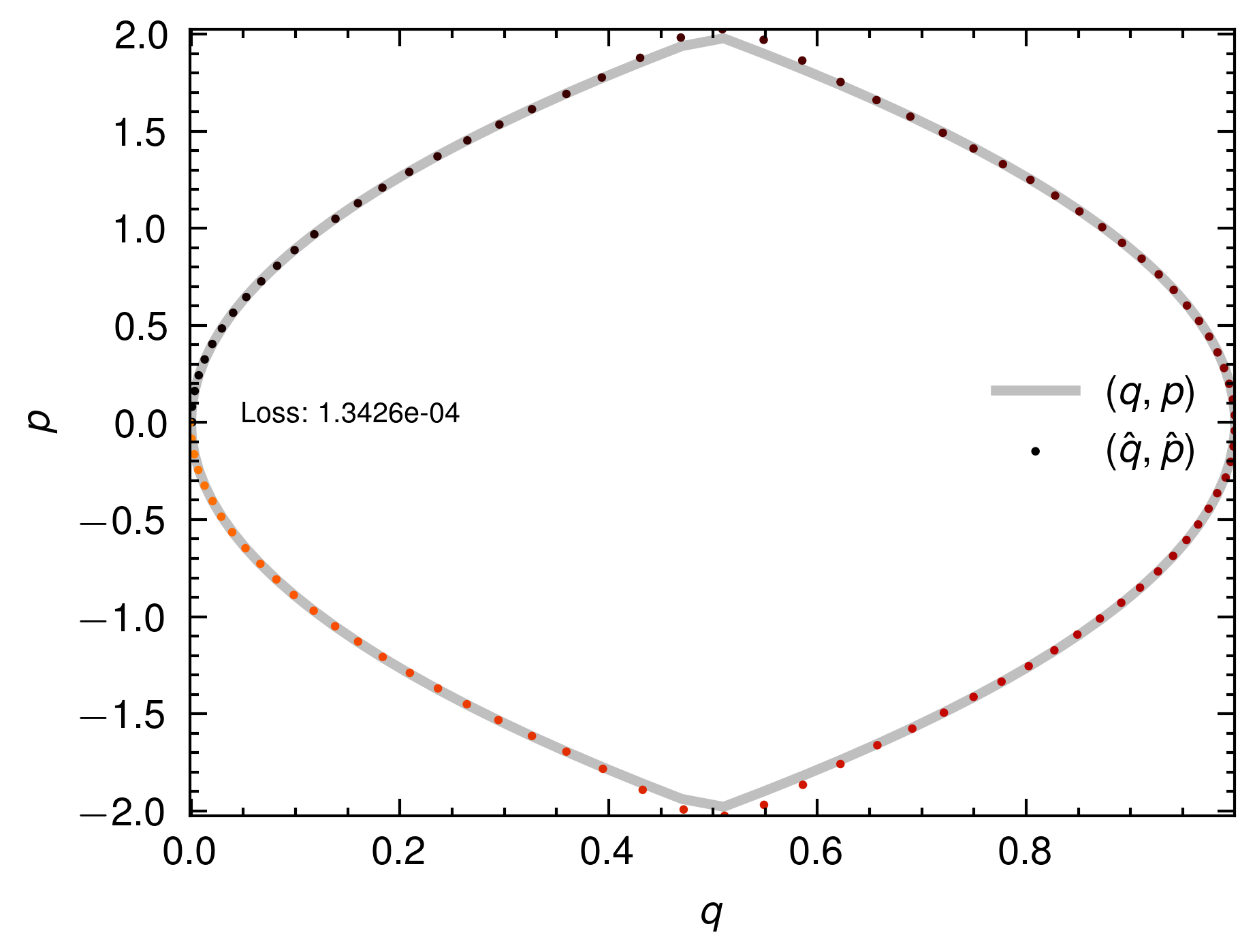}
            \caption{MambONet}
        \end{subfigure}
        \hfill
        \begin{subfigure}[b]{0.52\textwidth}
            \centering
            \includegraphics[width=\textwidth]{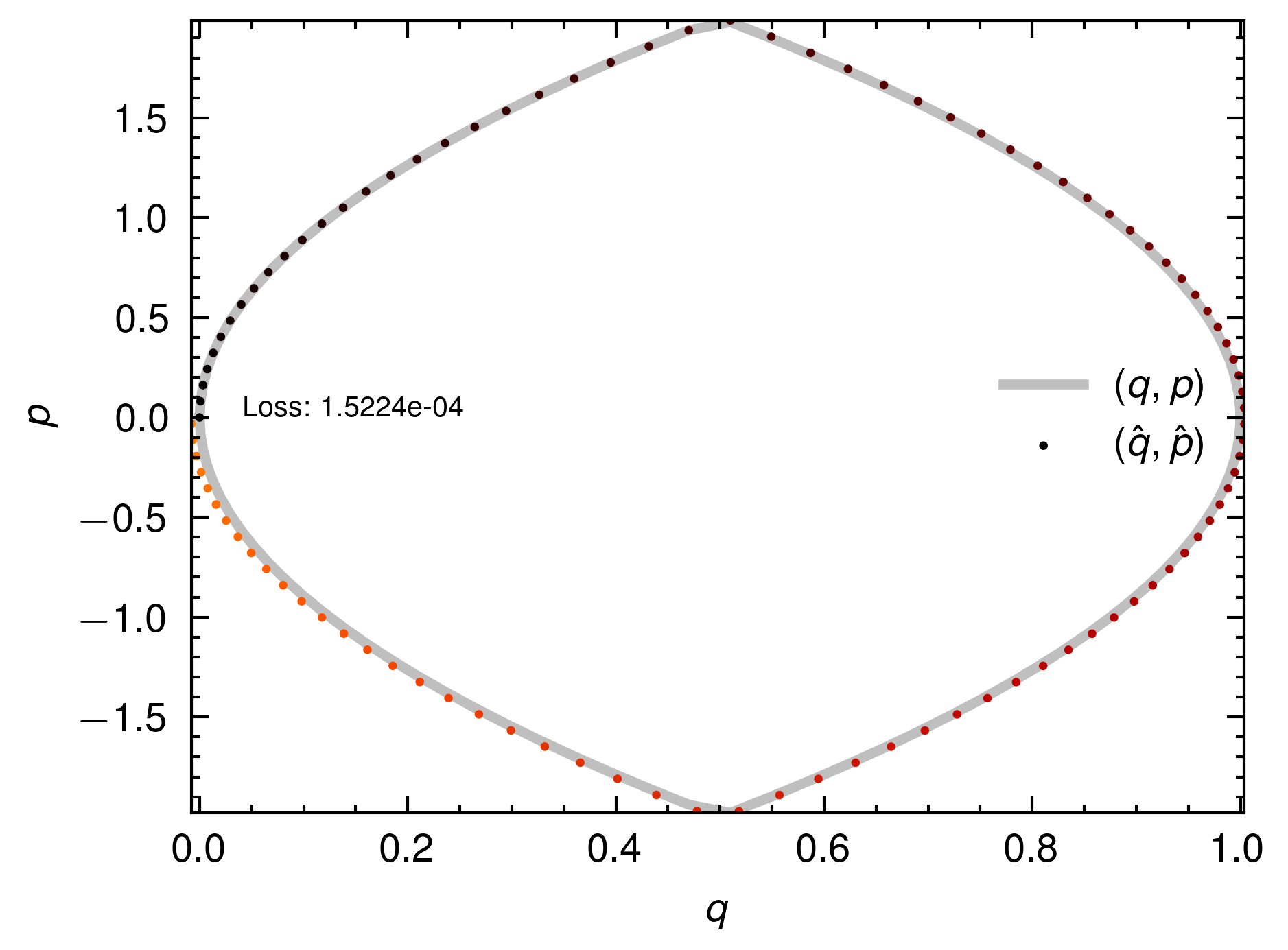}
            \caption{RK4}
        \end{subfigure}
        \caption{Phase plots for the Mirrored Free Fall potential. The true solution is shown in gray line, while the prediction is in scatter points. The plots should be read clockwise starting from (0.0, 0.0).}
        \label{fig:mff_phase_plots}
    \end{adjustwidth}
\end{figure}

Figure \ref{fig:mff_phase_plots} presents the phase plots for both MambONet and RK4 on the MFF potential, revealing intriguing differences in their error propagation characteristics. Both models struggle with this extrapolation task, showing decreased performance compared to their results on other potentials.

MambONet's predictions deviate considerably from true values near the non-differentiable point at $q=0.5$, unsurprising given that the MFF potential lies outside the training data's function space.
Remarkably, after this high-error point, MambONet's predictions ``recover,'' closely aligning with the true solution.
This suggests that errors at one time point don't necessarily impact subsequent predictions' accuracy.

Conversely, RK4 initially handles the non-differentiable point with less deviation but shows gradual error accumulation over time. From approximately (0.5, -2.0), the RK4 solution deviates significantly from the true trajectory, with this deviation growing continuously.

This contrast highlights a fundamental difference between operator learning approaches and traditional numerical methods. Operator networks like MambONet treat each time point independently, inherently preventing error propagation between time points. While this doesn't guarantee accuracy at every point, it means that errors don't necessarily compound over time.

Iterative methods like RK4, while potentially more accurate for short time intervals or well-behaved functions, are susceptible to error accumulation over longer time scales or in the presence of challenging features. Each step builds upon previous results, allowing small errors to compound.

These observations have significant implications for long-term predictions in dynamical systems and suggest several directions for future research:

\begin{enumerate}
    \item Investigating hybrid approaches combining strengths of operator learning and traditional numerical methods.
    \item Developing training strategies to improve extrapolation capabilities of operator learning models.
    \item Exploring operator learning methods in multi-scale simulations.
    \item Studying theoretical foundations of error behavior in operator learning models.
\end{enumerate}

In conclusion, the distinct error propagation characteristics of operator learning methods highlight both strengths and challenges in solving differential equations. While promising in preventing cumulative errors, their performance can be highly sensitive to input functions. Understanding and leveraging these properties will be crucial in expanding operator learning's applicability to a wider range of problems in computational physics and beyond.

\subsubsection{Comparison with Hamiltonian Neural Networks}
While both Hamiltonian Neural Networks (HNNs) \citep{greydanus2019hamiltonian} and our proposed Neural Hamilton method connect Hamiltonian mechanics with artificial intelligence, they are fundamentally opposite in their approaches upon closer examination. HNNs are designed to infer the Hamiltonian function from observed trajectories of dynamical systems that share the same form of Hamiltonian function (even if the specific values differ). Given an initial condition, HNNs learn the Hamiltonian and use Hamilton's equations to compute the time derivatives of $q$ and $p$. These derivatives are then integrated using an ODE solver to obtain the trajectories $q(t)$ and $p(t)$.

\begin{table}
\caption{Comparison of Hamiltonian Neural Networks and Neural Hamilton}
\label{tab:hnn_vs_neuralhamilton}
\centering
\renewcommand{\arraystretch}{1.25}
\begin{tabular}{ccc}
\toprule
Object & HNN & Neural Hamilton \\
\midrule
$q_0, p_0$ & Input & Given \\
$H(q,p)$ & Learn & Input \\
$\dot{q} = \frac{\partial H}{\partial p}, \dot{p} = -\frac{\partial H}{\partial q}$ & Given & Learn \\
$q(t), p(t)$ & Output & Output \\
\bottomrule
\end{tabular}
\end{table}

In contrast, our Neural Hamilton method takes the Hamiltonian (specifically, the potential function) as an input along with the same initial conditions and directly outputs the trajectories $q(t)$ and $p(t)$ through operator learning, without relying on any explicit physical equations. This means that our network effectively learns Hamiltonian mechanics itself, as it can generate the solutions to the equations of motion without being explicitly programmed with Hamilton's equations.

The key differences between these approaches are summarized in Table \ref{tab:hnn_vs_neuralhamilton}.

This comparison highlights a significant divergence in how machine learning can be applied to physical systems. HNNs are particularly useful when the underlying Hamiltonian is unknown but trajectory data is available. They excel at inferring the Hamiltonian function and then using established physical laws (Hamilton's equations) to predict future states via numerical integration with an ODE solver.

On the other hand, our Neural Hamilton method is advantageous in scenarios where rapid solutions for various potential functions are required without the need to solve differential equations explicitly. By learning the mapping from potential functions to trajectories directly, our method bypasses the need for explicit physical equations, essentially internalizing the dynamics dictated by Hamiltonian mechanics through data-driven operator learning.

Furthermore, this contrast underscores a broader trend in the application of AI to physics: a shift from learning components of physical theories (like the Hamiltonian function in HNNs) to learning the entire process of solving physical equations (as in our Neural Hamilton method). The success of our approach in capturing the dynamics without explicit knowledge of Hamilton's equations suggests that neural networks can internalize deep physical principles purely from data. This raises intriguing questions about the nature of physical laws and the potential for AI to discover or represent such laws in novel and efficient ways.

Future research could explore hybrid approaches that combine the strengths of both paradigms. For instance, developing models that can both learn the Hamiltonian and directly map potential functions to trajectories could lead to more powerful and versatile tools in computational physics. Such physics-informed neural operators would offer both interpretability and generalization across different physical systems, potentially advancing our ability to model complex dynamical phenomena.

\subsubsection{Extending Neural Hamilton to Unbounded Potentials}

While the Neural Hamilton approach has demonstrated remarkable effectiveness in solving Hamilton's equations for bounded potentials, its application to unbounded potentials presents a unique challenge due to the mathematical constraints of operator learning. However, we propose a method to extend the applicability of Neural Hamilton to certain classes of unbounded potentials through a transformation technique.

Consider a monotonically decreasing potential $V(q)$ defined on $[0, Q]$, where $Q < 1$ and $V(0) = V_0$. To adapt this unbounded potential for use with Neural Hamilton, we introduce a $C^2$ function $P(q)$ that extends the potential over the interval $(Q, 1]$, satisfying the following conditions:
\begin{equation}
\begin{gathered}
P(1) = V_0 \\
P(Q) = V(Q) \\
P'(Q) = V'(Q) \\
P''(Q) = V''(Q) \\
P(q) < V_0 \quad \text{for } Q < q < 1
\end{gathered}
\end{equation}
We then define a new potential function $\tilde{V}(q)$ as:
\begin{equation}
\tilde{V}(q) = \begin{cases} 
V(q) & \text{if } 0 \leq q \leq Q \\
P(q) & \text{if } Q < q \leq 1
\end{cases}
\end{equation}
This transformed potential $\tilde{V}(q)$ satisfies the input requirements for Neural Hamilton, allowing us to compute trajectories for the original unbounded potential within its domain of interest.

To extract the relevant dynamics from the Neural Hamilton output, we determine the time $T$ up to which the trajectory follows the original potential $V(q)$.
Given the conservation of energy $H = p^2/2 + V(q) = V_0$, we can derive:
\begin{equation}
T = \int_0^Q \frac{dq}{\sqrt{2(V_0 - V(q))}}
\end{equation}
As an illustrative example, we consider a free fall potential:
\begin{equation}
V(q) = -4(q - 0.5), \quad (0 \leq q \leq 0.5)
\end{equation}
We extend this potential using a cubic function $P(q) = 32q^3 - 48q^2 + 20q - 2$ for $0.5 < q \leq 1$, ensuring $C^2$ continuity at $q = 0.5$ and the required boundary conditions.

\begin{figure}
    \begin{adjustwidth}{-0.5cm}{-0.5cm}
    \centering
    \begin{subfigure}[b]{0.35\textwidth}
        \includegraphics[width=\textwidth]{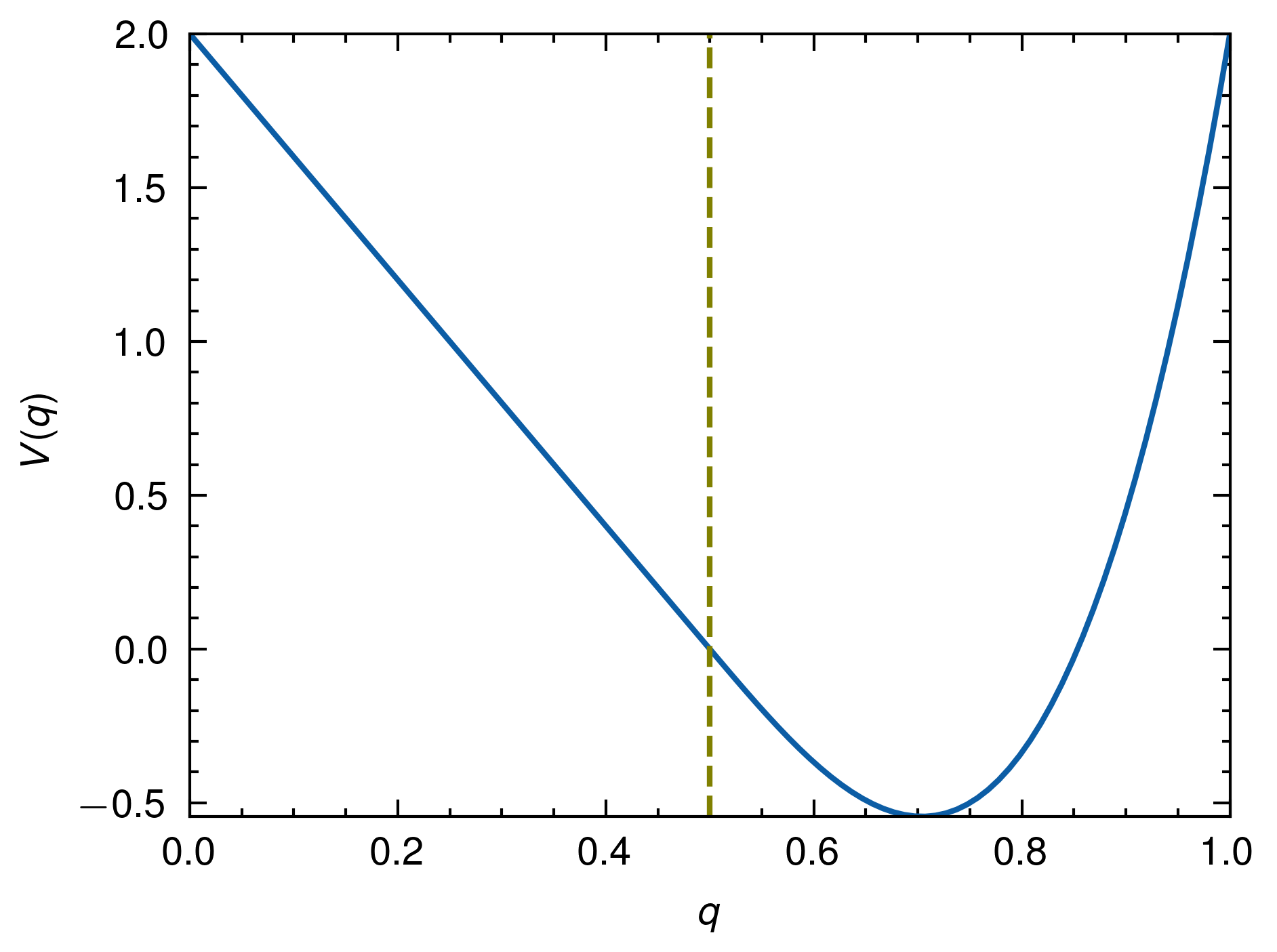}
        \caption{Extended potential}
    \end{subfigure}
    \hfill
    \begin{subfigure}[b]{0.35\textwidth}
        \includegraphics[width=\textwidth]{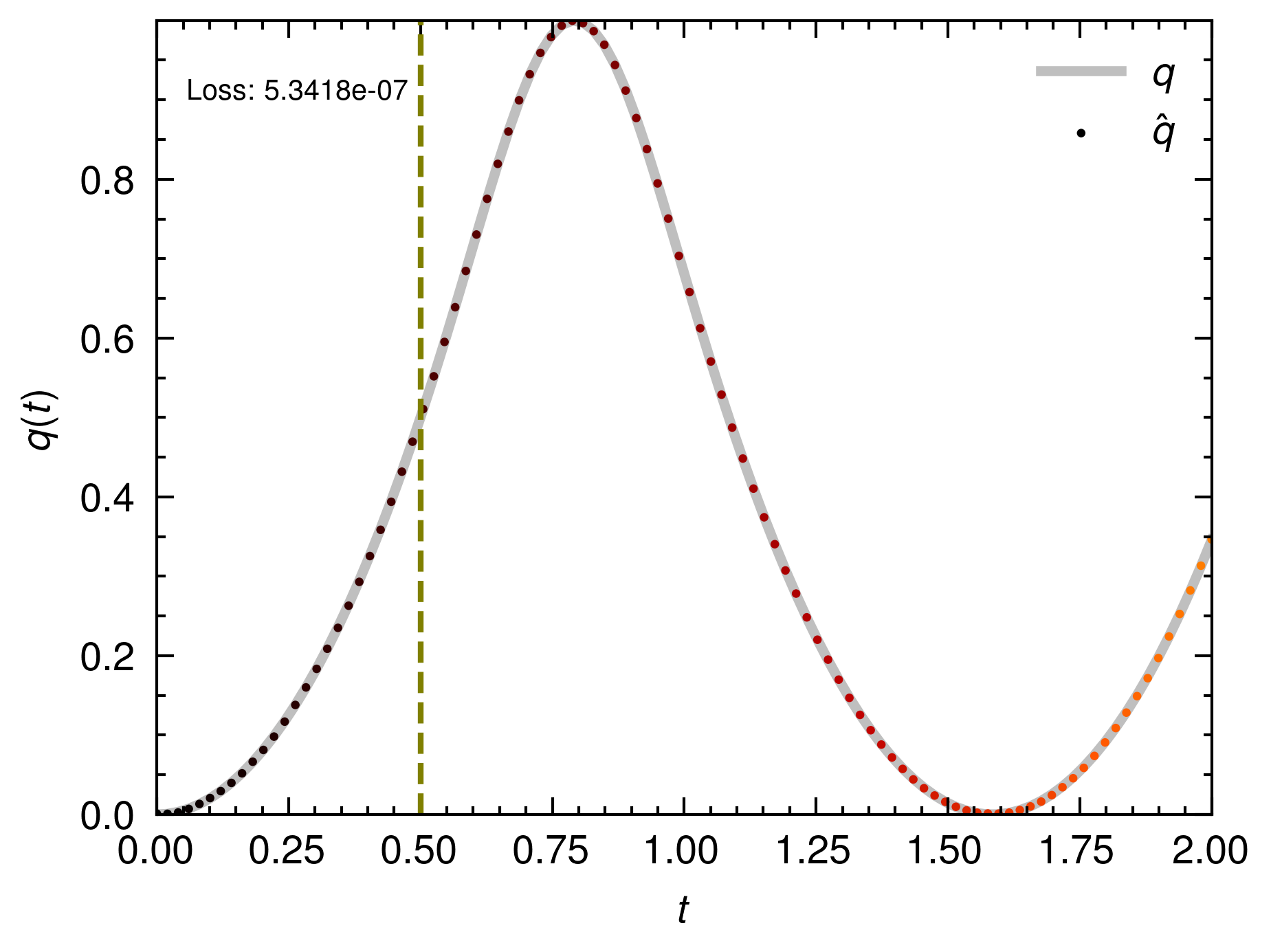}
        \caption{Position $q(t)$}
    \end{subfigure}
    \hfill
    \begin{subfigure}[b]{0.35\textwidth}
        \includegraphics[width=\textwidth]{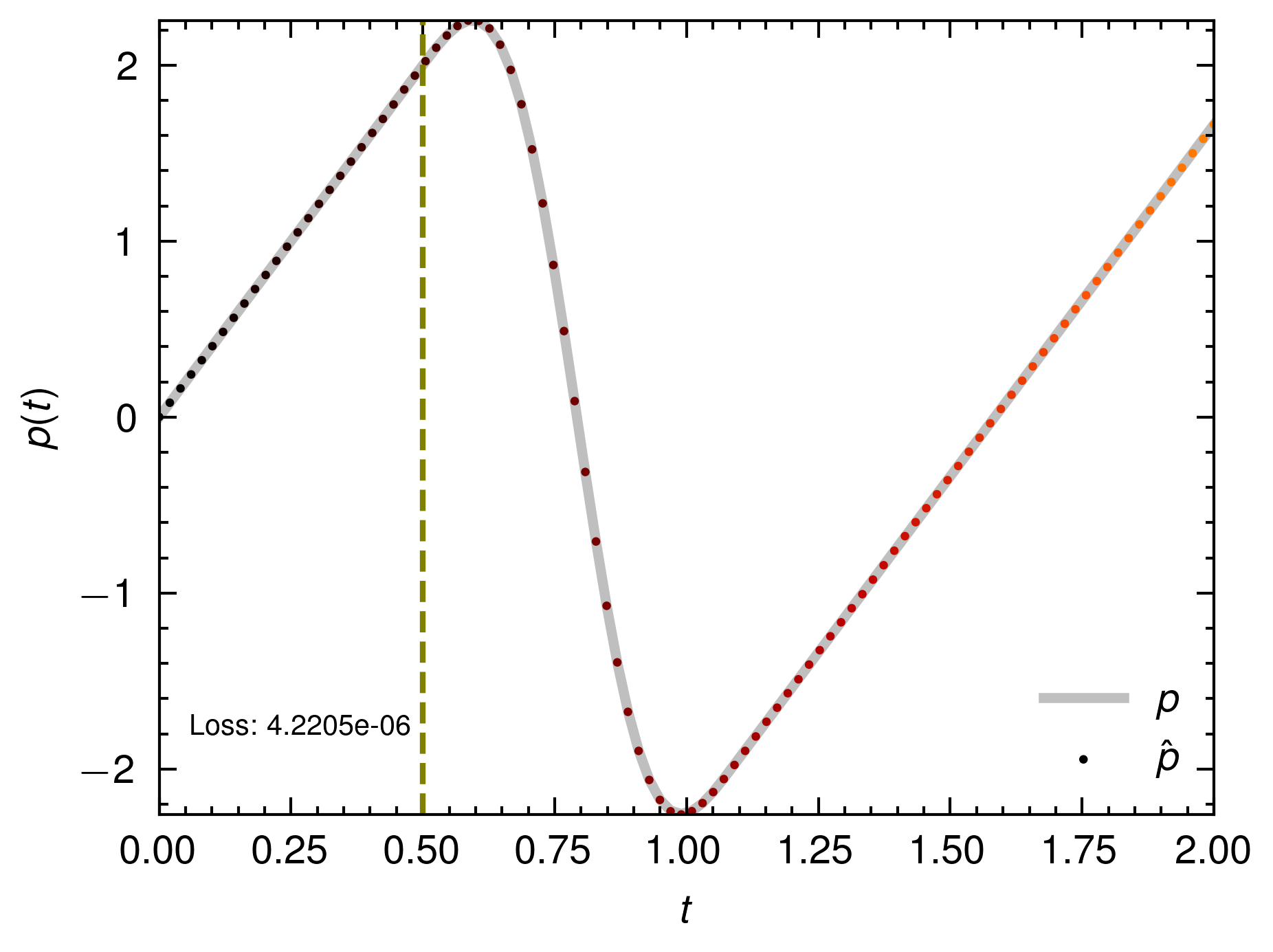}
        \caption{Momentum $p(t)$}
    \end{subfigure}
        \caption{(a) Extended potential function, with the olive dashed line at $q=0.5$ marking the boundary between the original free fall potential ($q \leq 0.5$) and the extended cubic potential ($q > 0.5$),
        (b) Position trajectory $q(t)$ and (c) Momentum trajectory $p(t)$ for the extended free fall potential, with the olive dashed line at $t=0.5$ separating the original ($t \leq 0.5$) and extended ($t > 0.5$) trajectory regions, as predicted by MambONet.}
    \label{fig:unbounded_results}
    \end{adjustwidth}
\end{figure}

Figures \ref{fig:unbounded_results} show the extended potentials, along with the resulting position and momentum trajectories from our trained MambONet.

For this example, we obtain $T = 0.5$. The trajectories for $0 \leq t \leq 0.5$ closely match the analytical solutions $q(t) = 2t^2$ and $p(t) = 4t$, demonstrating the effectiveness of our approach in handling unbounded potentials within their domain of interest.

This method extends the applicability of Neural Hamilton to a broader class of potentials, including unbounded ones, while maintaining the advantages of operator learning. Future work could focus on generalizing this approach to more complex unbounded potentials and multi-dimensional systems.

\section{Conclusion}

This study has demonstrated the capacity of neural networks to comprehend Hamiltonian mechanics through operator learning. Despite the constraints imposed by mathematical considerations, such as fixed initial conditions and a simplified kinetic term of $p^2/2$, we have shown that neural networks can grasp the operator relationship between potential functions and their corresponding position and momentum functions, without explicit equations. This achievement is underpinned by our mathematical proof that Hamilton's equations can be formulated as an operator learning problem, along with our novel algorithm for generating potential functions that satisfy mathematical constraints while facilitating effective learning.

Our research has revealed that employing architectures successful in language modeling, such as Transformers, Variational LSTMs, and Mamba models, significantly enhances the understanding of Hamiltonian dynamics compared to the standard DeepONet structure. When trained on sufficient data, these advanced models not only comprehend the dynamics but also outperform traditional numerical methods like RK4. The MambONet, in particular, exhibited superior performance across a wide range of potential functions, often surpassing other methods by a considerable margin.

Furthermore, we have shown that the Neural Hamilton methodology can be extended to certain unbounded potentials through appropriate transformations. While our proposed method is primarily applicable to monotonically decreasing or increasing potentials, it opens avenues for future research into handling more complex unbounded potentials. This extension broadens the applicability of our approach, though it acknowledges the need for case-specific strategies when dealing with highly intricate potentials.

It is important to note that this study serves as a proof of concept, focusing on one-dimensional potentials. Extending this work to higher dimensions presents challenges, particularly in preparing potential function data as the required function space expands significantly. Future research could explore methods to efficiently generate and process high-dimensional potential data, potentially leveraging advanced sampling techniques or dimensionality reduction methods.

In conclusion, our work not only advances the application of operator learning in physics but also bridges the gap between machine learning and fundamental physical principles. The success of our Neural Hamilton approach in capturing Hamiltonian dynamics without explicit equations suggests promising applications in various fields of physics and engineering. Future work could focus on extending these methods to more complex systems, including multi-dimensional and time-dependent Hamiltonians, as well as exploring the theoretical foundations of how neural networks internalize physical laws. This research opens up new possibilities for efficient simulation and analysis of complex physical systems, potentially revolutionizing our approach to computational physics and dynamical systems modeling.

\subsubsection*{Acknowledgments}
We thank Jinheung Kim, Donghun Yi, Soojin Lee, and Yeji Park for their insightful discussions and valuable feedback that helped shape this research.

\bibliography{ref}
\bibliographystyle{abbrvnat}

\appendix
\section{Proof of Theorems}

\subsection{Proof of Theorem 2}
\label{app:proof2}

We begin by restating Theorem 2:

\begin{numtheorem}{2}
For any initial condition $x_0 \in \mathbb{R}^{2m}$, there exists a unique maximal solution $x(t) \in C^1(I, \mathbb{R}^{2m})$ to the equation
\[
\dot{x} = \mathbf{J} \nabla H(x), \quad x(0) = x_0
\]
where $I \subset \mathbb{R}$ is the maximal interval of existence.
\end{numtheorem}

\begin{proof}
    Consider the Hamiltonian system $\dot{x} = \mathbf{J} \nabla H(x)$, $x(0) = x_0$, where $x = (q, p) \in \mathbb{R}^{2m}$, $\mathbf{J}$ is the constant symplectic matrix, and $H \in C^2(\mathbb{R}^{2m}, \mathbb{R})$. Let $f(x) = \mathbf{J} \nabla H(x)$.
    
    The function $f(x)$ is continuous in $x$ since $H \in C^2$ and $\mathbf{J}$ is constant.
    To establish local Lipschitz continuity of $f(x)$, consider a compact set $K \subset \mathbb{R}^{2m}$ containing $x_0$.
    The Hessian matrix $\mathbf{H}(x)$ is continuous on $K$, and by the Extreme Value Theorem \citep{rudin1976principles}, there exists $L_K > 0$ such that $\|\mathbf{H}(x)\| \leq L_K$ for all $x \in K$.
    By the Mean Value Theorem for vector-valued functions \citep{rudin1976principles}, for any $x_1, x_2 \in K$:
    \[
    \|\nabla H(x_1) - \nabla H(x_2)\| \leq L_K \|x_1 - x_2\|.
    \]
    
    Thus, $\nabla H(x)$ and consequently $f(x)$ are Lipschitz continuous on $K$.
    
    Given that $f(x)$ is continuous and locally Lipschitz continuous in $x$, the Picard-Lindelöf theorem guarantees the existence and uniqueness of a local solution $x(t)$ on some interval $[t_0 - \delta, t_0 + \delta]$, $\delta > 0$.
    
    This local solution can be extended to a maximal interval $I \subset \mathbb{R}$. The solution cannot be extended beyond $I$ if either $t$ approaches the endpoints of $I$ where $x(t)$ leaves every compact subset of $\mathbb{R}^{2m}$ (i.e., $\|x(t)\| \to \infty$), or if $f(x)$ ceases to be Lipschitz continuous. In this case, since $f(x)$ is defined and locally Lipschitz continuous everywhere in $\mathbb{R}^{2m}$, the only obstruction to extending the solution is if $\|x(t)\|$ becomes unbounded in finite time.
    
    Therefore, by the Picard-Lindelöf theorem and standard extension principles for ODEs \citep{coddington1956theory}, there exists a unique maximal solution $x(t) \in C^1(I, \mathbb{R}^{2m})$ to the Hamiltonian system, where $I$ is the maximal interval of existence containing $t = 0$.
\end{proof}

\textbf{Remarks:} The solution's maximal interval $I$ depends on the behavior of $H(x)$ at infinity. Without additional growth conditions on $H$, global existence (i.e., $I = \mathbb{R}$) cannot be guaranteed. The Picard-Lindelöf theorem also ensures continuous dependence of the solution on the initial condition $x_0$.

\subsection{Proof of Theorem 3}
\label{app:proof3}

We restate Theorem 3 as follows:

\begin{numtheorem}{3}
The mapping $\Phi: \mathcal{V} \to \mathcal{H}$ defined by
\[
\Phi(V)(q, p) = \sum_{i=1}^m \frac{p_i^2}{2 m_i} + V(q)
\]
is well-defined and injective, where $\mathcal{V} = C^2(\mathbb{R}^m, \mathbb{R})$ and $\mathcal{H} = C^2(\mathbb{R}^{2m}, \mathbb{R})$.
\end{numtheorem}

\begin{proof}
    We prove this theorem in two parts: first, we show that $\Phi$ is well-defined, and then we prove its injectivity.

    To show that $\Phi$ is well-defined, we need to verify that for each $V \in \mathcal{V}$, $\Phi(V) \in \mathcal{H}$.
    Consider $\Phi(V)(q, p) = T(p) + V(q)$, where $T(p) = \sum_{i=1}^m \frac{p_i^2}{2 m_i}$ is the kinetic energy term.
    $T(p)$ is $C^\infty$ in $(q, p)$ as it's a polynomial in $p$ and independent of $q$.
    $V(q)$ is $C^2$ in $q$ by definition of $\mathcal{V}$ and independent of $p$.
    The sum of these functions, $\Phi(V)(q, p)$, is therefore $C^2$ in $(q, p)$.
    The mixed partial derivatives involving both $q$ and $p$ exist and are continuous (in fact, they are zero).
    Thus, $\Phi(V) \in C^2(\mathbb{R}^{2m}, \mathbb{R}) = \mathcal{H}$ for all $V \in \mathcal{V}$, establishing that $\Phi$ is well-defined.

    To prove injectivity, assume $\Phi(V_1) = \Phi(V_2)$ for some $V_1, V_2 \in \mathcal{V}$. Then for all $(q, p) \in \mathbb{R}^{2m}$:
    \[
    \sum_{i=1}^m \frac{p_i^2}{2 m_i} + V_1(q) = \sum_{i=1}^m \frac{p_i^2}{2 m_i} + V_2(q).
    \]
    
    Subtracting the kinetic energy term from both sides yields:
    \[
    V_1(q) = V_2(q), \quad \forall q \in \mathbb{R}^m.
    \]
    
    This equality holds independently of $p$, demonstrating that $V_1 = V_2$. Therefore, $\Phi$ is injective.
\end{proof}

\section{Gaussian Random Fields}
\label{app:grf}

Gaussian Random Fields (GRFs) are stochastic processes that generalize the concept of multivariate normal distribution to infinite-dimensional space. In essence, a GRF is a collection of random variables, any finite number of which have a joint Gaussian distribution. GRFs are particularly useful in modeling spatial or temporal phenomena with uncertainty.

\subsubsection{Definition}
In our context, we focus on one-dimensional GRFs. For a one-dimensional case, let $x = (x_1, \ldots, x_m) \in \mathbb{R}^m$. The corresponding GRF is defined as:
\begin{equation}
    \mathbf{X} = (X_1, \ldots, X_m), \quad \mu(\mathbf{X}) = 0, \quad \text{Cov}(X_i, X_j) = k(x_i, x_j),
\end{equation}
where $k(x_i, x_j)$ is a kernel function. We denote this GRF as $\mathcal{G}(0, k(x_i, x_j))$. The mean is set to zero for simplicity, but in general, GRFs can have non-zero mean functions.

\subsubsection{Kernel Function}
The kernel function, also known as the covariance function, plays a crucial role in determining the properties of the GRF. It defines how the random variables at different points are related to each other. In our study, we use the squared exponential kernel:
\begin{equation}
    k(x_i, x_j; l) = \exp\left(-\frac{\|x_i - x_j\|^2}{2l^2}\right)
\end{equation}
where $l$ is the length scale parameter that controls the smoothness of the field. A larger $l$ results in smoother fields with longer-range correlations, while a smaller $l$ produces more rapidly varying fields.

The choice of the squared exponential kernel is motivated by its smoothness properties and its ability to model a wide range of phenomena. However, other kernels such as the Matérn kernel or the rational quadratic kernel could also be used, depending on the specific requirements of the problem at hand.

\subsubsection{Generation of GRF Realizations}
To generate realizations of the GRF, we need to sample from the multivariate Gaussian distribution defined by the kernel. This process can be computationally intensive, especially for large domains. To address this challenge, we utilize Rugfield \cite{Rugfield}, a Rust library specifically designed for efficient generation of 1D Gaussian Random Fields. Rugfield implements advanced algorithms that allow us to generate realizations of the random field quickly and accurately, even for large numbers of points.

\subsubsection{Application in Potential Function Generation}
The use of GRFs in our potential function generation provides a flexible and theoretically grounded way to create diverse, smooth functions. By varying the length scale and the number of GRF samples, we can generate a wide range of potential shapes, which is crucial for creating a comprehensive training dataset for our Hamilton equation solver.

In our algorithm, we sample GRF values at specific points, which are then used as control points for the cubic B-spline interpolation. This approach allows us to generate smooth, continuous potential functions with varying characteristics, essential for training a robust operator learning model for solving Hamilton's equations.

\section{Cubic B-splines}
\label{app:cbspline}

While Gaussian Random Fields provide us with a method to generate random points, our goal is to create sufficiently expressive and smooth functions for training the Operator network. Cubic B-Splines offer an ideal solution, providing smooth interpolation between control points with beneficial properties for our potential function representation.

\subsubsection{Motivation}
Simple interpolation methods like Cubic splines or Cubic Hermite splines are not ideal for our purposes due to several limitations:

\begin{enumerate}
    \item Limited expressiveness: With a small number of given points, piecewise polynomial splines are limited to the expressiveness of polynomials, which may not cover a sufficiently wide range of function spaces.
    \item Continuity issues: Particularly, Hermite splines guarantee $C^1$ continuity but not $C^2$ continuity. This lack of smoothness in the second derivative can negatively impact the learning process when dealing with potentials.
    \item Representation of complex functions: We need a method that can represent a wide variety of smooth functions, even with a limited number of control points.
\end{enumerate}

\subsubsection{Cubic B-Spline Definition}
Cubic B-splines are piecewise polynomial functions of degree 3, providing smooth interpolation between control points. They offer $C^2$ continuity, meaning that the function, its first derivative, and its second derivative are all continuous.

The general form of a cubic B-spline is defined as:
\begin{equation}
    \mathbf{C}(x) = \sum_{i=0}^n B^3_i(x)\mathbf{P}_i
\end{equation}
where $\mathbf{P}_i$ are the control points, and $B^3_i(x)$ are the cubic B-spline basis functions.

\subsubsection{B-Spline Basis Functions}
The basis functions are defined using the Cox-de Boor recursion formula \cite{de2001practical}:
\begin{equation}
    B^0_i(x) = \begin{cases}
        1 & \text{if } t_i \leq x < t_{i+1} \\
        0 & \text{otherwise}
    \end{cases}
\end{equation}
\begin{equation}
    B^p_i(x) = \frac{x - t_i}{t_{i+p} - t_i}B^{p-1}_i(x) + \frac{t_{i+p+1} - x}{t_{i+p+1} - t_{i+1}}B^{p-1}_{i+1}(x)
\end{equation}
Here, $t_i$ are the knot points that define the intervals of the piecewise function, and $p$ represents the degree of the spline (in our case, $p = 3$ for cubic splines).

\subsubsection{Clamped Cubic B-Spline with Uniform Knots}
In our implementation, we use a specific type of cubic B-spline known as a clamped cubic B-spline with uniform knots. The term "clamped" refers to a specific boundary condition where the first and last control points coincide with the endpoints of the curve. This ensures that our spline exactly passes through these endpoints, which is crucial for defining our potential function with specific boundary conditions.

For a domain $[a, b]$ divided into $n$ intervals, the uniform knots are defined as:
\begin{equation}
    \lambda_i = \frac{i}{n}, \quad \text{for } i = 0, 1, \ldots, n
\end{equation}

\subsubsection{Advantages for Potential Function Representation}
The use of clamped cubic B-splines with uniform knots offers several advantages for our potential function representation:

\begin{enumerate}
    \item Enhanced expressiveness: B-Splines can represent a wide variety of smooth functions, even with a limited number of control points.
    \item Guaranteed smoothness: The $C^2$ continuity ensures smooth transitions in the potential and its derivatives, which is crucial for the learning process.
    \item Local control: Changes to one control point primarily affect the curve locally, allowing for fine-tuned adjustments.
    \item Efficient computation: B-splines can be evaluated efficiently using recursive algorithms.
    \item Precise endpoint control: The clamped condition ensures that our potential function starts and ends at specified values.
    \item Uniform representation: The use of uniform knots ensures equal treatment of the entire domain.
\end{enumerate}

\subsubsection{Implementation}
For the implementation of Cubic B-Spline interpolation, we utilize Peroxide \cite{Peroxide}, a Rust numeric library. Peroxide provides efficient and accurate algorithms for B-Spline computation, allowing us to generate smooth potential functions from the random points produced by our Gaussian Random Field.

In the context of our potential function generation algorithm, we use the clamped cubic B-spline with uniform knots to interpolate between the points generated by the Gaussian Random Field. This ensures that our potential function $V(x)$ starts and ends at the specified values ($V(0) = V(1) = 2$), while providing a smooth, continuous representation of the potential in between these points.

\section{Hyperparameter Optimization}
\label{app:hyperparams}

\subsubsection{Grid Search for Model Architecture}
The initial phase of our hyperparameter optimization process involved a grid search to determine the optimal architecture for each model. During this phase, we fixed the learning rate and scheduler parameters to isolate the effects of architectural changes. For TraONet, VaRONet, and MambONet, we set the initial learning rate to $10^{-3}$, while for DeepONet, we used $10^{-2}$. We employed the ExpHyperbolicLR scheduler \cite{kim2024hyperboliclr} with an upper bound of 300, maximum iterations of 50, and infimum learning rates of $10^{-5}$ for DeepONet and $10^{-6}$ for the other models. The batch size was consistently set to 100 across all models.

\begin{table}
\caption{Grid Search Range and Results}
\label{tab:hyperparameter_optimization}
\centering
\renewcommand{\arraystretch}{1.25}
\begin{tabular}{lccc}
\toprule
Model & Parameter name & Candidates & Optimal value \\
\midrule
\multirow{3}{*}{DeepONet} & Hidden units & $\{128, 256, 512, 1024\}$ & $128$ \\
 & Hidden layers & $\{3, 4, 5, 6\}$ & $3$ \\
 & Branches & $\{10, 20, 30, 40\}$ & $10$ \\
\midrule
\multirow{4}{*}{TraONet} & Model dimension & $\{32, 64, 128\}$ & $64$ \\
 & Attention heads & $\{2, 4, 8\}$ & $8$ \\
 & Encoder/Decoder layers & $\{2, 3, 4\}$ & $3$ \\
 & Feed-forward dimension & $\{512, 1024, 2048\}$ & $512$ \\
\midrule
\multirow{3}{*}{VaRONet} & Hidden units & $\{64, 128, 256, 512\}$ & $512$ \\
 & LSTM layers & $\{2, 3, 4\}$ & $4$ \\
 & Latent dimension & $\{10, 20, 30, 40\}$ & $30$ \\
\midrule
\multirow{5}{*}{MambONet} & Model dimension & $\{32, 64, 128\}$ & $128$ \\
 & Mamba layers & $\{2, 3, 4\}$ & $4$ \\
 & Attention heads & $\{2, 4, 8\}$ & $4$ \\
 & Decoder layers & $\{2, 3, 4\}$ & $4$ \\
 & Feed-forward dimension & $\{256, 512, 1024\}$ & $1024$ \\
\bottomrule
\end{tabular}
\end{table}

We conducted the grid search over 50 epochs, evaluating performance using the average of five runs with different random seeds for each trial. This approach ensured robustness in our parameter selection process. Table \ref{tab:hyperparameter_optimization} presents the range of values considered for each architectural parameter and the optimal values identified through this process.

\subsubsection{Tree-structured Parzen Estimator Search for Learning Rate and Scheduler Parameters}
Following the optimization of model architectures, we employed the Tree-structured Parzen Estimator (TPE) method to fine-tune the remaining hyperparameters, including learning rate, scheduler parameters, and the KL weight for VaRONet. We utilized the optimal architectural configurations determined in the previous phase and conducted the TPE search over 50 epochs. As with the grid search, we averaged the performance over five runs with different random seeds for each trial, conducting 100 trials for each model.

\begin{table}
\caption{TPE Search Range and Optimal Values}
\label{tab:tpe_search_results}
\centering
\renewcommand{\arraystretch}{1.25}
\begin{tabular}{lccc}
\toprule
Model & Parameter name & Search range & Optimal value \\
\midrule
\multirow{3}{*}{DeepONet} & Initial learning rate & [$10^{-3}$, $10^{-1}$] & $7.3256 \times 10^{-3}$ \\
 & Upper bound & $\{300, 350, 400\}$ & $300$ \\
 & Infimum learning rate & [$10^{-6}$, $10^{-3}$] & $1.7369 \times 10^{-3}$ \\
\midrule
\multirow{3}{*}{TraONet} & Initial learning rate & [$10^{-4}$, $10^{-2}$] & $1.4645 \times 10^{-3}$ \\
 & Upper bound & $\{300, 350, 400\}$ & $300$ \\
 & Infimum learning rate & [$10^{-7}$, $10^{-4}$] & $4.5159 \times 10^{-5}$ \\
\midrule
\multirow{4}{*}{VaRONet} & KL weight & [$10^{-3}$, $10^{0}$] & $5.4963 \times 10^{-2}$ \\
 & Initial learning rate & [$10^{-4}$, $10^{-2}$] & $8.6721 \times 10^{-4}$ \\
 & Upper bound & $\{300, 350, 400\}$ & $300$ \\
 & Infimum learning rate & [$10^{-7}$, $10^{-4}$] & $6.2651 \times 10^{-6}$ \\
\midrule
\multirow{3}{*}{MambONet} & Initial learning rate & [$10^{-4}$, $10^{-2}$] & $9.7931 \times 10^{-4}$ \\
 & Upper bound & $\{300, 350, 400\}$ & $350$ \\
 & Infimum learning rate & [$10^{-7}$, $10^{-4}$] & $1.0310 \times 10^{-5}$ \\
\bottomrule
\end{tabular}
\end{table}

The search ranges and optimal values for these hyperparameters are detailed in Table \ref{tab:tpe_search_results}. This two-stage optimization process allowed us to efficiently explore the hyperparameter space while maintaining the computational feasibility of our study.

\subsubsection{Extended Epoch Training}
Upon determining the optimal hyperparameters through grid search and TPE, we proceeded to train our models for an extended period of 250 epochs. This approach was motivated by our use of hyperbolic-based learning rate schedulers, which we anticipated would maintain good performance even with increased epoch counts. The models trained using this extended epoch strategy were subsequently used for our final performance evaluations.

\subsubsection{Adaptation for Extended Dataset}
It is important to note that our initial hyperparameter optimization was conducted on the standard dataset comprising 10,000 potential functions. When applying these parameters to the extended dataset of 100,000 potentials, we encountered challenges, particularly with scheduler parameters.

While we expected the model architecture parameters to generalize well to the larger dataset, the increased number of steps per epoch (due to the consistent batch size) necessitated adjustments to the learning rate scheduler. This was particularly crucial for models like MambONet, which exhibited sensitivity to learning rates slightly higher than optimal.

To address this issue, we developed a technique to adapt the ExpHyperbolicLR scheduler for the extended dataset. Let $\eta_{N_1}(n)$ and $\eta_{N_2}(n)$ represent ExpHyperbolicLR schedulers with maximum epochs $N_1$ and $N_2$ respectively, where $N_1 \ll N_2$. Let $l(n)$ denote the asymptotic learning rate curve. In this context, $\eta_0$ represents the initial learning rate, and $\eta_\text{inf}$ represents the infimum learning rate, which is the lower bound that the learning rate approaches asymptotically.

The challenge arises because $\eta_{N_1}(N_1) < \eta_{N_2}(N_1) \approx l(N_1)$, which can lead to divergence in sensitive tasks. To mitigate this, we propose an adjusted ExpHyperbolicLR, $\tilde{\eta}(n)$, with a new asymptotic curve $\tilde{l}(n)$, such that:
\begin{equation}
\log \tilde{l}(N_1) = \frac{\log \eta_{N_1}(N_1) + \log l(N_1)}{2}
\end{equation}
To maintain the initial learning strategy, we keep the initial learning rate, upper bound, and maximum epochs consistent:
\begin{equation}
\tilde{\eta}_0 = \eta_0, \quad \tilde{U} = U, \quad \tilde{N} = N_1
\end{equation}
Given the asymptotic curve of ExpHyperbolicLR:
\begin{equation}
\log l(n) = \log \eta_0 - \frac{\log \eta_0 - \log \eta_{\inf}}{U} \times n
\end{equation}
We can derive the new infimum learning rate:
\begin{equation}
\log \tilde{\eta}_\text{inf} = \frac{1}{2}\left[\log\eta_0 + \log \eta_\text{inf} + \frac{U}{N_1}(\log \eta_{N_1}(N_1) - \log \eta_0)\right]
\end{equation}
This adjusted infimum learning rate was applied when training on the extended dataset, allowing us to maintain the benefits of our hyperparameter optimization while accommodating the increased data volume.

\newpage

\section{Test Potentials}
\label{app:test_potentials}

\vspace{-0.2cm}

\begin{figure}[h!]
    \begin{adjustwidth}{-0.5cm}{-0.5cm}
    \centering
    \begin{subfigure}[b]{0.35\textwidth}
        \includegraphics[width=\textwidth]{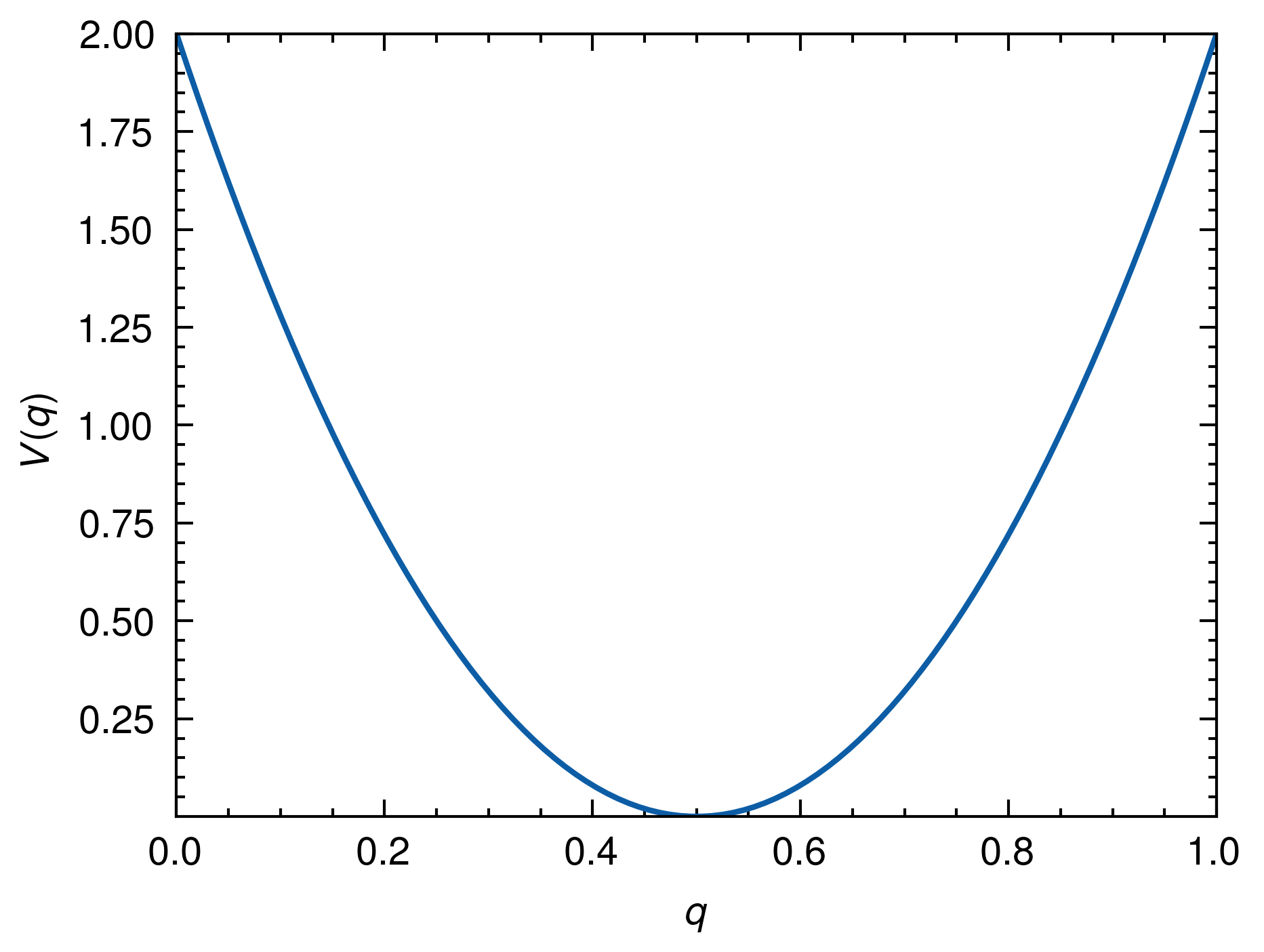}
    \end{subfigure}
    \hfill
    \begin{subfigure}[b]{0.35\textwidth}
        \includegraphics[width=\textwidth]{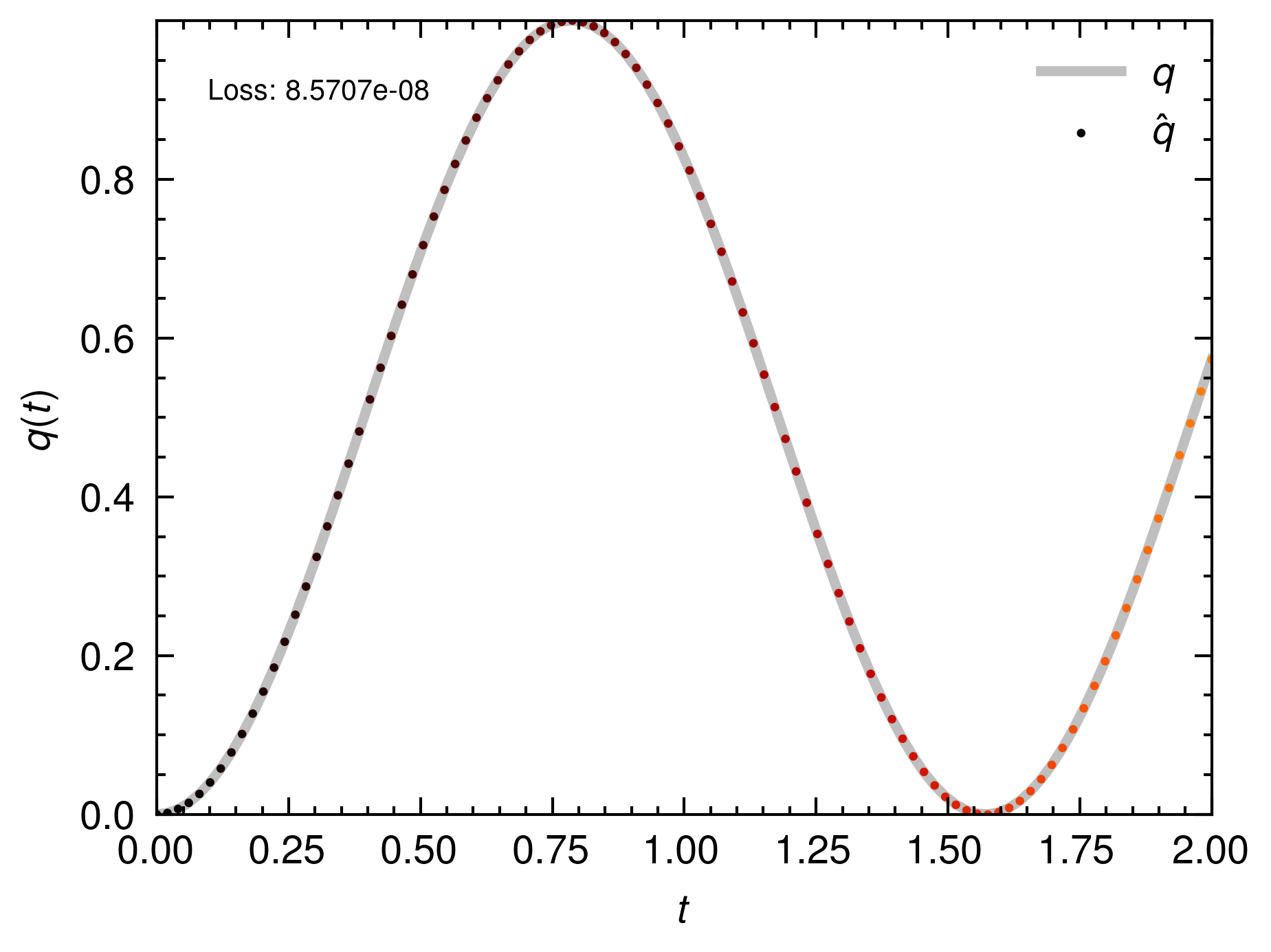}
    \end{subfigure}
    \hfill
    \begin{subfigure}[b]{0.35\textwidth}
        \includegraphics[width=\textwidth]{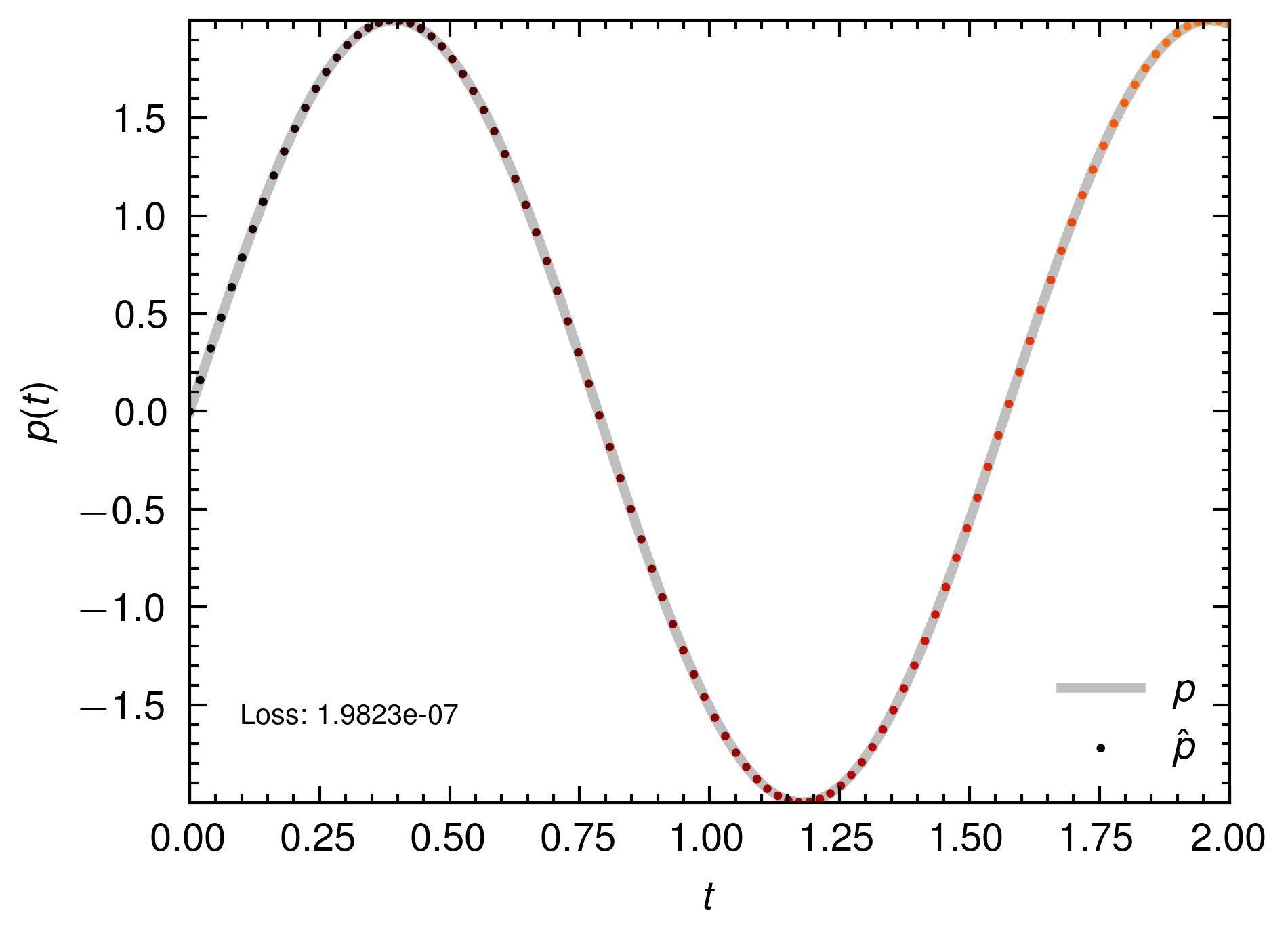}
    \end{subfigure}
    \begin{subfigure}[b]{0.35\textwidth}
        \includegraphics[width=\textwidth]{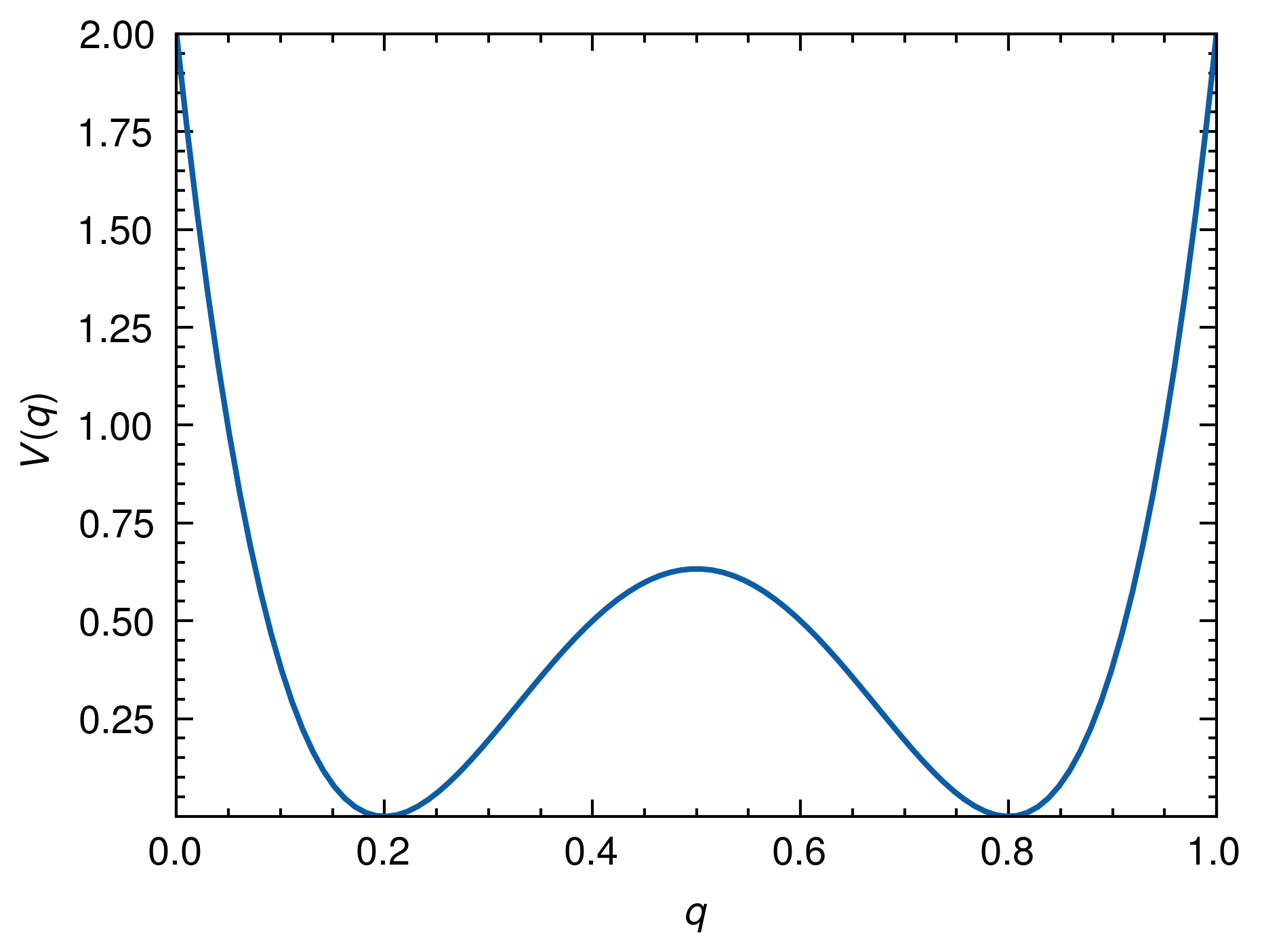}
    \end{subfigure}
    \hfill
    \begin{subfigure}[b]{0.35\textwidth}
        \includegraphics[width=\textwidth]{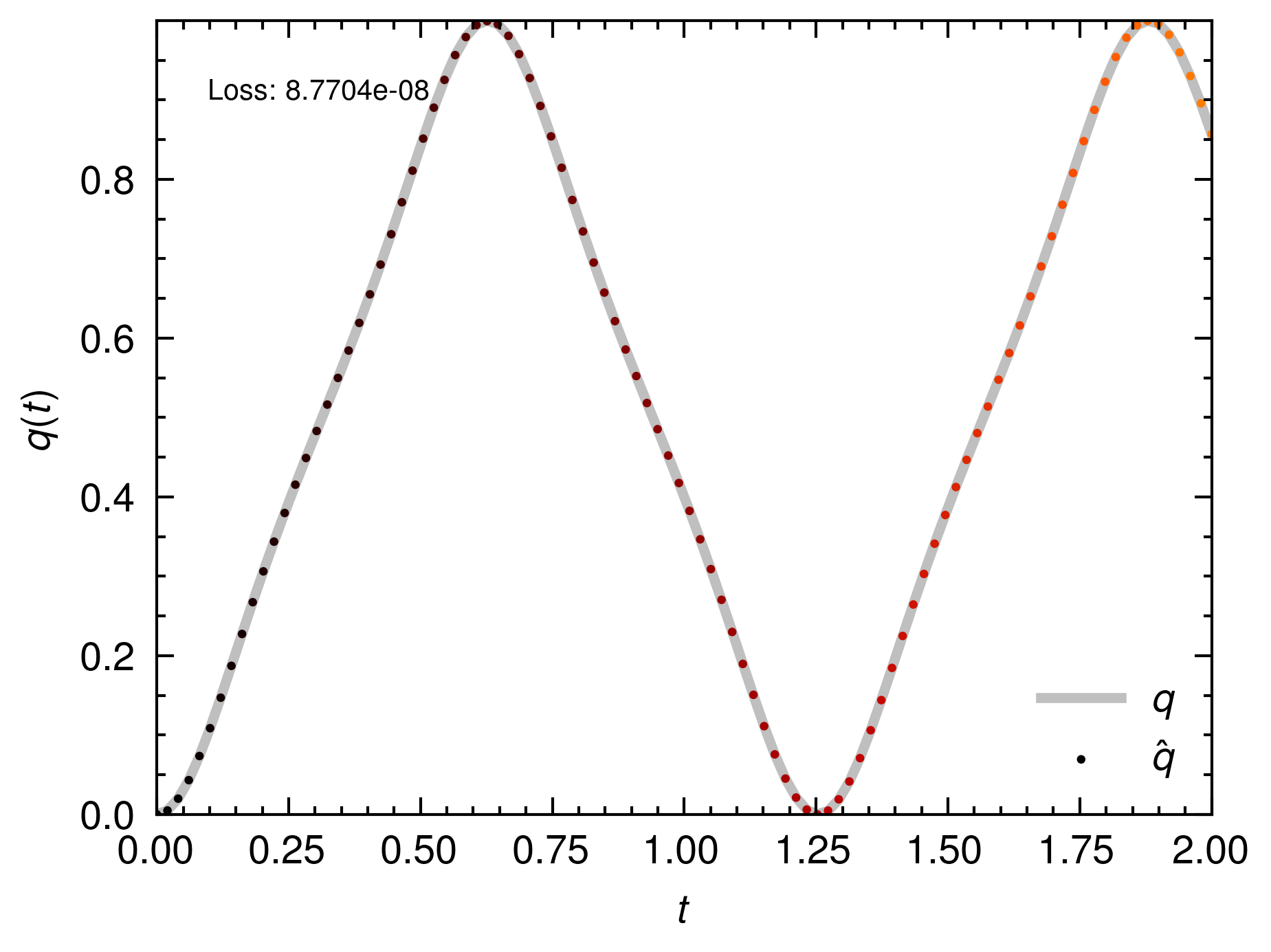}
    \end{subfigure}
    \hfill
    \begin{subfigure}[b]{0.35\textwidth}
        \includegraphics[width=\textwidth]{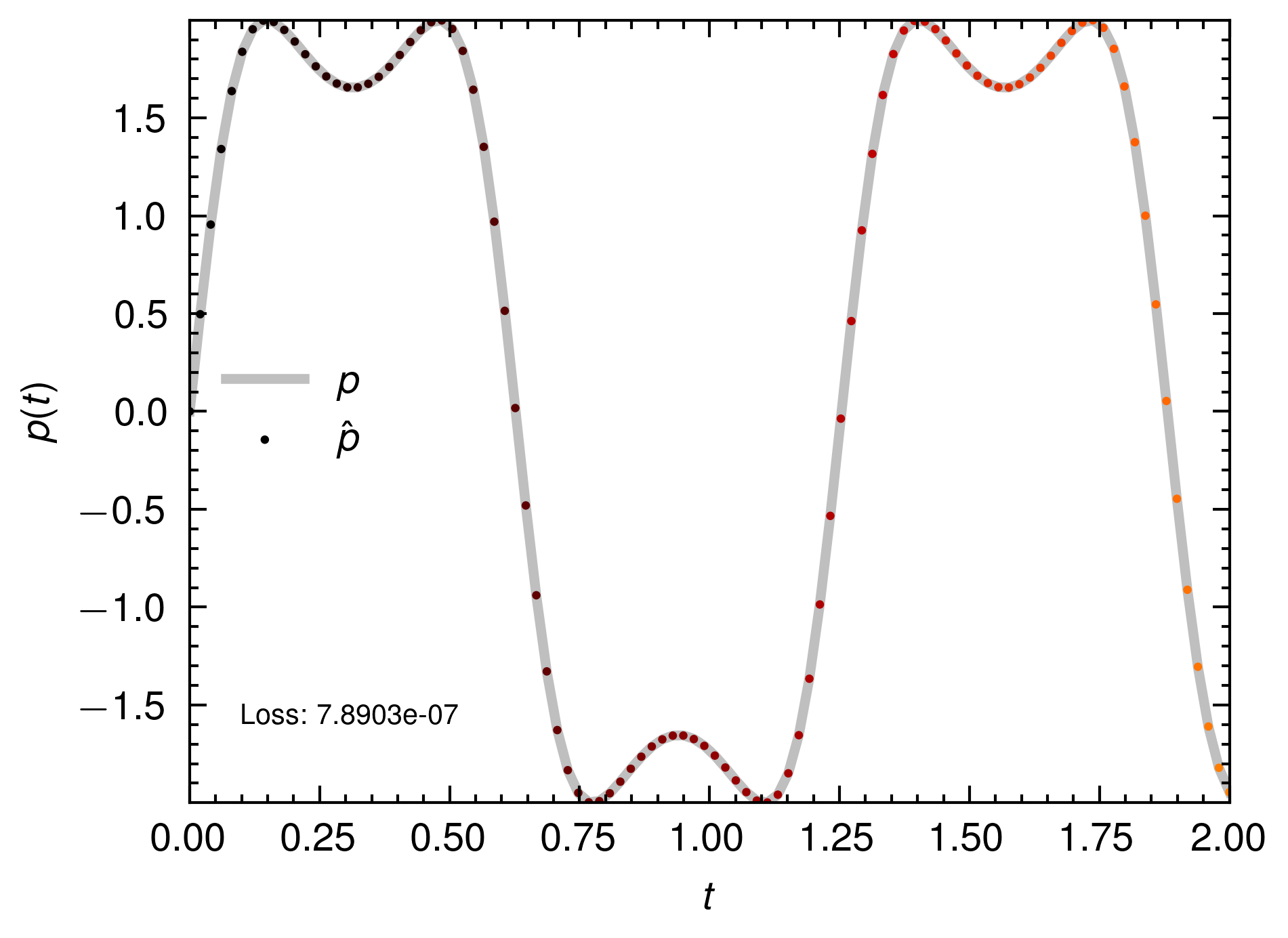}
    \end{subfigure}
    \begin{subfigure}[b]{0.35\textwidth}
        \includegraphics[width=\textwidth]{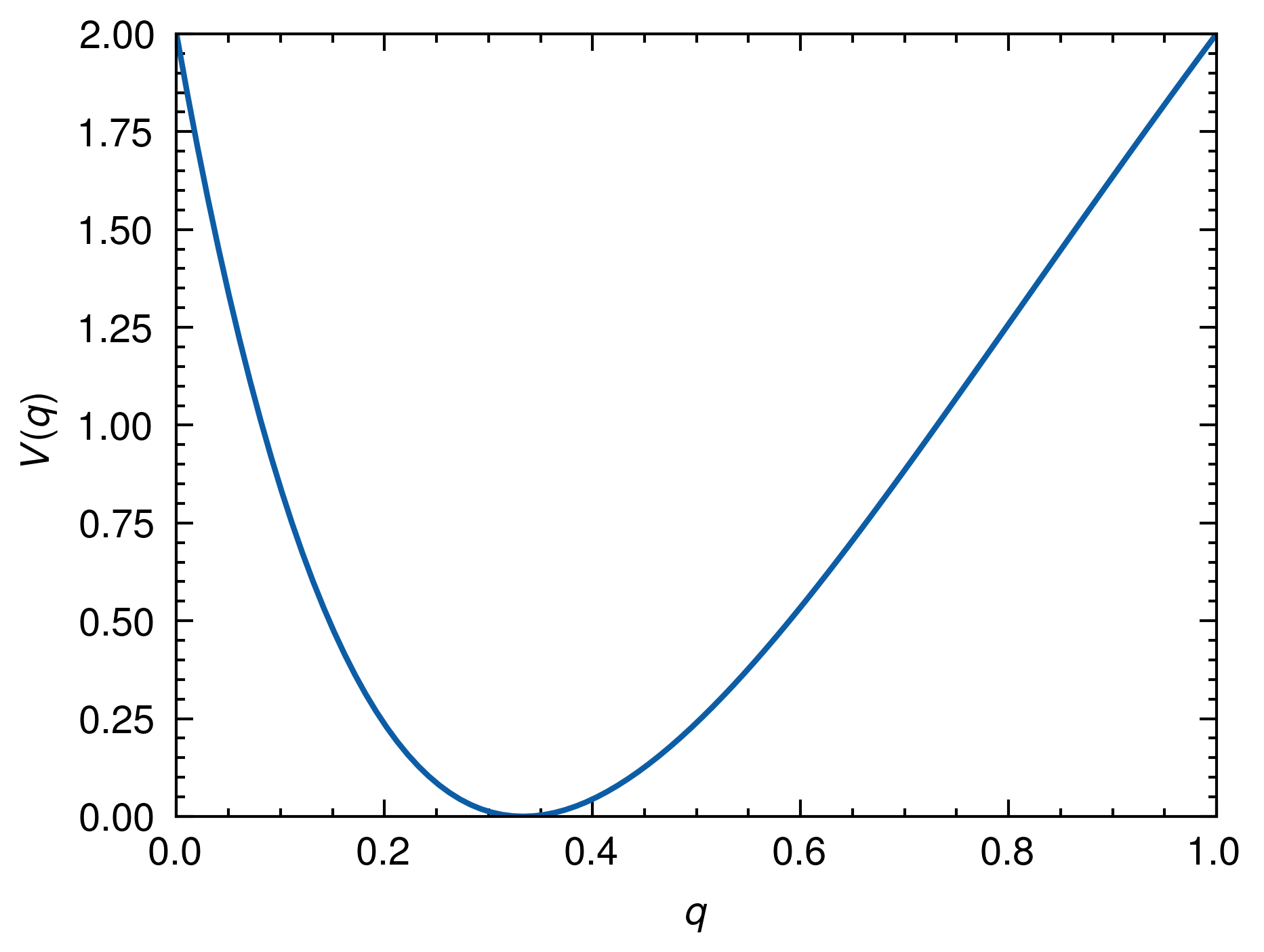}
    \end{subfigure}
    \hfill
    \begin{subfigure}[b]{0.35\textwidth}
        \includegraphics[width=\textwidth]{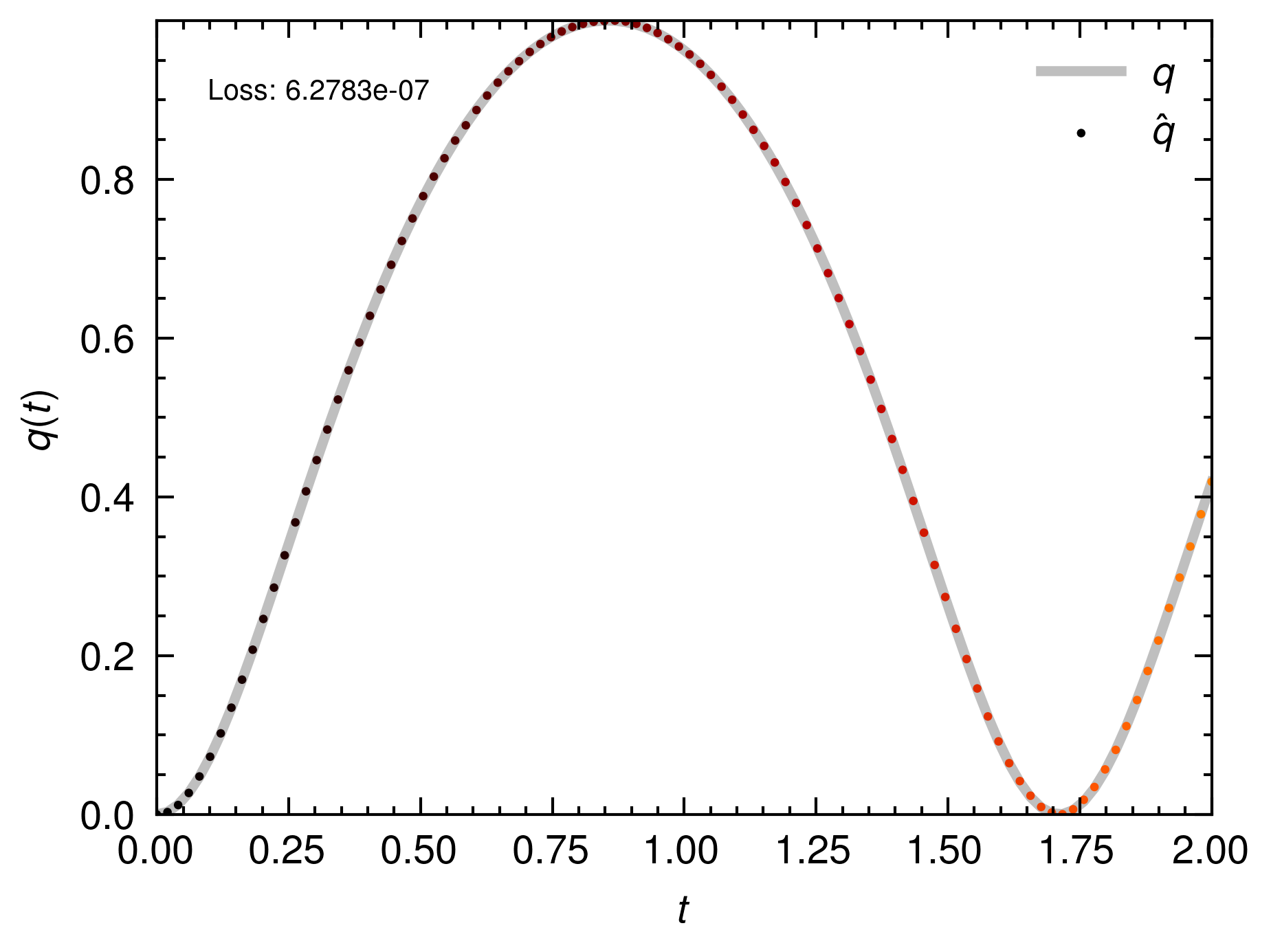}
    \end{subfigure}
    \hfill
    \begin{subfigure}[b]{0.35\textwidth}
        \includegraphics[width=\textwidth]{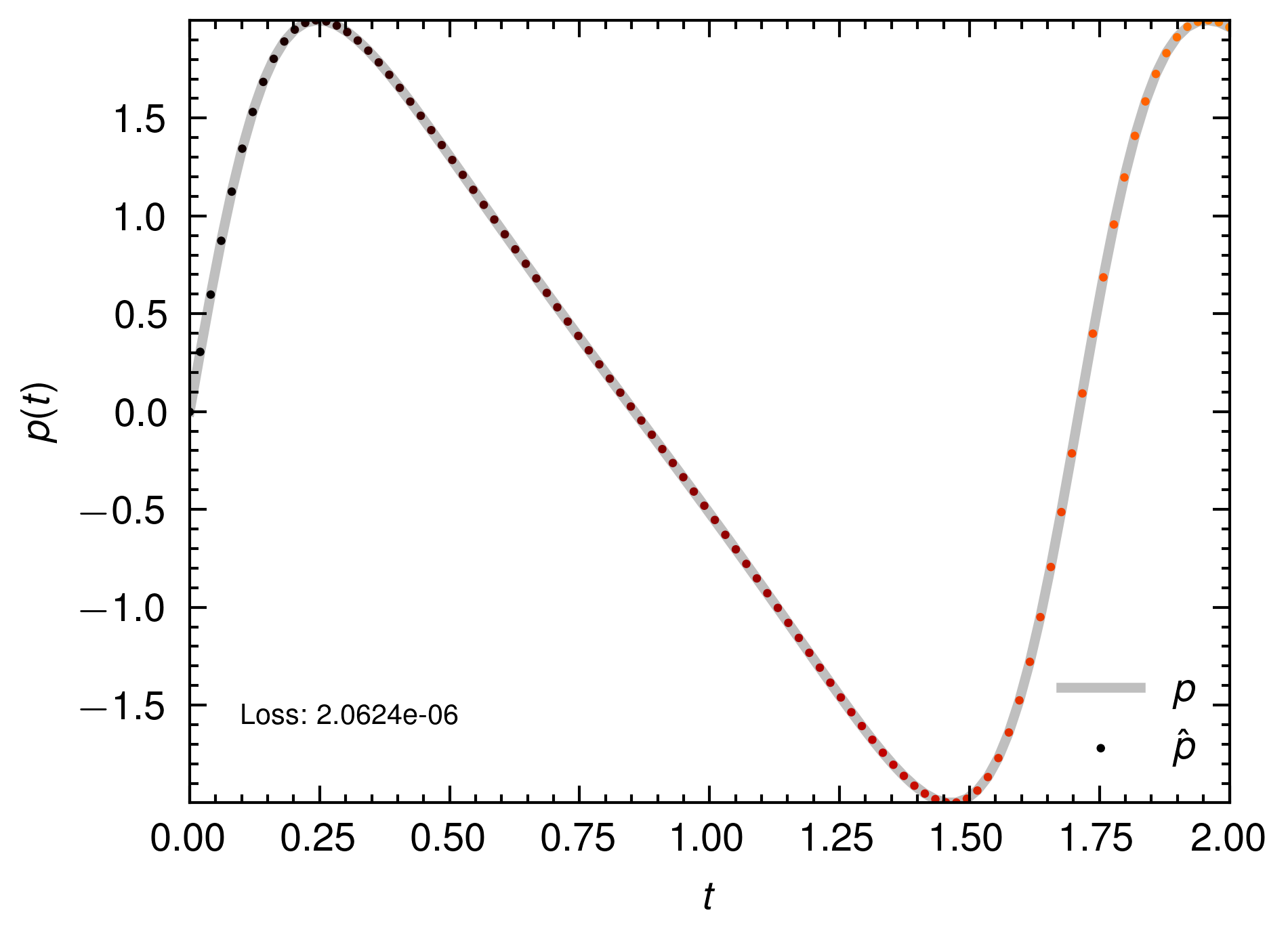}
    \end{subfigure}
    \begin{subfigure}[b]{0.35\textwidth}
        \includegraphics[width=\textwidth]{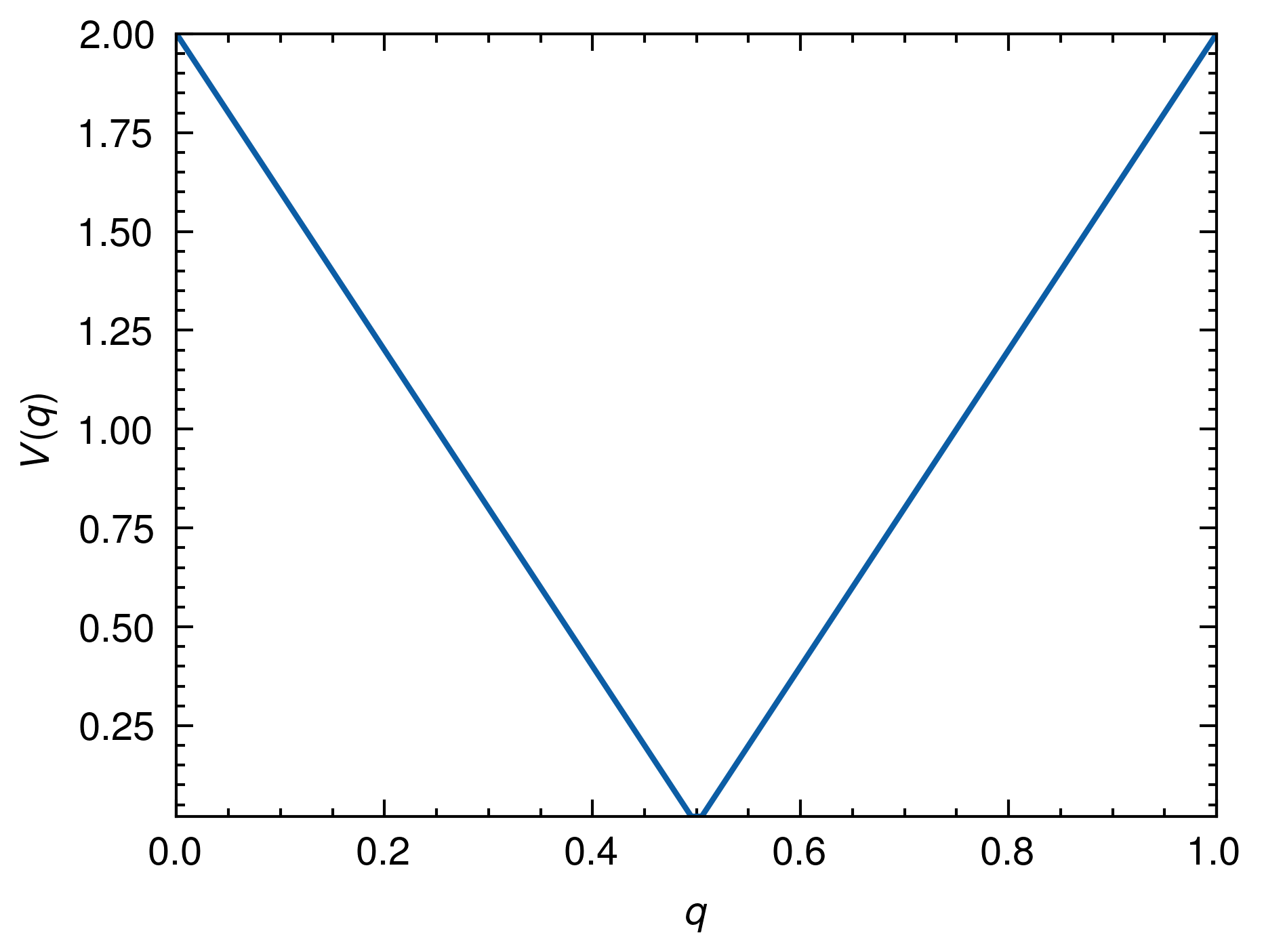}
    \end{subfigure}
    \hfill
    \begin{subfigure}[b]{0.35\textwidth}
        \includegraphics[width=\textwidth]{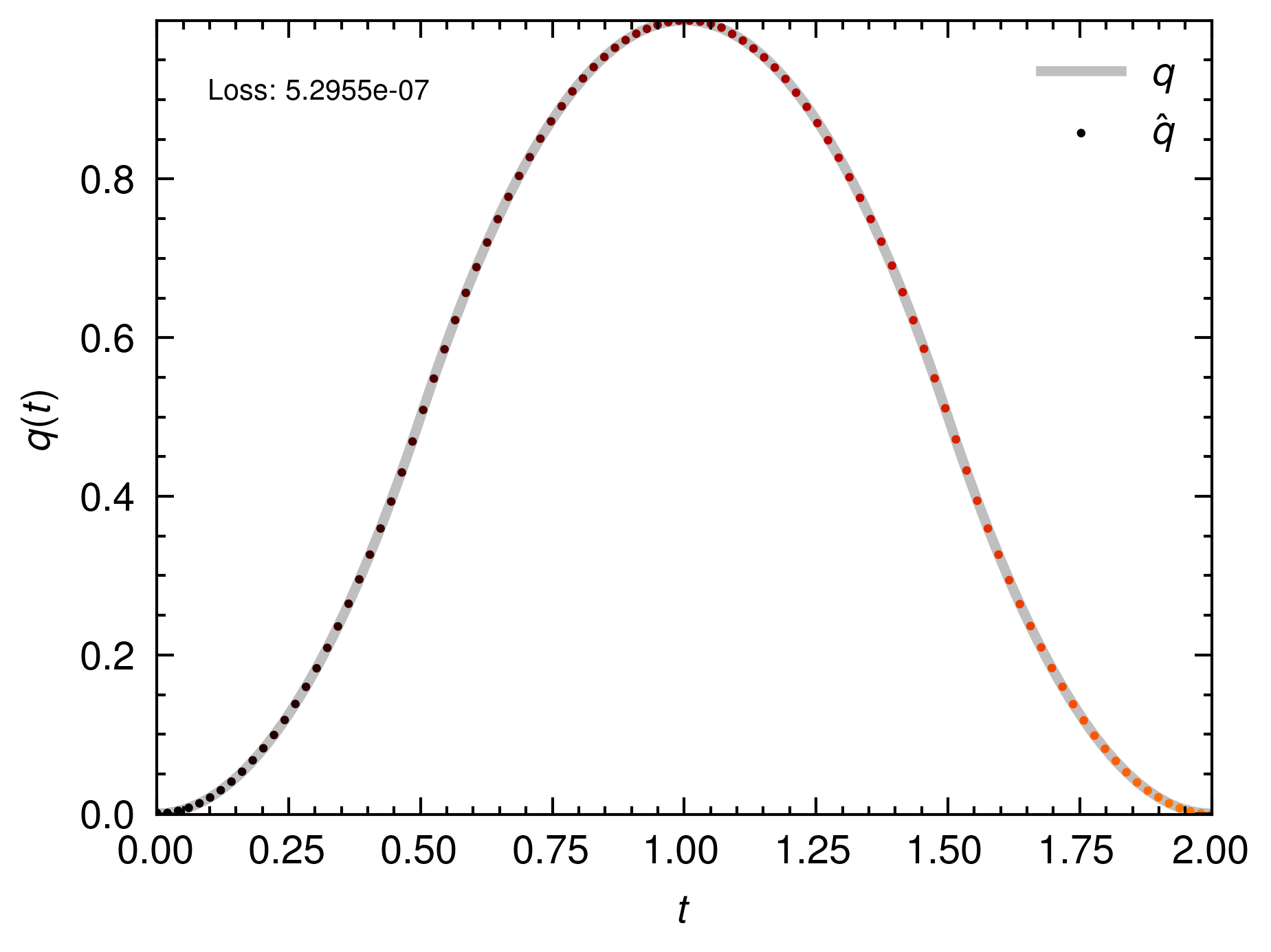}
    \end{subfigure}
    \hfill
    \begin{subfigure}[b]{0.35\textwidth}
        \includegraphics[width=\textwidth]{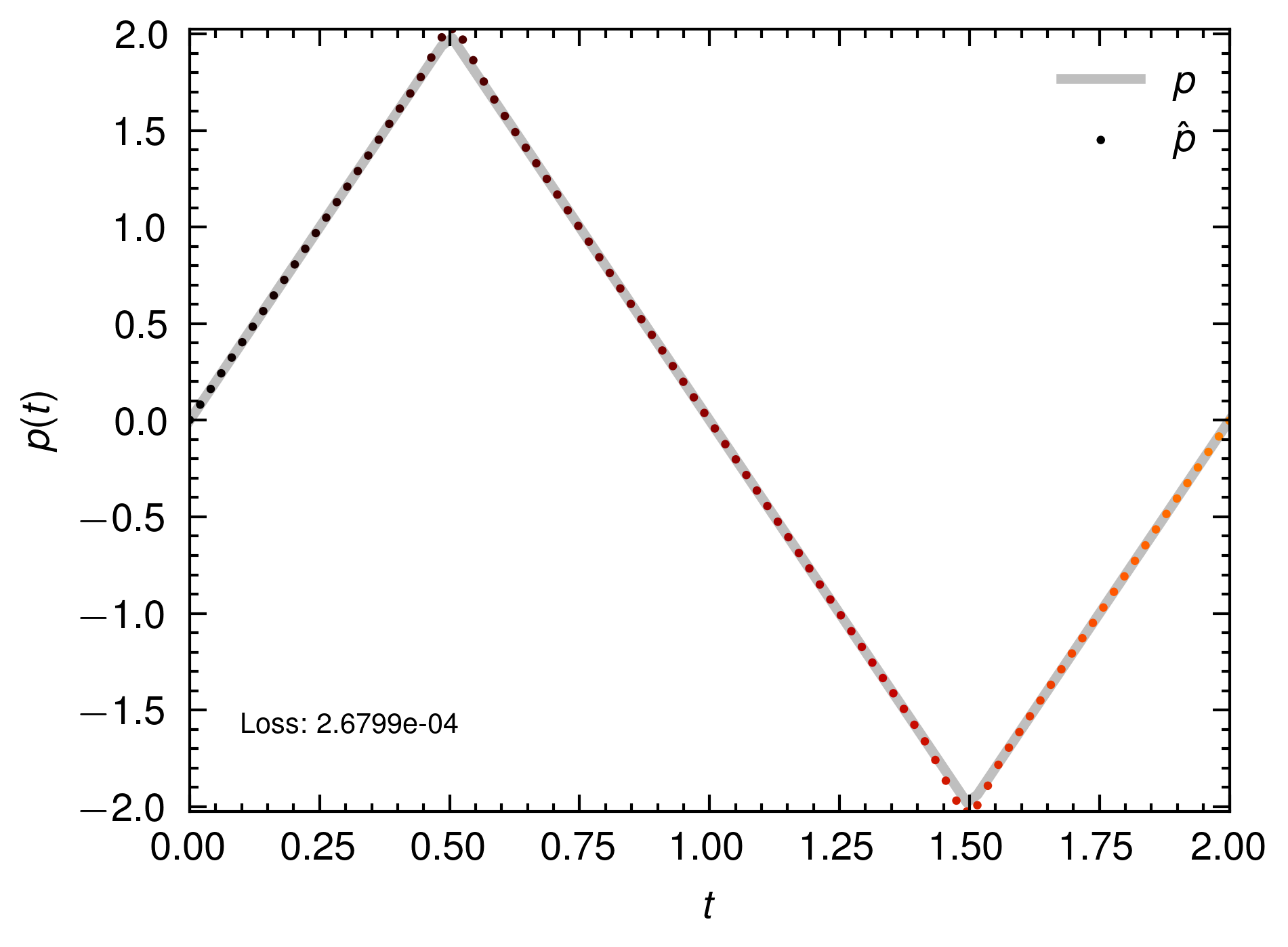}
    \end{subfigure}
    \begin{subfigure}[b]{0.35\textwidth}
        \includegraphics[width=\textwidth]{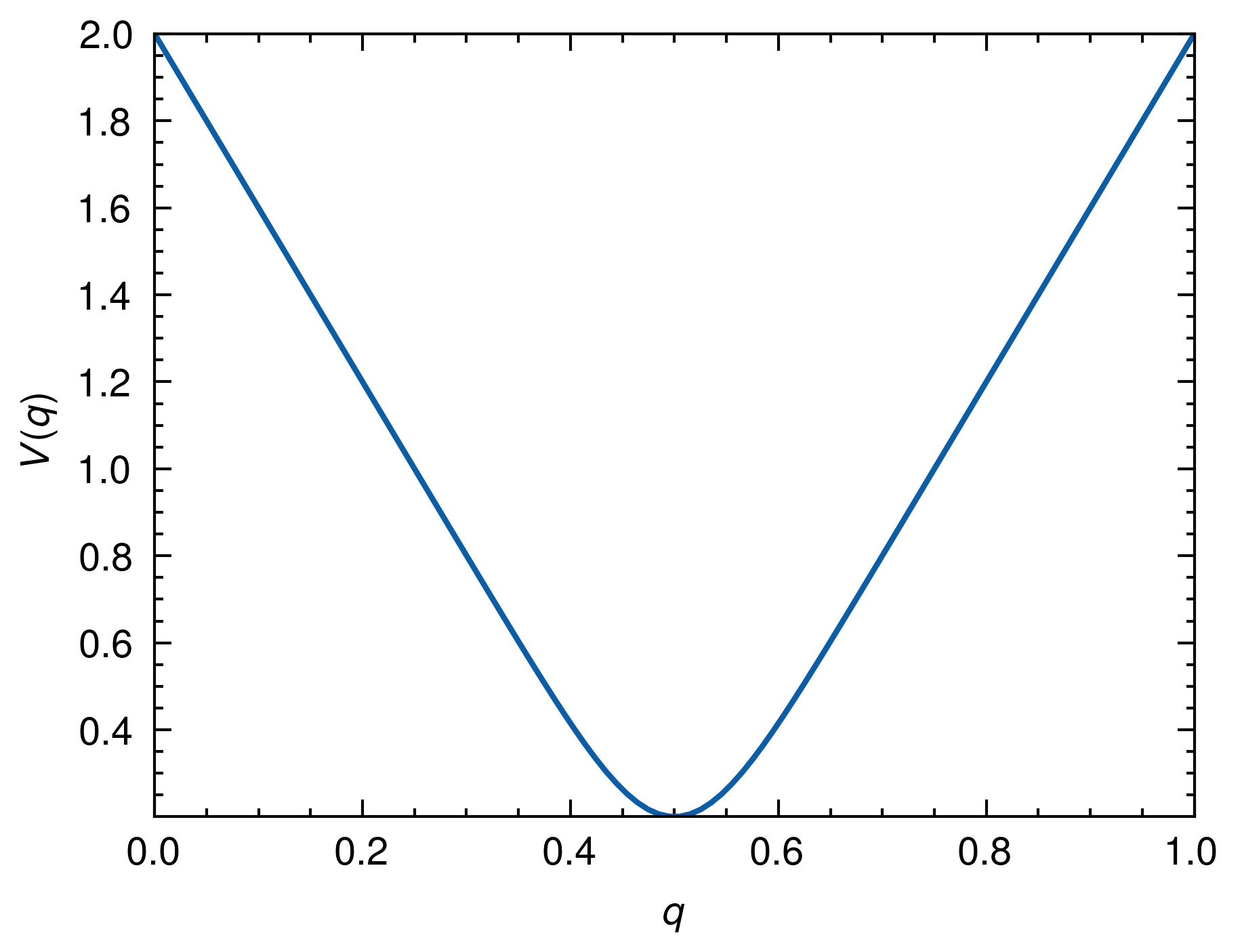}
    \end{subfigure}
    \hfill
    \begin{subfigure}[b]{0.35\textwidth}
        \includegraphics[width=\textwidth]{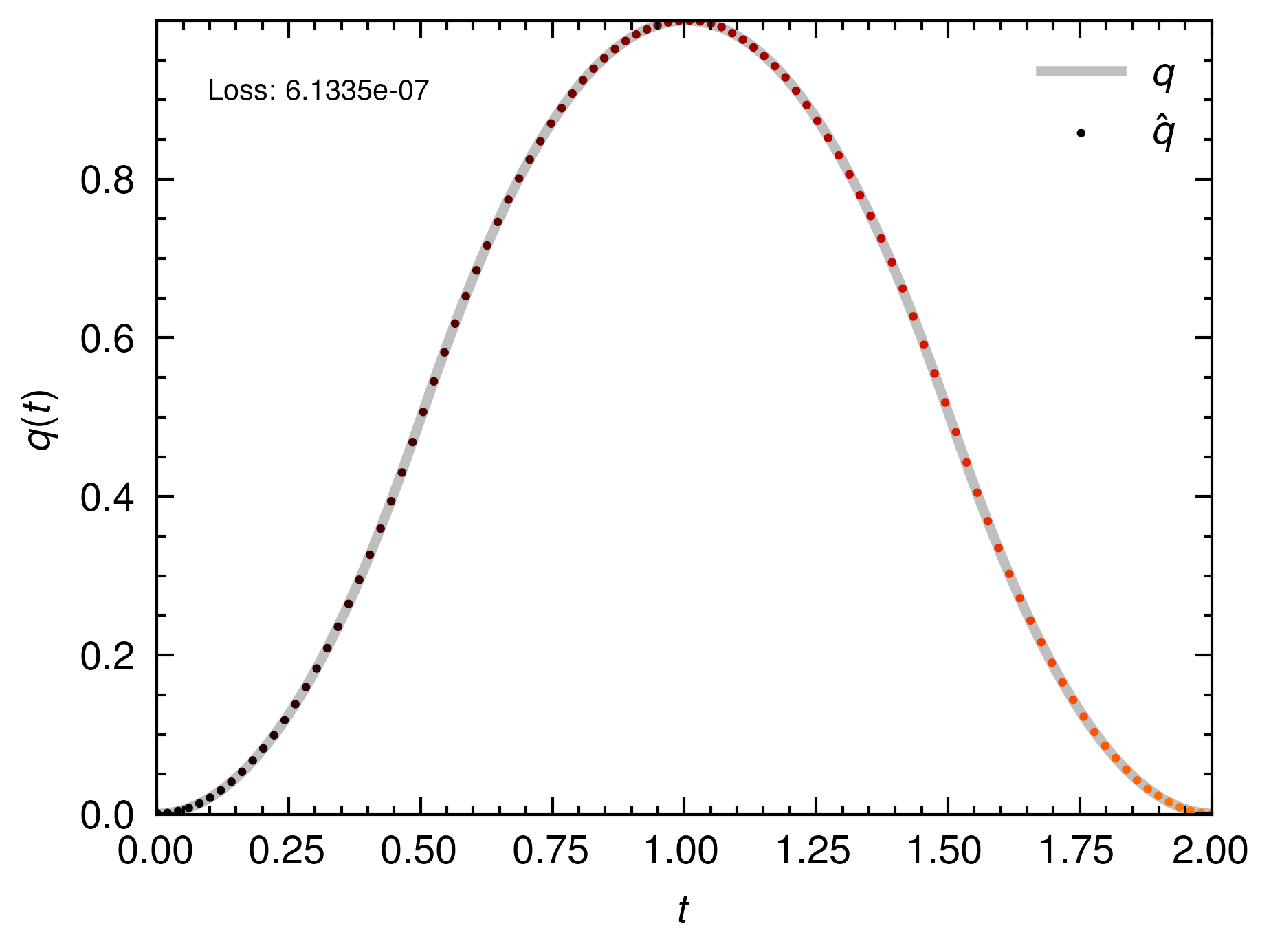}
    \end{subfigure}
    \hfill
    \begin{subfigure}[b]{0.35\textwidth}
        \includegraphics[width=\textwidth]{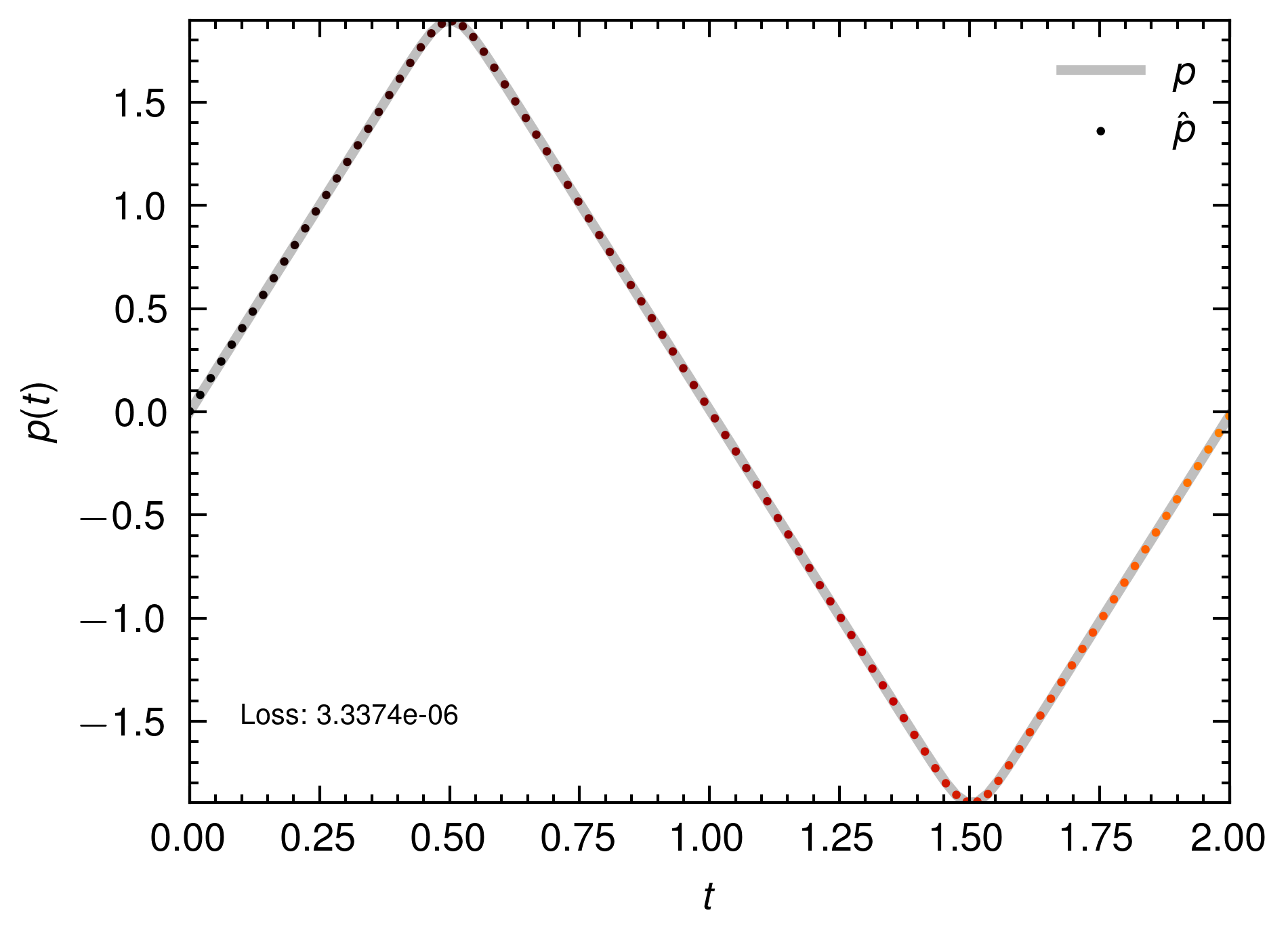}
    \end{subfigure}
    \caption{Test potentials and corresponding trajectories used in the experiments.
        Each row represents a different potential: Simple Harmonic Oscillator (SHO), Double-Well, Morse, Mirrored Free Fall (MFF), and Softened Mirrored Free Fall (SMFF), respectively.
        For position and momentum plots, the true values are represented by solid gray lines, while the predictions from MambONet trained on the extended dataset are shown as scatter points with colors transitioning from black to orange over time.The left column shows the potential function $V(x)$, the middle column shows the position $q(t)$, and the right column shows the momentum $p(t)$.}
    \label{fig:test_potentials}
    \end{adjustwidth}
\end{figure}

\end{document}